\documentclass[sigplan,10pt]{acmart}

\usepackage{subcaption}
\usepackage{algorithm}
\usepackage{algpseudocode}
\usepackage{xspace}
\usepackage{ulem}
\usepackage{booktabs}
\usepackage{bbm}
\usepackage{float}
\usepackage{caption}
\usepackage{wrapfig}
\AtBeginDocument{%
  \setcitestyle{maxbibnames=6}
}

\newcommand{\Feather}{\textsc{Feather}\xspace}

\copyrightyear{2026}\acmYear{2026}
\setcopyright{cc}
\setcctype{by}
\acmConference[SIGOPS ATC 2026]{The 2026 ACM SIGOPS Annual Technical Conference}{November 15-18, 2026}{Hong Kong, China}
\acmBooktitle{The 2026 ACM SIGOPS Annual Technical Conference (SIGOPS ATC 2026), November 15-18, 2026, Hong Kong, China}
\acmDOI{10.1145/3828523.3857087}
\acmISBN{979-8-4007-2818-1/2026/11}

\begin{document}
\title{Requests of a \Feather Must Flock Together: Batch Size vs. Prefix Homogeneity in LLM Inference}
\renewcommand{\shorttitle}{Requests of a \Feather Must Flock Together}

\author{Saksham Rathi}
\affiliation{%
  \institution{Indian Institute of Technology Bombay}
  \city{Mumbai}
  \country{India}}
\email{sakshamrathi@cse.iitb.ac.in}

\author{Preeti}
\affiliation{%
  \institution{Indian Institute of Technology Bombay}
  \city{Mumbai}
  \country{India}}
\email{preeti@cse.iitb.ac.in}

\author{Mythili Vutukuru}
\affiliation{%
  \institution{Indian Institute of Technology Bombay}
  \city{Mumbai}
  \country{India}}
\email{mythili@cse.iitb.ac.in}

\begin{abstract}
Auto-regressive token generation in large language models is memory-bound because it requires ``attending to'' key and value tensors (KV cache) of all previous tokens. Prior work aims to improve the efficiency of this \textit{decode} process by batching multiple requests together, and maximizing batch size subject to GPU memory constraints. The key observation of our work is that with prefix-sharing workloads, smaller, prefix-homogeneous batches -- where \textit{all} requests share a common prefix -- can achieve higher decode throughput than larger, heterogeneous batches, due to better spatial and temporal locality during KV cache accesses. However, current prefix-aware schedulers maximize prefix reuse within a batch only to reduce KV cache memory footprint, but do not stop batch formation at smaller homogeneous batches that could have performed better. Further, we show that shared prefix detection in existing schedulers relies on radix-tree traversals, incurring substantial CPU overhead that is often comparable to GPU execution time. This paper presents \Feather, a prefix-aware scheduler that uses reinforcement learning (RL) to learn the optimal tradeoff between batch size and prefix homogeneity. We also introduce Chunked Hash Tree (CHT), a lightweight data structure that enables fast prefix detection and efficient request selection for the RL scheduler, avoiding expensive tree traversals. We integrate \Feather into vLLM and SGLang, and our evaluation shows that \Feather achieves 2--10$\times$ higher end-to-end throughput as compared to existing schedulers by reducing the total number of KV cache accesses, while doing no worse than the status quo when there is insufficient prefix sharing. 
\end{abstract}


 
\maketitle

\section{Introduction}
\label{sec:intro}
Large language models (LLMs)~\cite{wei2022emergent, openai2024gpt4technicalreport,deepseekai2025deepseekv3technicalreport} are rapidly transforming modern computing, powering applications such as conversational assistants~\cite{dam2024completesurveyllmbasedai,10.1145/3580305.3599572}, code generation~\cite{hui2024qwen25codertechnicalreport}, and retrieval-augmented generation~\cite{NEURIPS2020_6b493230,yao2025cacheblendfastlargelanguage}. As these models transition from research to production, cloud AI engines~\cite{openai2024chatgpt,anthropic2024claude,google2024gemini} must improve serving latency and resource efficiency for both online and offline inference. While prior work has proposed scheduling strategies~\cite{pang2025optimizing,zhong2024distservedisaggregatingprefilldecoding,zheng2025bucketservebucketbaseddynamicbatching,agrawal2024tamingthroughputlatencytradeoffllm}, parallelization techniques~\cite{su2025seesawhighthroughputllminference,wu2024loongserveefficientlyservinglongcontext}, memory management methods~\cite{kwon2023efficient,prabhu2025vattentiondynamicmemorymanagement,zhang2025jengaeffectivememorymanagement}, quantization~\cite{chee2024quip2bitquantizationlarge,lin2024awqactivationawareweightquantization}, and kernel-level optimizations~\cite{yi2026pat,MLSYS2025_96894468}, efficiently serving LLM workloads, especially under realistic high-throughput conditions, remains an open problem.

\begin{figure}[t]
\captionsetup[subfigure]{labelformat=simple}
    \renewcommand\thesubfigure{(\alph{subfigure})}
\centering
\begin{subfigure}{0.55\columnwidth}
    \centering
    \includegraphics[width=\linewidth]{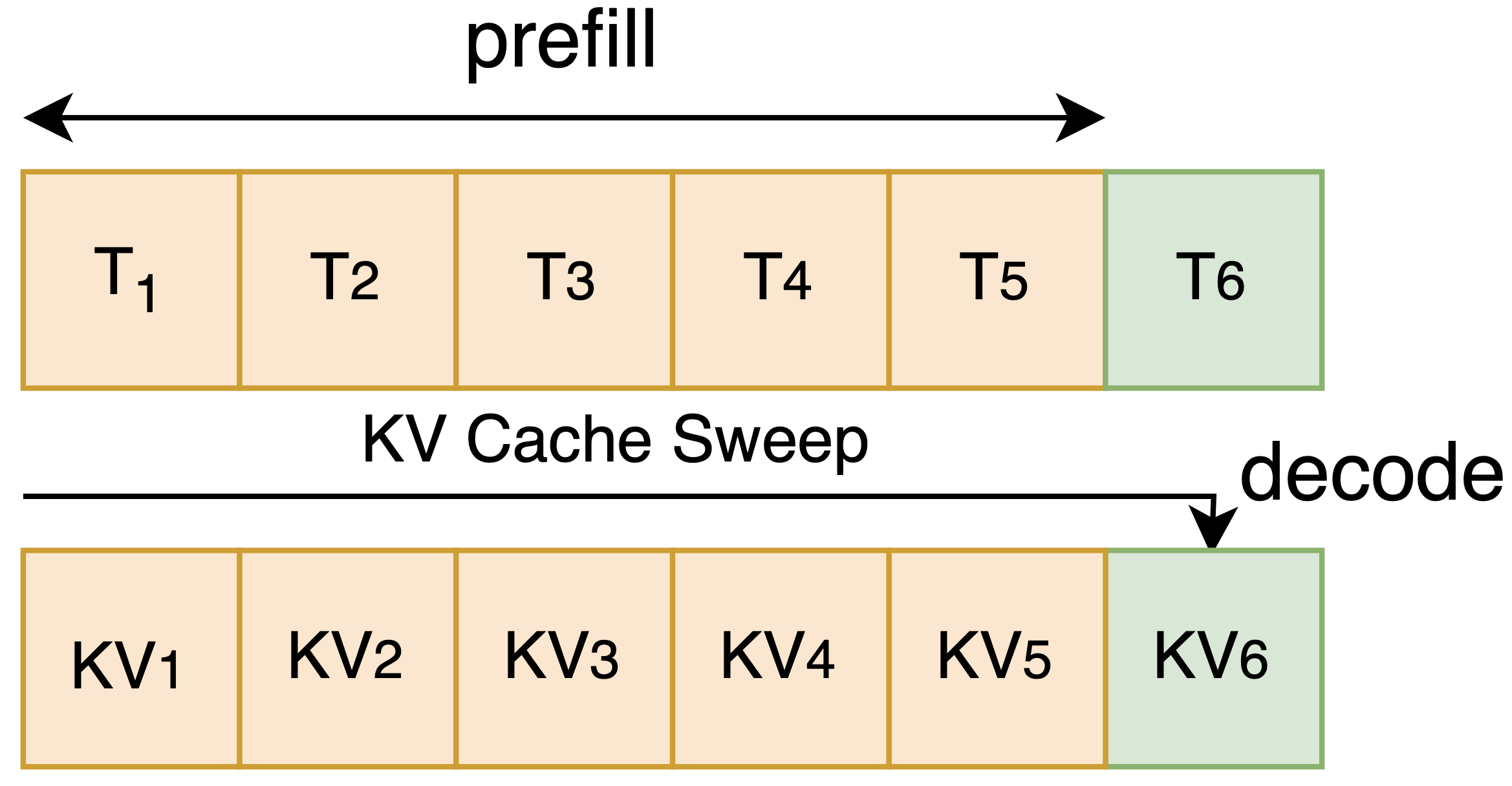}
    \caption{Prefill and Decode}
    \label{fig:prefill_decode}
\end{subfigure}
\hspace{5pt}
\begin{subfigure}{0.25\columnwidth}
    \centering
    \includegraphics[width=\linewidth]{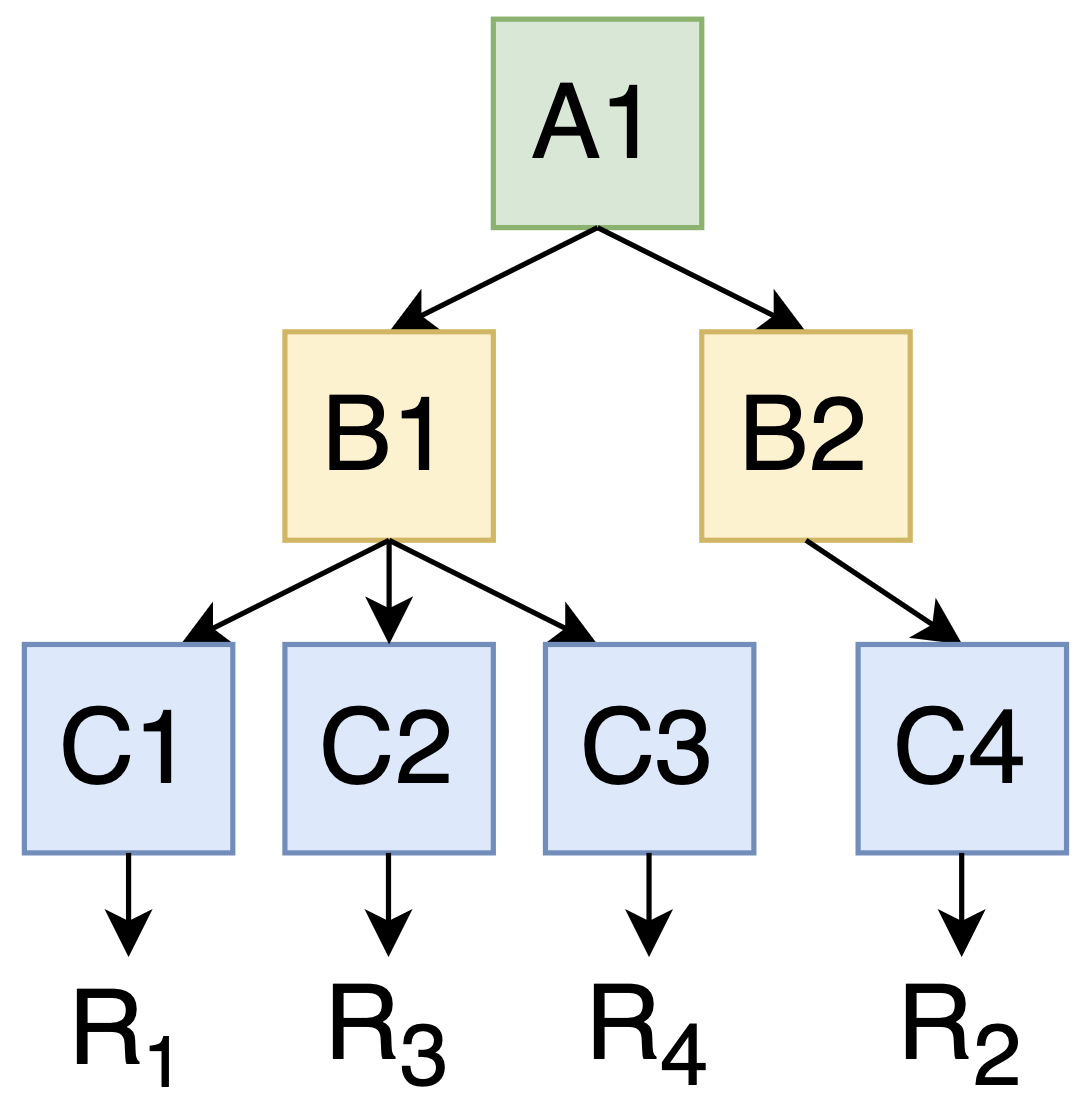}
    \caption{Radix Tree}
    \label{fig:radix_tree_four_requests}
\end{subfigure}
\caption{LLM Inference Primitives}
\label{fig:prefill_decode_radix_tree}

\end{figure}

An LLM inference request goes through two phases: \textit{prefill} and \textit{decode}. The prefill phase processes all prompt tokens in parallel to produce the first output token, making it compute-bound. The decode phase, by contrast, generates tokens auto-regressively and is fundamentally memory-bound. 
Each newly generated token attends to the keys and values of all previous tokens, resulting in a sweep of the entire KV cache for each decode iteration, as shown in Figure~\ref{fig:prefill_decode}. As context lengths grow, this memory traffic becomes the dominant bottleneck. 
Popular LLM inference engines like vLLM~\cite{kwon2023efficient} adopt batching to improve GPU utilization during the decode phase, scheduling multiple requests together in a batch using policies like First Come First Serve (FCFS). More advanced frameworks explore techniques such as interleaving~\cite{agrawal2024tamingthroughputlatencytradeoffllm}, disaggregation~\cite{zhong2024distservedisaggregatingprefilldecoding}, and spatial co-scheduling~\cite{Kamath_2025,lin2025boostingllmservingspatialtemporal}, along with dynamically adjusting batch sizes based on runtime constraints~\cite{pang2025optimizing}, to address this compute-memory mismatch. 

Orthogonally, workloads such as few-shot prompting~\cite{brown2020languagemodelsfewshotlearners,NEURIPS2021_5c049256}, system prompts~\cite{thebigpromptlibrary}, and RAG~\cite{NEURIPS2020_6b493230} exhibit significant \textit{prefix sharing} across requests. For instance, when an LLM agent repeatedly analyzes a massive codebase or a document repository, the same background is used as a prefix to the prompt given in every sub-task. As context lengths grow to millions of tokens, optimizing for these shared prefix workloads becomes critical to achieving good inference performance. Existing work primarily relies on reusing KV caches for shared prefixes, to eliminate redundant KV computation and reduce KV cache memory footprint. Systems such as SGLang~\cite{zheng2024sglangefficientexecutionstructured} and vLLM~\cite{kwon2023efficient} use radix trees (Figure~\ref{fig:radix_tree_four_requests}) or hashing-based structures to keep track of shared KV caches. While vLLM's FCFS would have scheduled requests in arrival order ($R_1, R_2, R_3, R_4$), SGLang performs a depth-first traversal of the radix tree and schedules requests in the order $R_1, R_3, R_4, R_2$, to maximize KV cache reuse and reduce KV cache evictions.
A complementary line of work~\cite{yi2026pat,yao2025deftdecodingflashtreeattention,ye2024chunkattentionefficientselfattentionprefixaware,MLSYS2025_96894468} exploits the shared prefix structure within the attention kernel itself, by jointly computing attention for groups of queries that share prefixes. Such prefix-aware attention kernels avoid redundant loading of shared KV blocks for each request separately, thereby reducing memory traffic.

Despite these advances, we identify two critical gaps in current scheduling strategies for prefix-shared workloads. First, prior work focuses on maximizing batch size as the primary knob to improve token generation throughput in the decode phase. However, we show via controlled experiments that prefix homogeneity---the extent to which \textit{all} requests in a batch share a common prefix---has an even stronger impact on performance. Further, we find that the performance gains increase with the length of the prefix shared across all the requests in a batch (\S\ref{subsec:significance_of_prefix_homogeneity}). This is because when a prefix is shared by all requests, there is significant \textit{spatial} and \textit{temporal} locality in the memory accesses of the shared KV cache located in the GPU DRAM. Such highly regular and sequential memory access patterns allow hardware prefetchers to detect uniform strides, and proactively move data from DRAM to the L2 cache. These benefits of long sequential memory access patterns to GPU DRAM have been documented in prior work as well~\cite{7445236, Peng_2017}, though not in the context of prefix-homogeneous batches. We show that these locality benefits eventually translate to both higher memory bandwidth utilization and lower total amount of data fetched from DRAM, and hence higher decode throughput.

\begin{figure}[t]
  \centering
  \includegraphics[width=\linewidth]{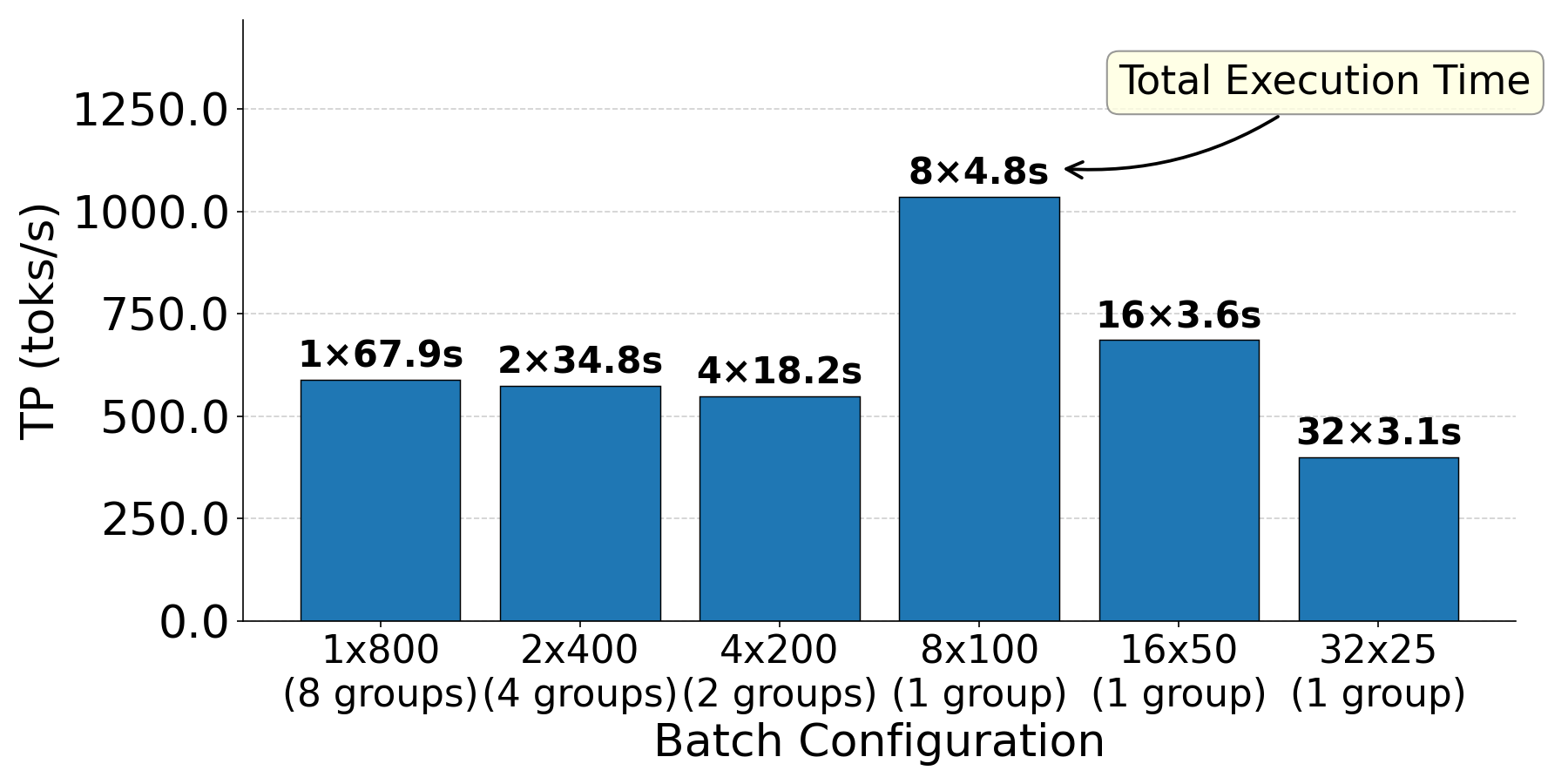}
  \caption{Batch Size vs. Prefix Homogeneity}
  \label{fig:prefill_decode_homo_hetero_grouped_throughput_tok100}
  
\end{figure}

We also show that moderately small prefix-homogeneous batches can leverage these locality benefits and perform better than larger heterogeneous batches (\S\ref{subsec:batch_size_vs_prefix_homogeneity}). For example, Figure~\ref{fig:prefill_decode_homo_hetero_grouped_throughput_tok100} shows the throughput and execution times of 800 requests, where 8 groups of 100 requests each share a large prefix of 10K tokens between them. Processing these requests in 8 prefix-homogeneous batches of 100 requests each leads to nearly 2$\times$ higher throughput than when using larger heterogeneous batches, though very small homogeneous batches (32 batches of 25 requests each) perform poorly as well due to sub-optimal compute utilization. Existing prefix-aware schedulers have no mechanism to recognize this tradeoff between batch size and prefix homogeneity; e.g., in Figure~\ref{fig:radix_tree_four_requests}, a DFS-traversal-based scheduler will use $R_2$ to maximize batch size, even if a smaller batch (only $R_1, R_3, R_4$) could have performed better due to a longer common prefix. 

The second big gap with current prefix-aware schedulers is that they incur prohibitive CPU overheads in maintaining and traversing radix trees, which can account for 50--90\% of total latency (\S\ref{subsec:overhead_dynamic_prefix_detection}). Even though radix tree based schedulers optimize the GPU execution time through better reuse of KV caches as compared to simpler schedulers like FCFS, we show that the additional overheads on the critical execution path to detect prefix sharing, all done while the GPU is waiting, result in an overall worse end-to-end performance.

This paper introduces \Feather\footnote{Please see \S\ref{sec:about_feather} for an explanation of this name.} (\S\ref{sec:design}), a prefix-aware scheduler that improves throughput by efficiently finding an optimal balance between batch size and prefix homogeneity. \Feather is built on two components: \textit{(i) Chunked Hash Tree (CHT)}, a lightweight data structure for fast prefix detection, similar in spirit to vLLM's block table~\cite{kwon2023efficient}. Instead of performing token-level radix tree traversals, CHT represents each request as a vector of cumulative prefix hashes. To efficiently select a request that maximizes the length of the prefix shared across the active batch, CHT maintains a \textit{min-heap}, updated lazily, and ordered by per-request \textit{missing count}---the number of prefix chunks not yet covered by the batch. This enables the identification of the best candidate in $\mathcal{O}(\log W)$ time among $W$ waiting requests, a significant improvement over DFS traversal~\cite{zheng2024sglangefficientexecutionstructured}. \textit{(ii) RL-based batching policy}. Deciding when to stop adding requests to a batch presents a trade-off: larger batches improve utilization but dilute prefix homogeneity. We cast this as a reinforcement learning problem in which the scheduler observes batch characteristics (e.g., current size, shared prefix length), evaluates the impact of adding another request to the batch, and stops batch formation at a point that maximizes throughput. This adaptive approach consistently outperforms static heuristics across diverse workloads and hardware configurations. Together, these two components enable \Feather to construct prefix-homogeneous batches efficiently, both improving system throughput and reducing scheduling overhead.

We implement \Feather on vLLM~\cite{kwon2023efficient} and SGLang~\cite{zheng2024sglangefficientexecutionstructured} and evaluate it across diverse workloads and configurations (\S\ref{sec:eval}). Our results show that \Feather achieves $2$--$10\times$ higher end-to-end throughput over vLLM's FCFS and other baselines for shared prefix lengths ranging from 1K to 10K tokens, while also improving DRAM bandwidth utilization. \Feather is fully hardware-agnostic and yet performs better than recent prefix-aware kernels that must be tuned to specific hardware configurations. 
Our RL policy ensures that our performance never falls below that of the baselines, even if the workload does not allow for prefix-homogeneous batches, and CHT reduces prefix-detection overhead by up to $1000\times$ compared to DFS-based approaches for long sequences. 

In summary, this paper makes the following contributions: (i) We demonstrate the tradeoff between batch size and prefix homogeneity in the decode phase performance of LLM inference, and identify previously overlooked CPU overheads in prefix-aware scheduling. (ii) We design the Chunked Hash Tree (CHT) data structure for low-overhead prefix detection, and develop a reinforcement learning-based policy for adaptive batch construction. (iii) We build \Feather, a prefix-aware scheduler integrated into vLLM and SGLang, and demonstrate improvements in throughput and latency on diverse workloads. Our code is available at \textcolor{blue}{\url{https://github.com/networkedsystemsIITB/Feather}}.

\section{Background and Related Work}
\label{sec:background}
We now discuss the background and prior work on batching, scheduling, and prefix-aware optimizations.

\subsection{Background}

\subsubsection*{LLM Inference and Attention}
LLM inference proceeds auto-regressively in two phases: prefill and decode. Their latencies are known as \textit{time to first token} (TTFT) and \textit{time between tokens} (TBT), respectively. Each decode step comprises self-attention, Query-Key-Value-Output (QKVO) projections, and Multi-Layer Perceptron (MLP) layers. Self-attention captures global token dependencies as:

\[
\text{Attention}(\mathbf{Q}, \mathbf{K}, \mathbf{V}) = \text{softmax}\left(\frac{\mathbf{Q}\mathbf{K}^T}{\sqrt{d}}\right)\mathbf{V}
\]
where $d$ denotes the head dimension. During the decode phase, the query $\mathbf{Q}$ of the current token attends to all prior keys $\mathbf{K}$ and values $\mathbf{V}$. These are cached in GPU memory as the \textbf{KV cache}, eliminating redundant computation at the cost of increased memory usage. The asymmetry between prefill and decode, together with the growing memory footprint of the KV cache, motivates much of the prior work.

\subsubsection*{GPU Execution Model}
GPUs are hierarchically organized for parallelism across hundreds of thousands of threads. The basic compute unit is the Streaming Multiprocessor (SM), and modern GPUs typically contain a few hundred SMs. Each SM includes an L1 cache, shared memory, tensor cores for matrix multiplication (GEMM), and execution units. SMs access global memory through a shared L2 cache. This hierarchy, from small, fast on-chip memory to large, high-latency global memory, is critical for LLM serving, as KV cache access patterns directly influence memory bandwidth utilization. Like CPUs, GPUs use hardware prefetching and benefit from regular, sequential memory access patterns~\cite{7445236, Peng_2017}.

\subsection{Related Work}
\subsubsection*{Batching and Scheduling}
vLLM~\cite{kwon2023efficient} adopts a first-come, first-served (FCFS) scheduling policy as a practical baseline for LLM serving. Sarathi-Serve~\cite{agrawal2024tamingthroughputlatencytradeoffllm} reduces prefill-decode interference through \textit{chunked prefills} and stall-free batching, while DistServe~\cite{zhong2024distservedisaggregatingprefilldecoding} eliminates it by disaggregating the two phases across GPUs. At the kernel level, POD-Attention~\cite{Kamath_2025} overlaps prefill and decode on a single GPU via fused attention with SM-aware scheduling, and Bullet~\cite{lin2025boostingllmservingspatialtemporal} enables fine-grained resource sharing through dynamic SM allocation. At the request level, BucketServe~\cite{zheng2025bucketservebucketbaseddynamicbatching} groups requests by sequence length to improve batching efficiency, and \citet{pang2025optimizing} dynamically adapts the batch size to runtime constraints. Complementary to prior work, \textit{\Feather operates at the scheduler level and learns when to stop batch construction based on prefix homogeneity, introducing a scheduling signal not explicitly exploited in existing systems.}

\subsubsection*{Prefix Sharing}
\begin{figure}[t]
    \centering
    \includegraphics[width=\linewidth]{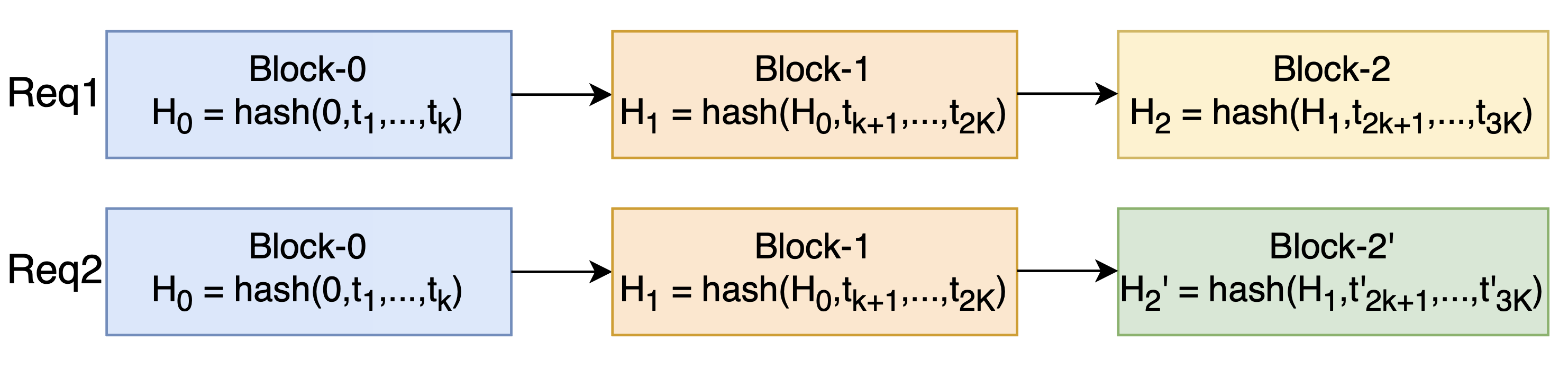}
    \caption{An Example of a vLLM Merkle Tree}
    \label{fig:vllm_merkle_tree_example}
    
\end{figure}

Many real-world workloads~\cite{brown2020languagemodelsfewshotlearners,NEURIPS2021_5c049256,thebigpromptlibrary,NEURIPS2020_6b493230} exhibit substantial prefix overlap, and enabling KV cache reuse avoids redundant computation and reduces memory footprint. However, realizing these benefits requires coordination between serving and memory management. SGLang \cite{zheng2024sglangefficientexecutionstructured} enables prefix sharing via \textit{RadixAttention}, which uses a radix tree to organize token sequences, and enables KV reuse across shared prefixes. Similarly, vLLM~\cite{kwon2023efficient} implements prefix caching~\cite{vllm_prefix_caching_docs} using a Merkle-tree-inspired hashing scheme (Figure~\ref{fig:vllm_merkle_tree_example}), where each KV block hash depends on its tokens and preceding prefix hashes, enabling reuse when hash sequences match. BatchLLM~\cite{zheng2025batchllmoptimizinglargebatched} targets large offline workloads by grouping prefix-sharing requests and reordering them to maximize KV reuse. BlendServe~\cite{zhao2024blendserveoptimizingofflineinference} combines a resource-aware prefix tree with scheduling to jointly optimize prefix reuse and compute–memory demand. \textit{\Feather uses chunked hashing instead of radix trees or hash chains, for more efficient prefix-homogeneous batch construction.}

\subsubsection*{Prefix-Aware Attention Kernels}
Prior work~\cite{yi2026pat,yao2025deftdecodingflashtreeattention,ye2024chunkattentionefficientselfattentionprefixaware,MLSYS2025_96894468} also optimizes attention kernels by grouping queries with shared prefixes to load KV blocks once and reuse them across queries. These approaches, however, rely on GPU-specific kernel designs. In contrast, \textit{\Feather improves inputs to the existing attention kernels by constructing prefix-homogeneous batches upfront, making it hardware-agnostic and naturally composable with kernel-level techniques.}

\begin{figure*}[t]
\captionsetup[subfigure]{labelformat=simple}
\renewcommand\thesubfigure{(\alph{subfigure})}
\centering
\begin{subfigure}{0.43\textwidth}
    \centering
    \includegraphics[width=\linewidth]{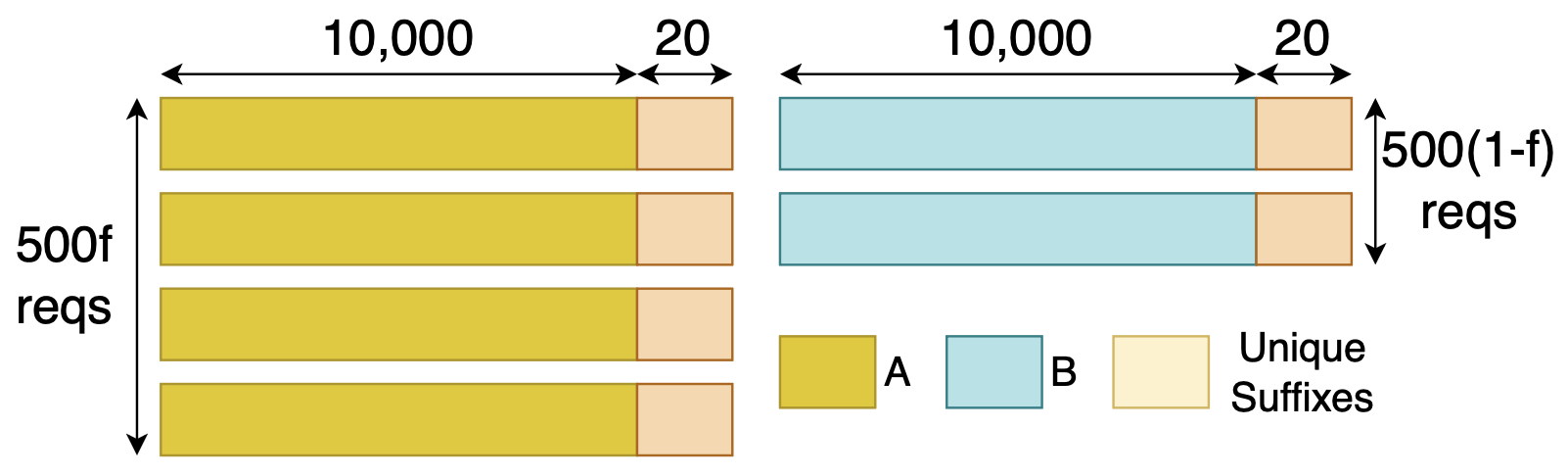}
    \caption{Visualization}
    \label{fig:two_fam_vis}
\end{subfigure}
\hfill
\begin{subfigure}{0.25\textwidth}
    \centering
    \includegraphics[width=\linewidth]{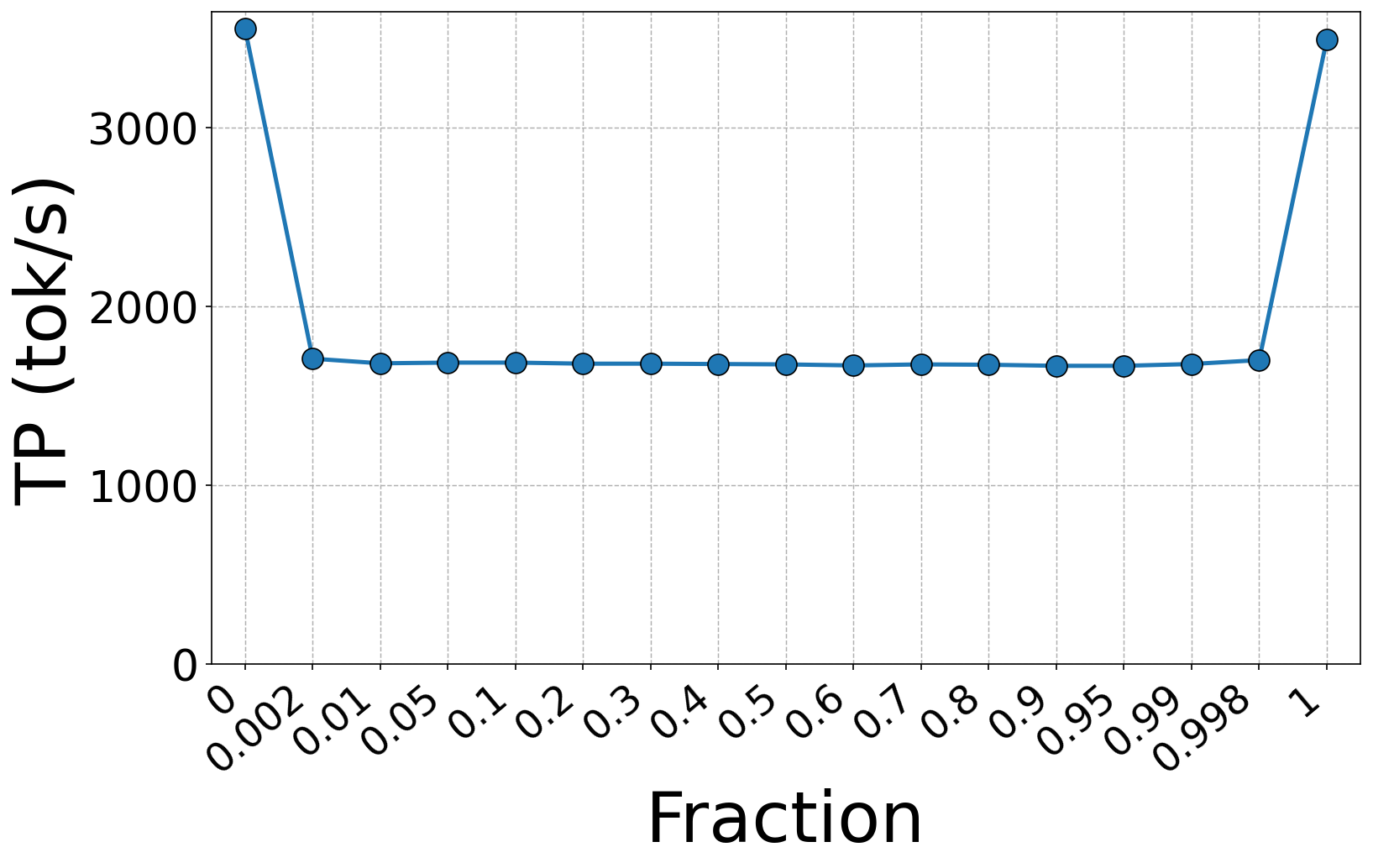}
    \caption{Throughput}
    \label{fig:two_fam_tp}
\end{subfigure}
\hfill
\begin{subfigure}{0.25\textwidth}
    \includegraphics[width=\linewidth]{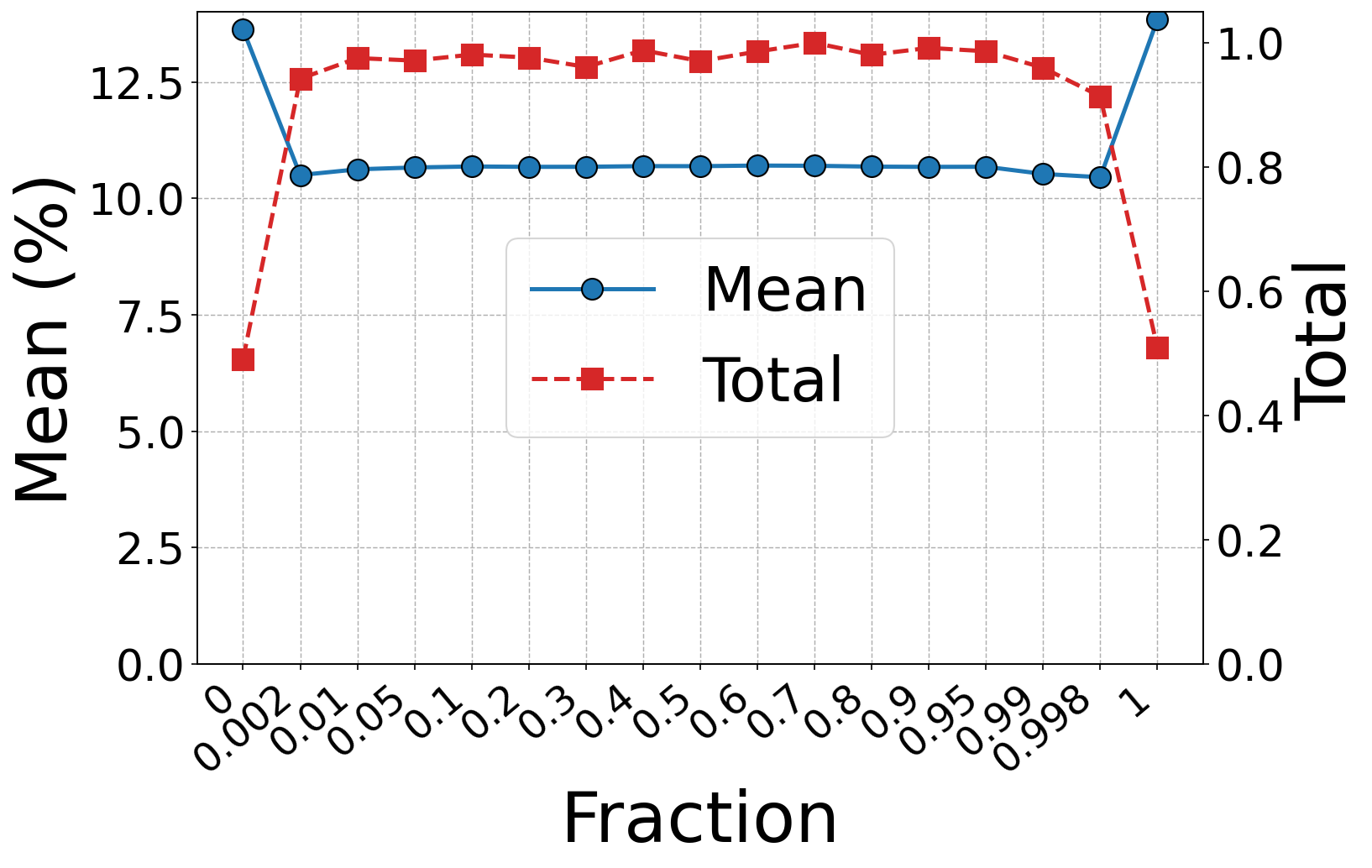}
\caption{Bandwidth Utilization}
\label{fig:two_fam_mean_dram}
\end{subfigure}
\caption{Prefix Homogeneity - Two Prefix Groups}
\label{fig:two_fam_fraction}

\end{figure*}

\section{Motivation}
\label{sec:motivation}
Through a set of controlled experiments (here and in \S\ref{sec:additional_results_section_3}), we validate three hypotheses that underpin the design of \Feather: (i) prefix homogeneity, the extent to which \textit{all} requests in a batch share a common prefix, has a strong impact on the decode throughput because it increases the locality of KV cache accesses. Importantly, these gains can be achieved \textit{without any modifications to existing attention kernels}. (ii) There is a trade-off between batch size and prefix homogeneity, where moderately sized homogeneous batches \textit{can} outperform large heterogeneous batches; and (iii) existing prefix detection approaches incur substantial CPU overhead.

\subsubsection*{Experimental Setup} 
All experiments use the Llama 3 8B \cite{llama3modelcard} model and are run on an RTX 6000 Ada GPU~\cite{nvidia_rtx6000_datasheet} with 48 GB of GDDR6 memory and a 96 MB L2 cache. We use the vLLM \cite{kwon2023efficient} inference engine with a default maximum batch size of 500 unless stated otherwise. We first focus on decode throughput, measured in output tokens per second. To isolate decode behavior, we perform a warm-up run so that the prefix KV caches are resident in GPU memory. This ensures that per-token compute remains constant within an experiment and differences arise primarily from memory accesses and batching. We also report end-to-end throughput, which includes the time taken to prefill the queries. For this model, a 10k-token request generates 1.2 GB of KV cache data across all layers. Within each layer, vLLM uses its paged memory management system and allocates the KV cache in blocks of 16 tokens contiguously in GPU memory, which is sufficient for locality benefits to kick in. We sample requests from the L-Eval~\cite{an2023leval} dataset and prune the size of each request as required by the experiment. We use the same FlashAttention~\cite{dao2022flashattentionfastmemoryefficientexact} backend (FlashAttention-2, via vLLM~0.13.0) throughout all the experiments, so that observed throughput differences are attributable only to batch composition rather than kernel-level variation.

\subsection{Significance of Prefix Homogeneity}
\label{subsec:significance_of_prefix_homogeneity}
We start by observing the impact of prefix homogeneity on decode phase throughput. When \textit{all} requests in a batch share a common prefix, their attention operations traverse the same KV cache regions concurrently. This yields two beneficial effects: (i) spatial locality: Sequential access to contiguous KV tensors enables the hardware prefetcher to fetch upcoming cache lines from DRAM into L2~\cite{7445236, Peng_2017}; (ii) temporal locality: Cache lines fetched for one request are reused by neighboring requests before eviction. Together, these effects increase L2 hit rates and DRAM utilization, improving throughput. In contrast, heterogeneous prefixes interleave KV accesses across disjoint regions, disrupting locality and reducing effective memory bandwidth. We will now present experiments that validate these claims.

\begin{figure*}[t]
\centering
\begin{minipage}{0.38\textwidth}
    \centering
\includegraphics[width=\columnwidth]{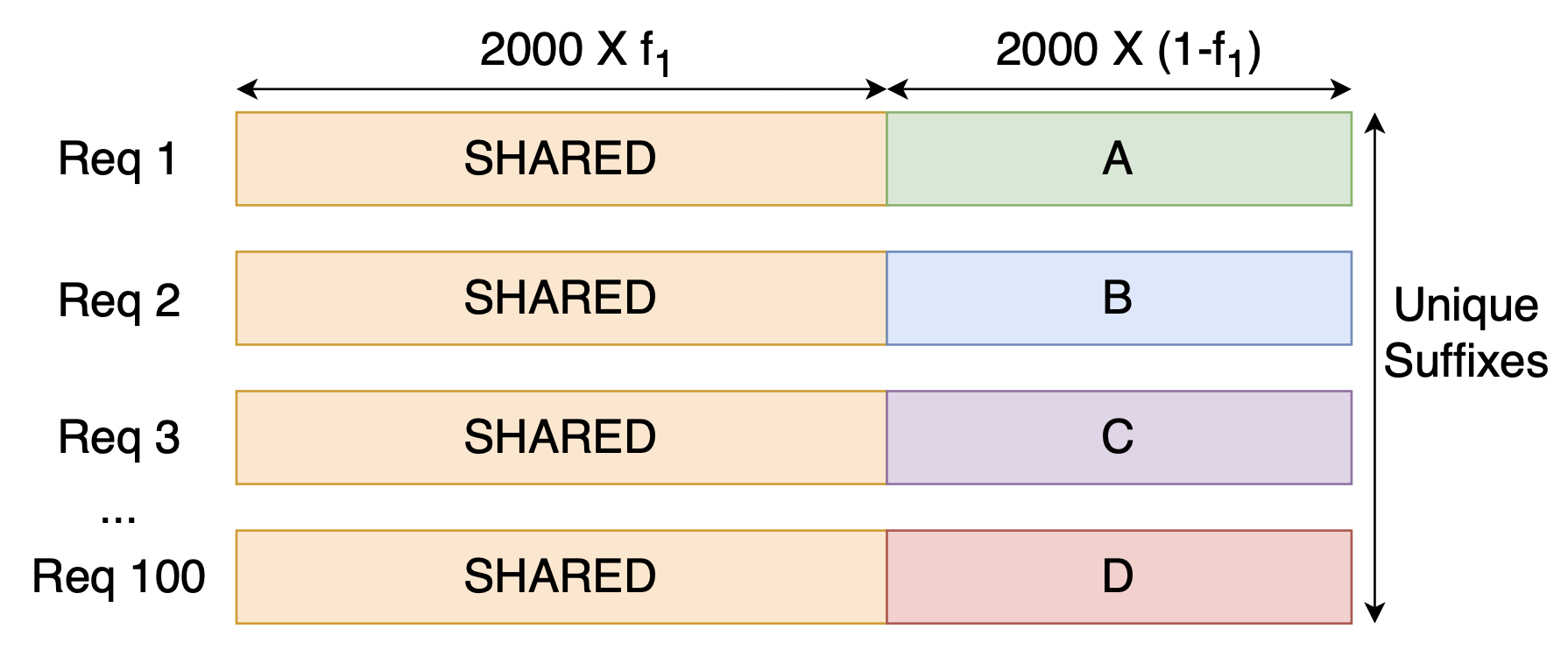}
\caption{Fractional Sharing Visualization}
\label{fig:frac_share_vis}
\end{minipage}
\hfill
\begin{minipage}{0.28\textwidth}
    \centering
    \includegraphics[width=\columnwidth]{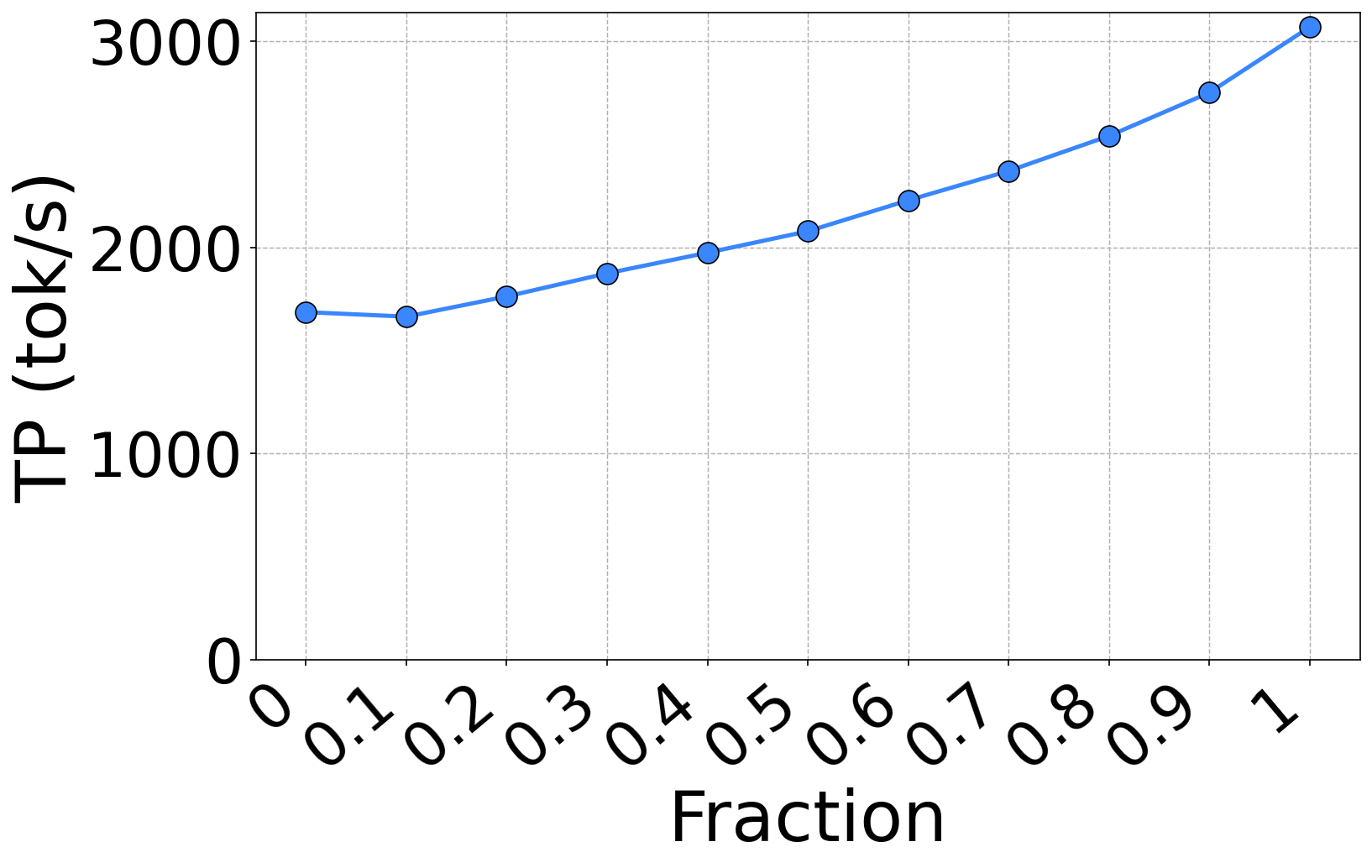}
    \caption{Throughput vs. Fraction of Prefix Shared}
    \label{fig:frac_share_tp}
\end{minipage}
\hfill
\begin{minipage}{0.32\textwidth}
    \centering
    \includegraphics[width=\columnwidth]{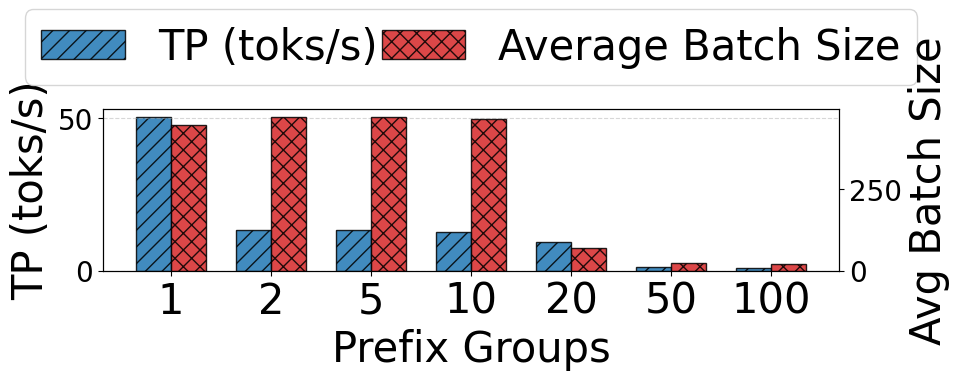}
    \captionof{figure}{Number of Prefix Groups}
    \label{fig:hbm_evic_tp_bs}
\end{minipage}

\end{figure*}

\textbf{Experiment 1:} We consider two prefixes, $A$ and $B$, each 10K tokens long. We form a batch of $500$ requests, each using the prefix $A$ or $B$, followed by a unique suffix of length $20$ tokens (Figure~\ref{fig:two_fam_vis}). Each request generates 50 decode tokens. Let $f \in [0,1]$ denote the fraction of requests using prefix $A$. When $f$ is 0 or 1, the batch is homogeneous; otherwise, two prefix groups coexist. We perform a warm-up run so that KV caches for both prefixes are resident in GPU memory.

Figure~\ref{fig:two_fam_tp} shows decode throughput as a function of $f$. We see that the throughput is highest for fully homogeneous batches. However, even a single request from the other prefix group (e.g., $f = 0.002$) causes throughput to drop by nearly $2\times$. This demonstrates that decode throughput is sensitive to even minimal prefix heterogeneity. To understand this effect, we capture DRAM bandwidth utilization using DCGMI~\cite{nvidia_dcgm_2026}. We first plot the mean bandwidth utilization for each $f$, and then we sum the utilization over the entire experiment (i.e., compute the area under the memory bandwidth curve) to approximate total memory accesses. As shown in Figure~\ref{fig:two_fam_mean_dram}, homogeneous batches achieve over 40\% higher bandwidth utilization than mixed-prefix batches due to greater spatial locality, enabling more efficient hardware-level prefetching into the GPU cache. Furthermore, greater temporal locality in prefix-homogeneous batches reduces the total data fetched from DRAM. Because most SMs process queries from the dominant prefix even when $f = 0.002$, we attribute the performance gains primarily to improved L2 cache reuse rather than increased L1 cache locality within an SM.

\textbf{Takeaway 1:} \textit{Decode throughput is maximized when \textit{all} requests share a prefix; even a single divergent prefix disrupts locality and sharply reduces effective bandwidth.}

\textbf{Experiment 2:} Next, we evaluate the impact of the shared prefix length on performance. For this, we construct a batch of 100 requests, each 2K tokens long and sharing a common prefix of length $2K \times f$ tokens, where $f \in [0, 1]$. The remaining $2K \times (1-f)$ tokens are unique across requests (Figure~\ref{fig:frac_share_vis}). The KV caches fit into GPU memory for all values of $f$. Figure~\ref{fig:frac_share_tp} shows that throughput increases smoothly with $f$. It is lowest at $f = 0$ and improves steadily as sharing increases, reaching over 60\% higher throughput at $f = 1$. As $f$ grows, a larger portion of the KV cache is shared across requests; since decode repeatedly traverses the KV cache, greater spatial and temporal locality improve throughput.

\textbf{Takeaway 2:} \textit{In prefix-homogeneous batches, decode throughput increases smoothly with the length of the shared prefix.}

\textbf{Experiment 3:} Next, we evaluate the effect of the number of prefix groups on performance using an experiment with $N$ shared prefixes $P_1, P_2, \dots, P_N$, each of length $10K$ tokens and having same number of requests. Figure~\ref{fig:hbm_evic_tp_bs} shows throughput and average batch size as $N$ increases. As observed earlier, throughput drops from $N = 1$ to $N = 2$ due to reduced temporal and spatial locality arising from prefix heterogeneity. However, from $N = 2$ to $N = 20$, throughput remains stable, indicating that additional heterogeneity has a limited impact once locality is disrupted. Finally, throughput drops sharply from $N = 20$ to $N = 100$, as GPU memory can no longer hold KV caches for all active prefixes. The impact is also visible through the reduced average batch size for $N = 50$ and $N = 100$. Prior work~\cite{zheng2025batchllmoptimizinglargebatched,zheng2024sglangefficientexecutionstructured} mitigates the throughput drop from $N = 20$ to $N = 50$ due to KV cache evictions by maximizing prefix reuse and minimizing memory footprint using depth-first traversal of radix trees. However, to the best of our knowledge, no prior work has observed the drop in performance from $N = 1$ to $N = 2$ prefix groups.

\textbf{Takeaway 3:} \textit{While prefix heterogeneity causes a drop in throughput, the amount of heterogeneity has a marginal impact on the loss of locality. However, very high heterogeneity leads to minimal KV reuse and higher GPU memory evictions.}

\begin{figure}[t]
    \centering
    \captionsetup[subfigure]{labelformat=simple}
    \renewcommand\thesubfigure{(\alph{subfigure})}
    \begin{subfigure}{0.12\columnwidth}
        \centering
        \includegraphics[width=\linewidth]{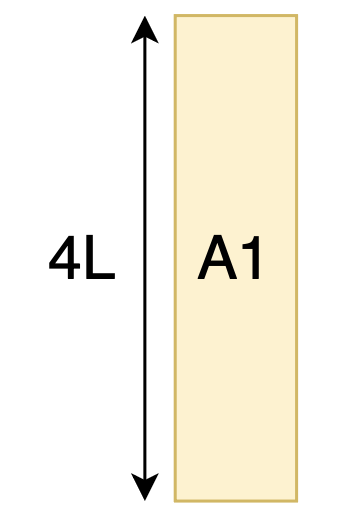}
        \caption{1-level}
        \label{fig:vis_complete_sharing}
    \end{subfigure}
    \hfill
    \begin{subfigure}{0.17\columnwidth}
        \centering
        \includegraphics[width=\linewidth]{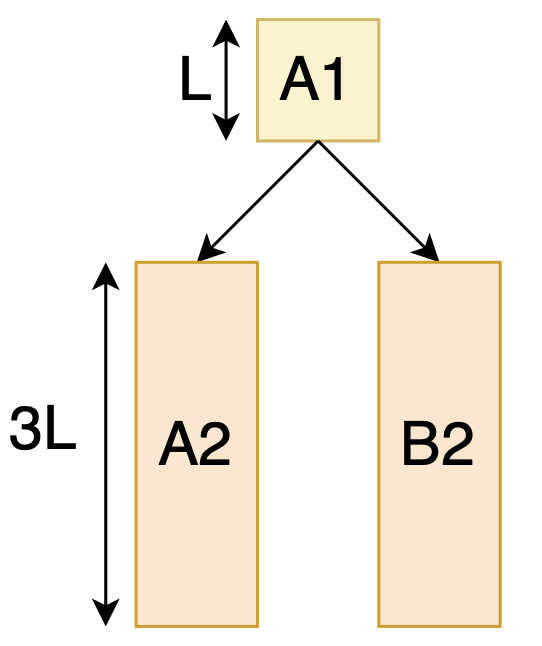}
        \caption{2 levels}
        \label{fig:2_levels}
    \end{subfigure}
    \hfill
    \begin{subfigure}{0.27\columnwidth}
        \centering
        \includegraphics[width=\linewidth]{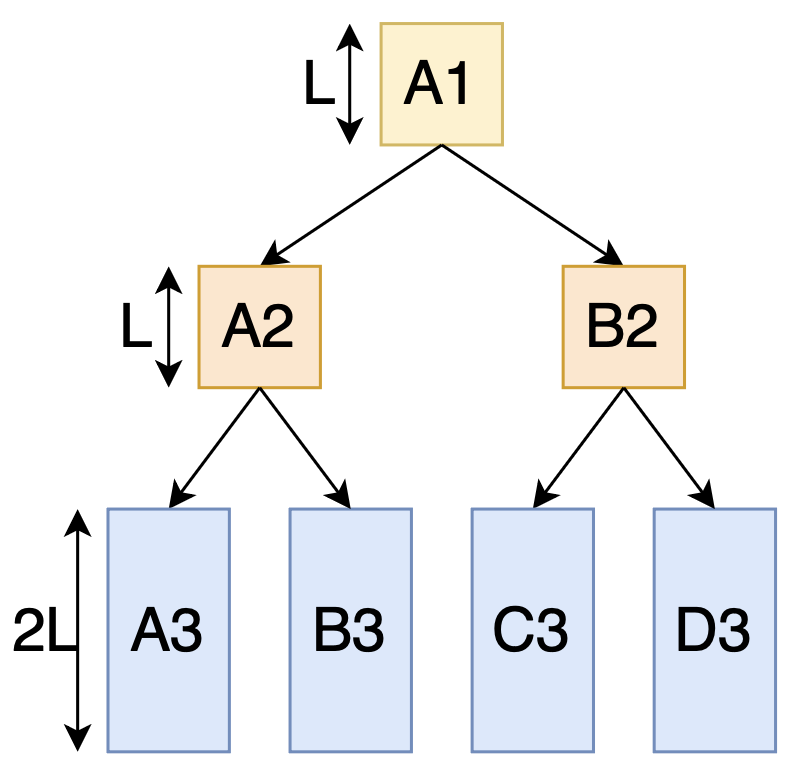}
        \caption{3 levels}
        \label{fig:3_levels}
    \end{subfigure}
    \hfill
    \begin{subfigure}{0.40\columnwidth}
        \centering
        \includegraphics[width=\linewidth]{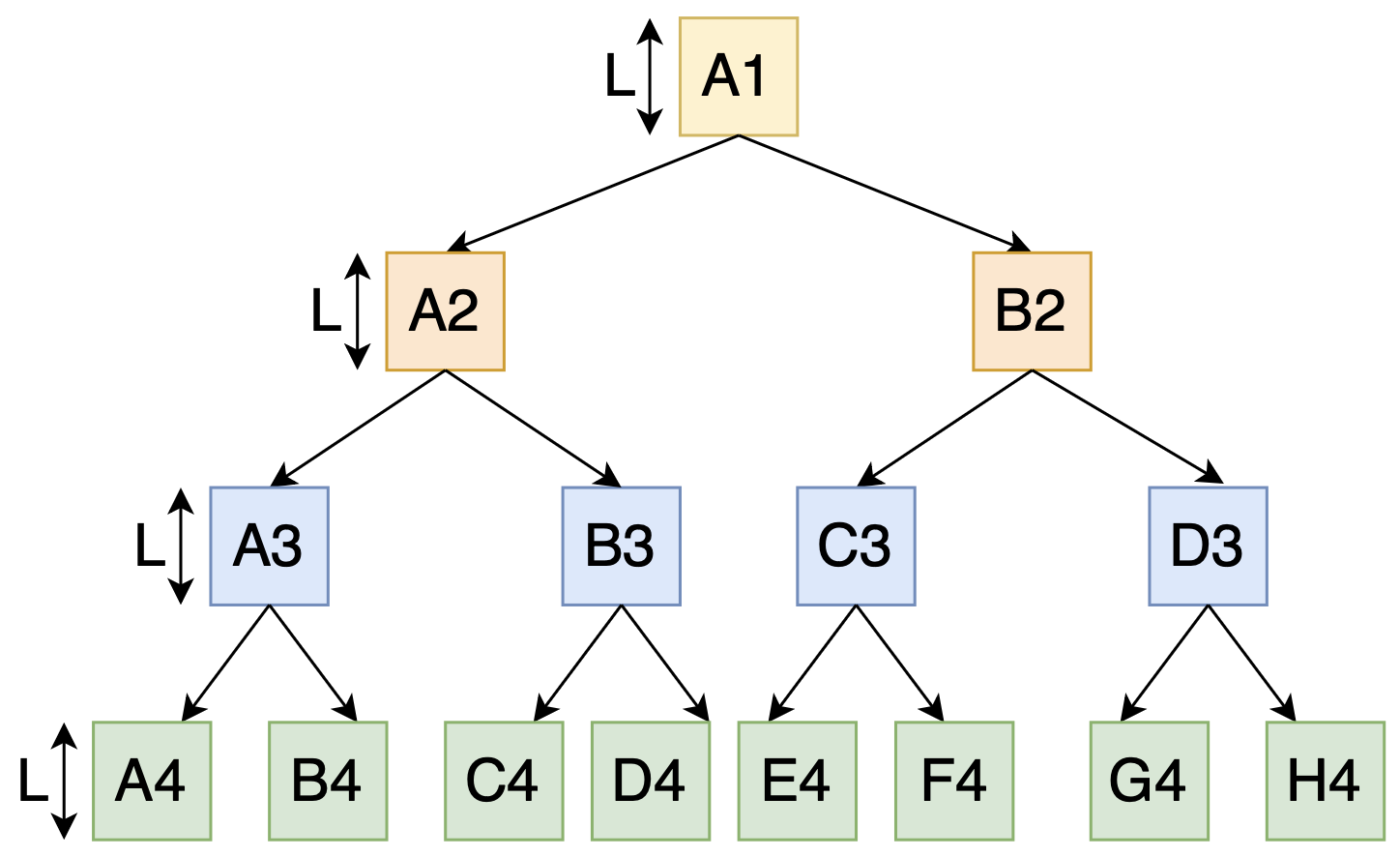}
        \caption{4 levels}
        \label{fig:4_levels}
    \end{subfigure}
    \caption{Radix Tree Sharing Levels}
    \label{fig:sharing_configurations}
    
\end{figure}

\begin{figure*}[t]
\centering
\begin{minipage}{0.30\textwidth}
\centering
    \includegraphics[width=\columnwidth]{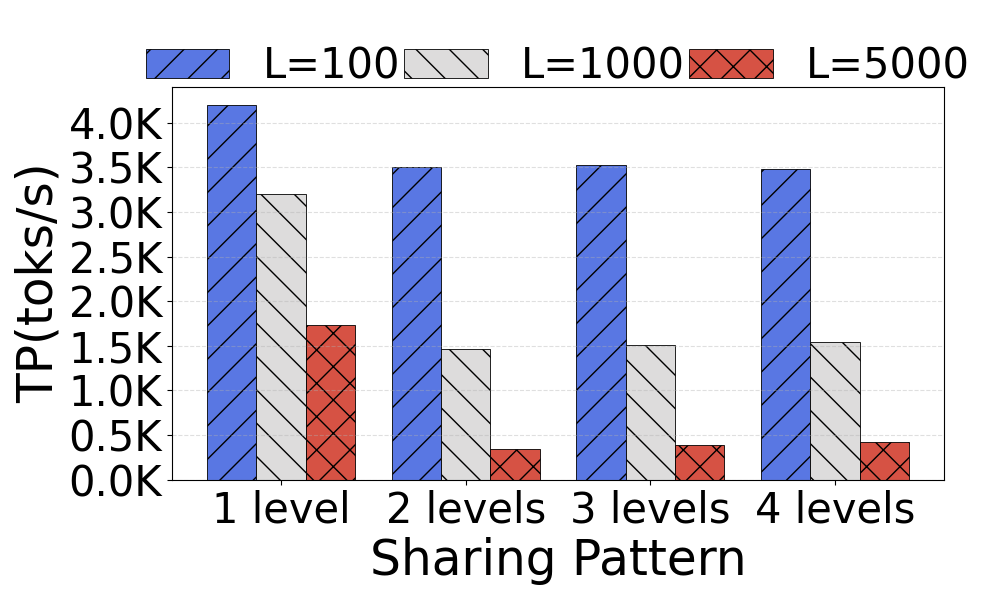}
    \caption{Radix Tree Sharing Patterns}
    \label{fig:tp_vs_l_1_9_11}
\end{minipage}
\hfill
\begin{minipage}{0.34\textwidth}
    \centering
    \includegraphics[width=\columnwidth]{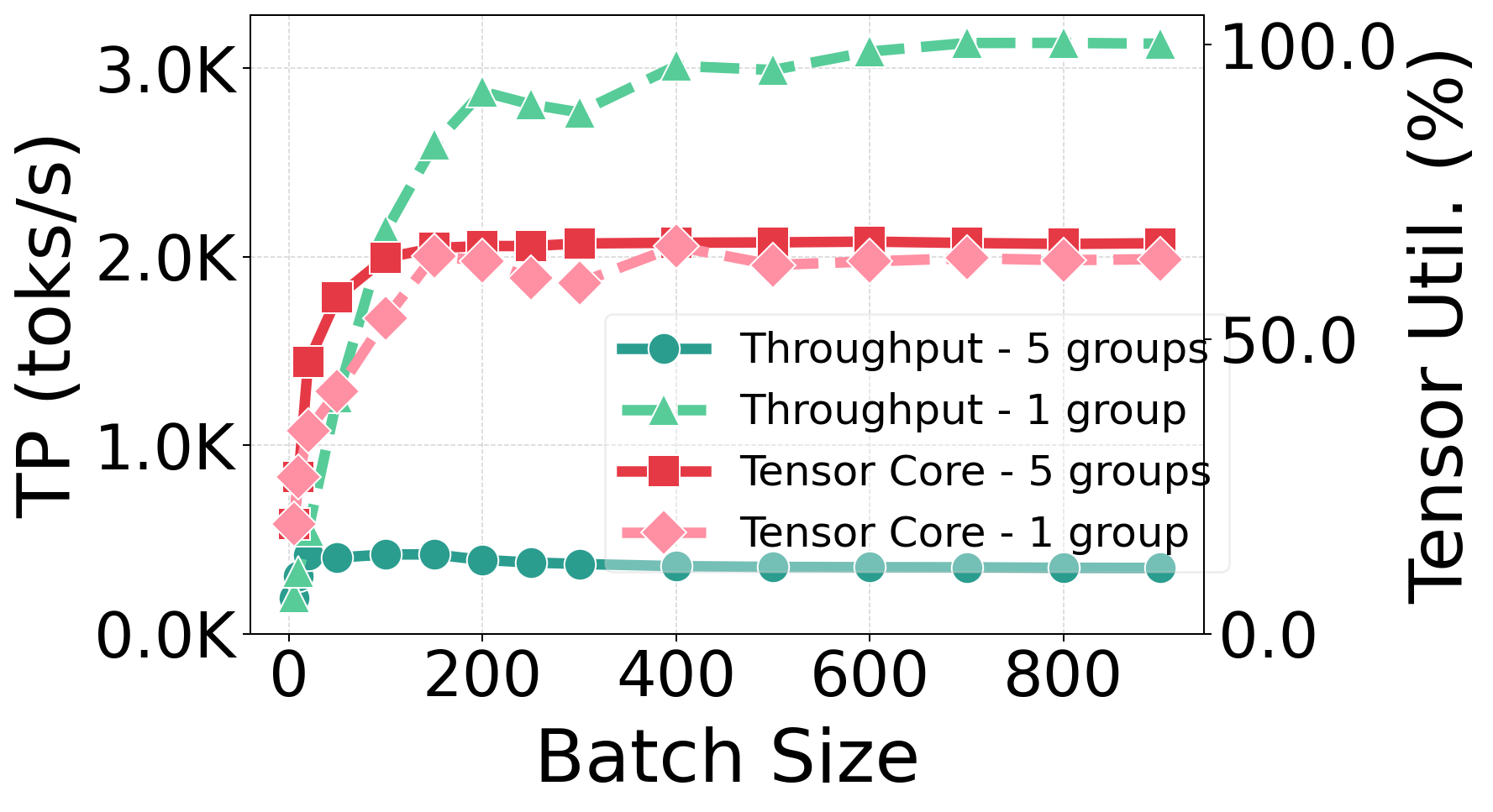}
    \captionof{figure}{Throughput vs. Batch Size}
    \label{fig:batch_size_tp_motivation}
\end{minipage}
\hfill
\begin{minipage}{0.34\textwidth}
    \centering
    \includegraphics[width=\columnwidth]{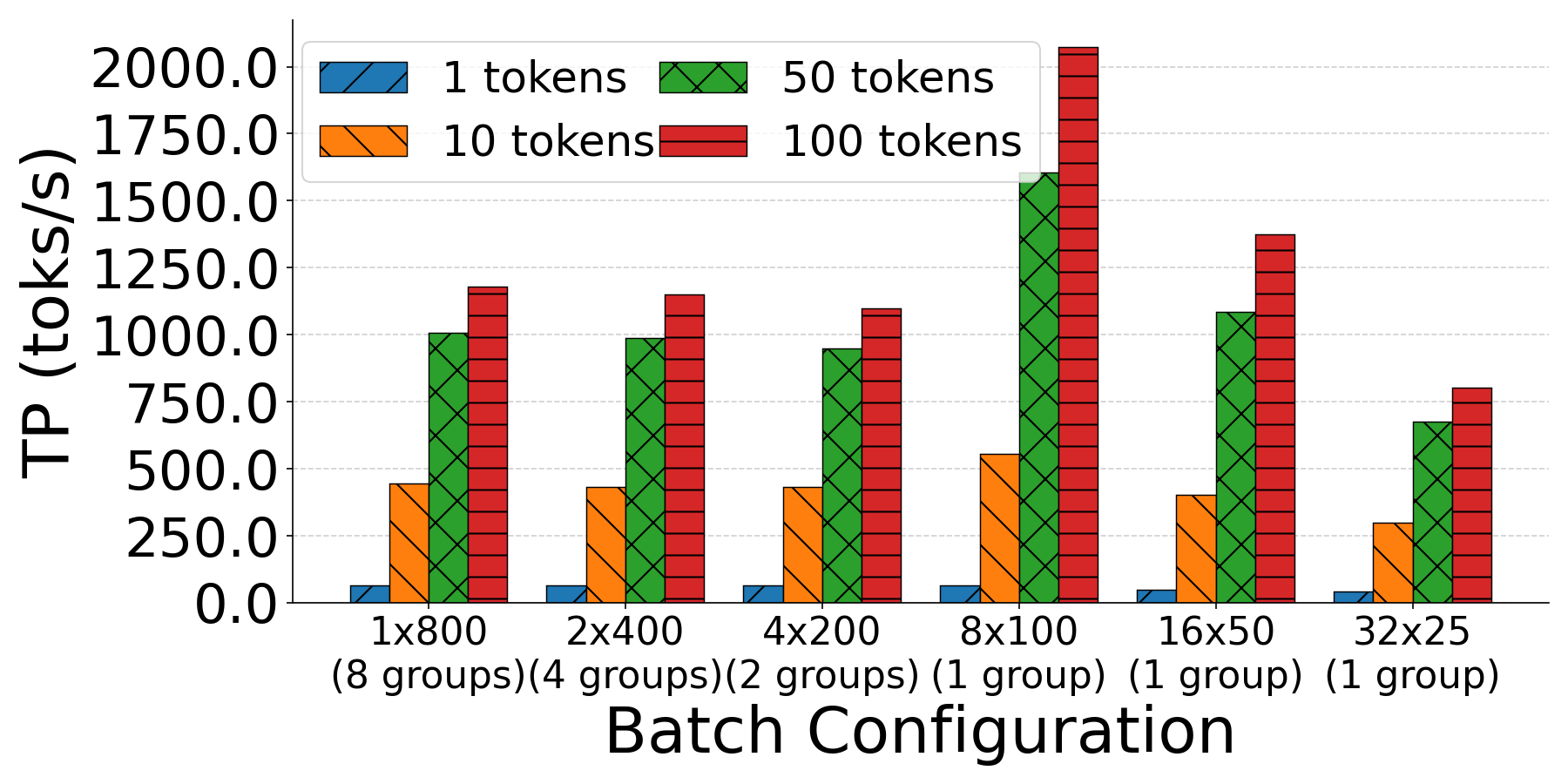}
    \captionof{figure}{Batch Size vs. Homogeneity}
    \label{fig:prefill_decode_tp_batch_config_decode}
\end{minipage}

\end{figure*}

\textbf{Experiment 4:} Until now, all the experiments have considered requests that have a single level of prefix sharing. We now ask the question: what is the impact of more complicated sharing patterns on performance? For this, we construct radix trees with 1 to 4 levels of shared prefixes (Figure~\ref{fig:sharing_configurations}). There are two branches in each level, and requests are equally distributed across the leaves. The sequence length per request is fixed at $4L$, and Figure~\ref{fig:tp_vs_l_1_9_11} shows the decode throughput for three values of $L$. As expected, a single level of radix tree has the highest performance because of maximum locality. Interestingly, radix trees with 2, 3, and 4 levels exhibit similar throughput across different values of $L$, despite differing branching structures. Why so? During the initial segment of length $L$, all configurations access the same KV region. However, once branching occurs, the batch becomes heterogeneous, locality degrades, and the marginal cost of additional prefix groups is small, as seen earlier in Figure~\ref{fig:hbm_evic_tp_bs}.

\textbf{Takeaway 4:} \textit{Performance is driven mainly by the length of the top-level prefix shared by all requests, while the number of radix-tree levels and branches has a negligible impact.}

\subsection{Batch Size vs Prefix Homogeneity}
\label{subsec:batch_size_vs_prefix_homogeneity}
So far, we have established that prefix homogeneity has a significant impact on performance. But when the workload does not have enough requests from a prefix group and prefix-homogeneous batches are too small, the compute will remain under-utilized, leading to a loss in throughput. Therefore, there exists a trade-off between batch size and prefix homogeneity. We now explore this trade-off.

\textbf{Experiment 5:} We study decode throughput vs. batch size using two workloads: (i) heterogeneous, with requests spread across 5 distinct 10K-token prefixes, and (ii) homogeneous, with all requests sharing one such prefix. Figure~\ref{fig:batch_size_tp_motivation} shows throughput and tensor core utilization as a function of batch size for both workloads. We see that throughput plateaus at a point before the maximum batch size, and the saturation throughput is higher in the homogeneous case due to greater locality. Further, the batch size at which throughput saturates is higher for the homogeneous workload because requests can access the same KV cache without using additional memory bandwidth. Interestingly, the compute utilization is almost exactly the same across both workloads. Similar observations have also been made in prior work~\cite{recasens2025mindmemorygapunveiling}.

\textbf{Takeaway 5:} \textit{Decode is memory bandwidth bound and increasing batch size beyond a point does not necessarily improve throughput. However, the point at which diminishing returns begin depends on prefix homogeneity (and hardware config).}

\textbf{Experiment 6:} Given that batch size has a diminishing impact on performance, we ask if moderately small homogeneous batches can outperform larger heterogeneous ones. To answer this, we use eight prefix groups with 100 requests each (800 total, each 10K tokens) and evaluate the following batching strategies: (i) 1$\times$800 (8 prefix groups/batch); (ii) 2$\times$400 (4 prefix groups/batch); (iii) 4$\times$200 (2 prefix groups/ batch); (iv) 8$\times$100 (homogeneous); (v) 16$\times$50 (homogeneous); and (vi) 32$\times$25 (homogeneous). We vary the decode length from 1 to 100 tokens and measure the end-to-end throughput, including both the prefill and decode phases for all batches. Figure~\ref{fig:prefill_decode_tp_batch_config_decode} shows this throughput across the different batching strategies. For small decode lengths, throughput is dominated by prefill, and large heterogeneous batches perform the best due to higher tensor core utilization; splitting 1$\times$800 into smaller batches slightly reduces throughput due to lower compute efficiency, as locality benefits are not yet dominant. As decode length increases, performance trends reverse: multiple decode iterations repeatedly sweep the KV cache and reap greater benefits due to locality. Therefore, at longer decode lengths, fully homogeneous batches (8$\times$100) significantly improve throughput despite smaller batch sizes. Notably, even 16$\times$50 continues to outperform 1$\times$800 at large decode lengths, showing that locality gains can outweigh reduced tensor core utilization. Only when batches become too small (32$\times$25) does throughput decline again, as reduced arithmetic intensity and occupancy dominate.
 
\textbf{Takeaway 6:} \textit{Moderately small prefix-homogeneous batches can achieve higher throughput than larger heterogeneous ones. Gains increase with longer decode lengths as repeated KV sweeps amplify locality; however, excessively small batches underutilize compute and degrade performance.}

\begin{table}[t]
\centering
\small
\resizebox{\columnwidth}{!}{\begin{tabular}{lcccc}
\toprule
\textbf{Policy} & \textbf{GPU Exec (s)} & \textbf{Batch Time (s, \%)} & \textbf{Throughput (toks/s)} \\
\midrule
vLLM FCFS & 229.36 & 3e-6 (0.00\%) & 436.25  \\
SGLANG DFS-W & 177.78 & 92.15 (51.8\%) & 562.61 \\
SGLANG LPM & 149.27 & 46.62 (31.2\%) & 669.92  \\
PAT (vLLM Block Table) & 316.45 & 91.72 (29.0\%) & 316.63  \\
vLLM RadixTree & 139.33 & 62.95 (45.2\%) & 717.35  \\
vLLM User IDs & 105.43 & 0.03 (0.03\%) & 948.54 \\
\bottomrule
\end{tabular}}
\caption{Overhead of Dynamic Prefix Detection}
\label{tab:cost_schedulers}

\end{table}

\subsection{Overhead of Dynamic Prefix Detection}
\label{subsec:overhead_dynamic_prefix_detection}
Next, we study how state-of-the-art inference engines identify prefix sharing and the costs involved. Systems such as SGLang \cite{zheng2024sglangefficientexecutionstructured} and BatchLLM \cite{zheng2025batchllmoptimizinglargebatched} do not explicitly form prefix-homogeneous batches, but use radix trees to detect sharing, and schedule requests to minimize KV recomputation. 

\textbf{Experiment 7:} We construct a workload of 2K requests, each with a 10K-token prefix, and compare scheduling CPU time to total execution time. SGLang proposes two prefix-aware policies that use its radix tree: \textit{Longest Prefix Match}, which prioritizes requests with the longest matching cached prefixes, and \textit{DFS-Weight}, which performs a depth-first traversal and prioritizes branches with more active requests. vLLM uses a much simpler hash-based block table, which is used for grouping requests in the prefix-aware attention kernel implementation of PAT~\cite{yi2026pat}. We compare the overheads of this block table, along with our own radix-tree-based scheduler implemented on top of vLLM. We also compare against an idealized policy where prefix group identifiers are provided explicitly, eliminating dynamic prefix discovery.

Table~\ref{tab:cost_schedulers} shows that SGLang's LPM and DFS-Weight incur scheduler overhead (measured as time taken by all functions invoked during batch formation) comparable to GPU execution time, even though these policies improve performance over vLLM's FCFS. In contrast, when prefix group identifiers are provided explicitly, CPU overhead drops significantly. This gap arises from the structure of radix trees: they require token-by-token comparisons during insertion, dynamic node splitting on partial matches, and repeated tree traversals for scheduling requests. With long prompts and high concurrency, these operations incur significant overhead. Similarly, vLLM's block table design is inefficient in this regard. However, these overheads are not sufficiently highlighted in prior work~\cite{zheng2025batchllmoptimizinglargebatched,zhao2024blendserveoptimizingofflineinference}, which is mostly based on offline inference.

\textbf{Takeaway 7:} \textit{Existing approaches introduce significant CPU overhead when detecting prefix sharing, which is sometimes even comparable to the GPU execution time.}

\section{Design and Implementation}
\label{sec:design}
Our experiments so far have demonstrated that the length of the prefix shared across \textit{all} active requests is an important determinant of system performance. However, this metric interacts with batch size in a complex way: larger batches improve GPU utilization but often reduce shared prefix alignment. Further, detecting this shared prefix length incurs high overhead in existing systems. We now describe the design of \Feather, an efficient scheduler that finds the right balance between batch size and prefix homogeneity.

\subsection{Overview}
\label{sec:system_overview}
\Feather is based on the following two key ideas.

\subsubsection*{Key Idea 1: Learn the Stopping Decision via RL (\S\ref{subsec:batching_rl})}
The trade-off between batch size and prefix homogeneity raises a key question: when should the scheduler stop adding requests? The answer depends on both hardware and workload, making fixed heuristics brittle. We model batch construction as a sequential decision process and learn a stopping policy via Reinforcement Learning. At each step, the scheduler evaluates a candidate request using various state features and decides whether it should be added to the batch or if the batch should be dispatched to the GPU for execution.

\subsubsection*{Key Idea 2: Shared Prefix Detection via Chunked Hash Tree (\S\ref{subsec:chunked_hash_tree})}
Tree-based approaches incur significant overhead (\S\ref{subsec:overhead_dynamic_prefix_detection}), and finding the best waiting request requires a DFS traversal that is $\mathcal{O}(E)$, where $E$ is the number of edges in the radix tree. In the worst case, this can go up to $\mathcal{O}(W\times T)$, where $W$ is the number of waiting requests, and $T$ is the sequence length. We instead represent each request as a sequence of hashes over fixed-size chunks and maintain a working set of hashes for the active batch. This is similar to Merkle-style hash chains but does not require sequential hash computation, and each chunk can be processed independently without prefix dependencies. Prefix matching is thus reduced to a set-overlap problem: selecting the request with the maximum overlap with the working set. However, a naive implementation that scans all requests and counts overlaps requires $\mathcal{O}(W \cdot C)$ work per decision, where $C$ is the number of chunks in a request, which remains too expensive for frequent scheduling. We avoid this by maintaining a missing count per request—the number of chunks absent from the working set—which changes only when the working set changes (i.e., upon request addition or completion). Overlaps are thus updated incrementally, reducing selection to finding the request with the smallest missing count. To do so efficiently, we maintain a min-heap over missing counts, with lazy updates, yielding $\mathcal{O}(\log W)$ selection cost.

\subsubsection*{System Architecture}
\begin{figure}[t]
\centering
\includegraphics[width=\columnwidth]{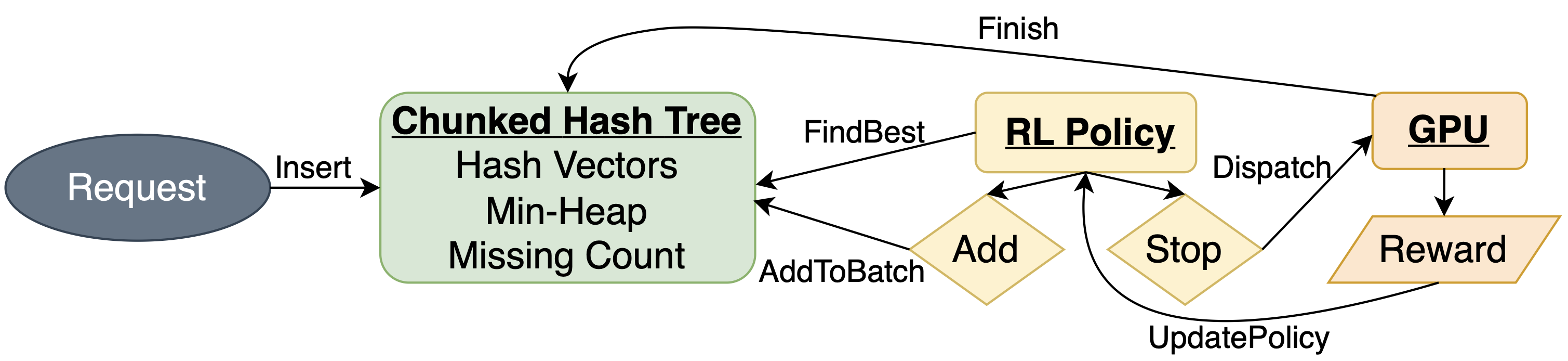}
\caption{\Feather Pipeline}
\label{fig:pipeline}

\end{figure}

Figure~\ref{fig:pipeline} shows the overall scheduling pipeline. We consider an inference server that continuously receives requests, which are inserted into the waiting queue of CHT. The scheduler builds batches incrementally in a loop. It first finds the best candidate by retrieving the waiting request with the highest overlap with the current working set. Then the RL policy evaluates the current system state and decides whether to \textsc{Add} the request or \textsc{Stop}. If \textsc{Add} is chosen, the request is added to the scheduler batch, and all the data structures, such as the working set and the min-heap, are updated accordingly; otherwise, the batch is dispatched to the GPU for an execution step. The decode throughput measured after the GPU execution is used as a reward for the RL policy. Completed requests are removed from the active batch, and the working set is updated.

\begin{figure*}[t]
\captionsetup[subfigure]{labelformat=simple}
    \renewcommand\thesubfigure{(\alph{subfigure})}
\centering
\begin{subfigure}{0.24\textwidth}
    \centering
    \includegraphics[width=\linewidth]{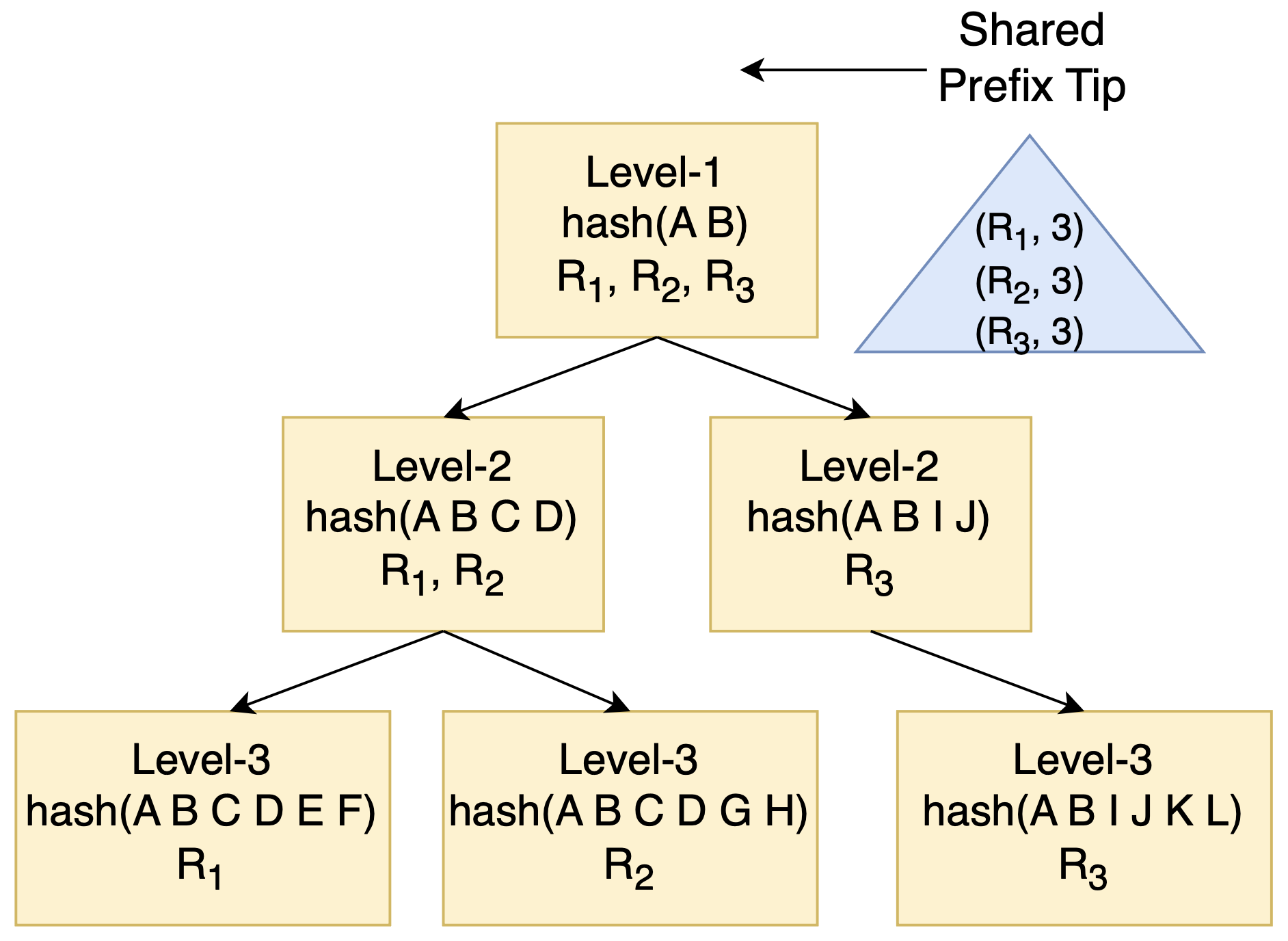}
    \caption{Insertion}
    \label{fig:chunked_hash_tree_insertion}
\end{subfigure}
\hfill
\begin{subfigure}{0.28\textwidth}
    \centering
    \includegraphics[width=\linewidth]{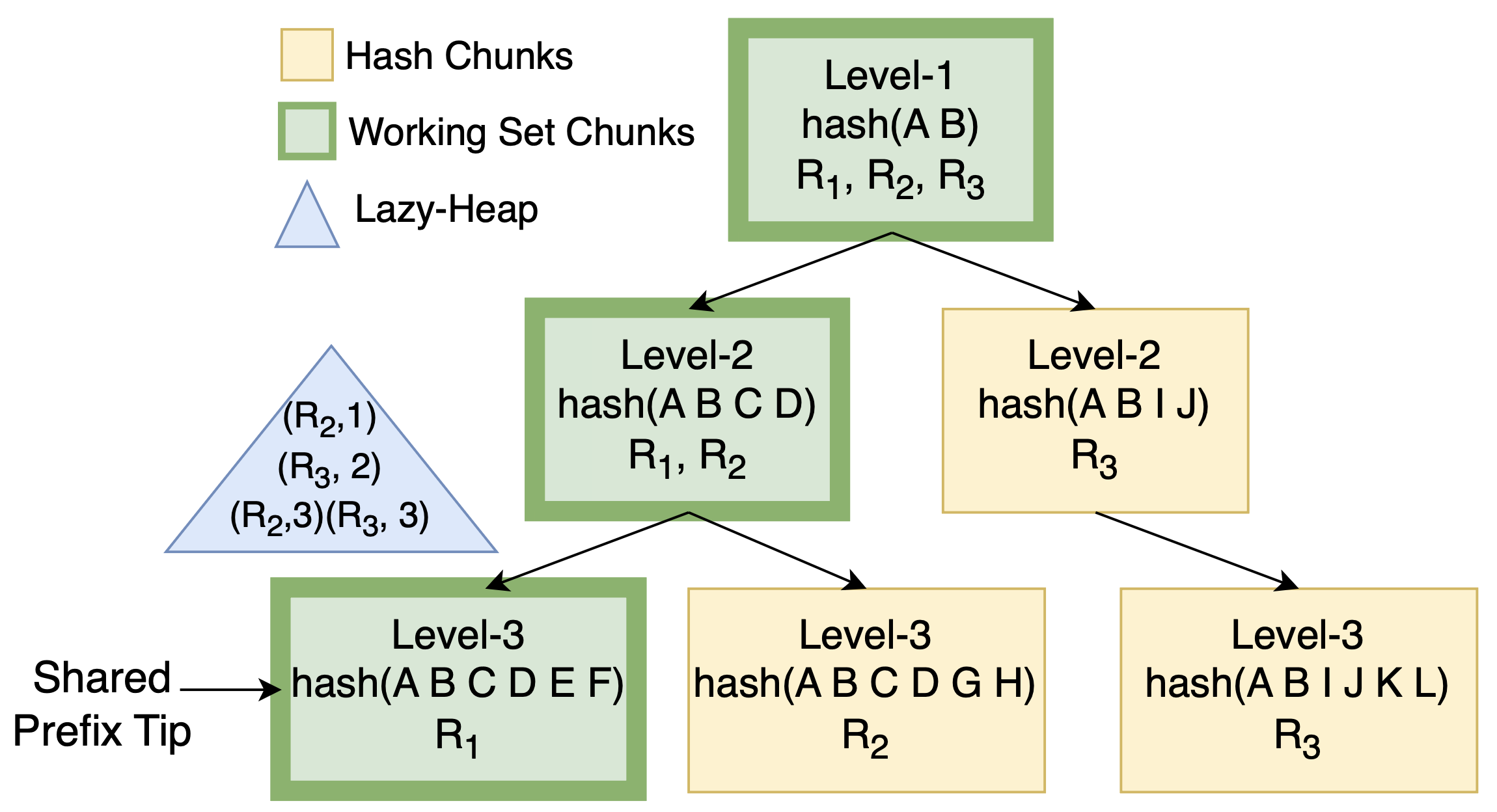}
    \caption{Adding to Batch}
    \label{fig:chunked_hash_tree_activation}
\end{subfigure}
\hfill
\begin{subfigure}{0.24\textwidth}
    \centering
    \includegraphics[width=\linewidth]{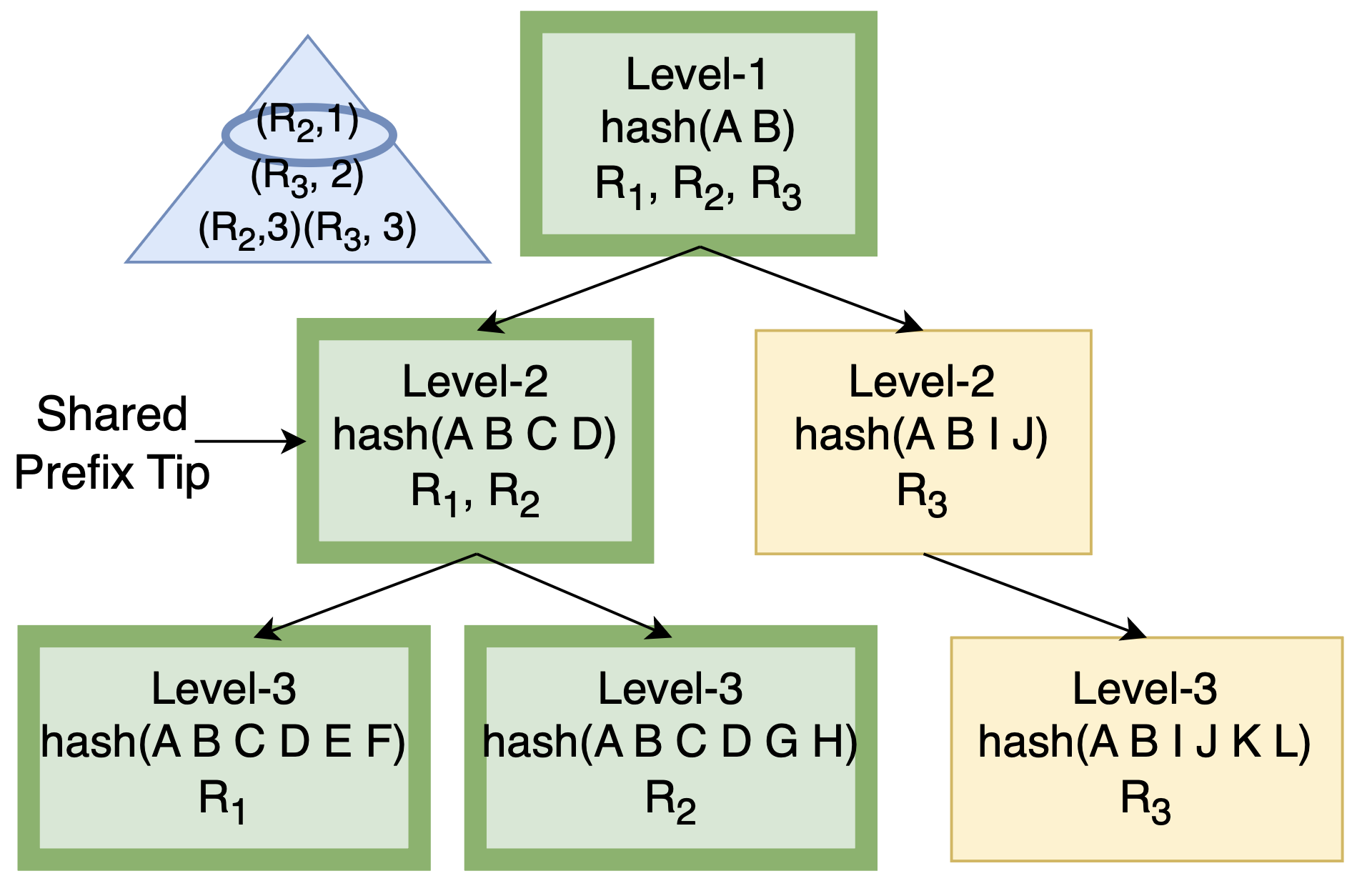}
    \caption{Finding the Best Request}
    \label{fig:chunked_hash_tree_find_best_request}
\end{subfigure}
\hfill
\begin{subfigure}{0.21\textwidth}
    \centering
    \includegraphics[width=\linewidth]{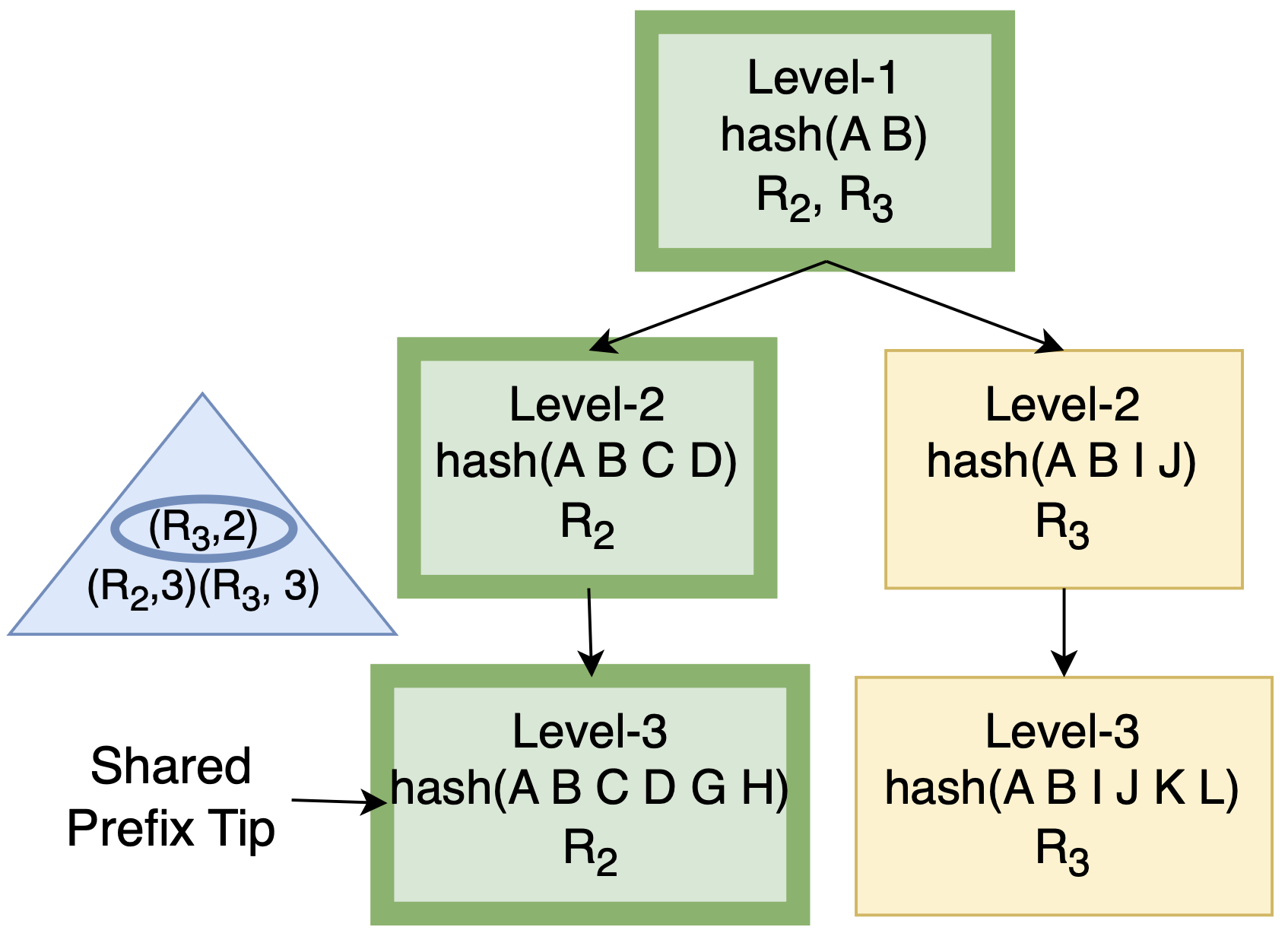}
    \caption{Request Completion}
    \label{fig:chunked_hash_tree_finish}
\end{subfigure}
\caption{Chunked Hash Tree Operations}
\label{fig:chunked_hash_tree}

\end{figure*}

\subsection{Chunked Hash Tree (CHT)}
\label{subsec:chunked_hash_tree}
We now describe the operations of CHT (a more formal treatment is in \S\ref{sec:algorithms}). We begin with how token sequences are mapped to a compact representation in our system.

\subsubsection*{Hash Vectors}
Given a request as a token sequence, we partition it into non-overlapping chunks of size $K$ and build a sequence of hashes, one for each chunk, where each hash is cumulative over all tokens up to that chunk. For example, in Figure~\ref{fig:chunked_hash_tree_insertion}, a request with tokens A-F and $K = 2$ is represented as three incremental hashes: hash(A-B), hash(A-D), and hash(A-F). Two requests will have the same hash at level $l$ iff they share the same first $lK$ tokens. Each request is represented as an ordered hash vector. We use an incremental xxHash-64 state, ensuring each token is processed once, yielding a compute cost of $\mathcal{O}(T)$, independent of the number of chunk boundaries, where $T$ is the sequence length.

\subsubsection*{Insertion}
\textsc{Insert} is invoked when a new request arrives at the scheduler. We first compute the cumulative prefix-hash vector and store it in a map from request IDs to hash vectors for faster reuse. We also maintain a reverse index that maps each hash to the set of requests containing that hash. For each hash in the hash vector, we insert the request into this reverse index. The overall cost is dominated by the hash computation phase, which runs in $\mathcal{O}(T)$.

\subsubsection*{Working Set, Min-Heap and Shared Prefix Tip}
To evaluate how well a request matches the active batch, we maintain a working set that contains (level, hash) tuples representing prefix hashes at that level covered by currently active requests. For each request, we also maintain a missing count, which is the number of levels for which this request's hashes are not present in the working set. We insert these missing counts into a min-heap, which enables efficient retrieval of the most suitable request for addition to the batch. \textsc{Insert} updates both of these structures. Moreover, we do not perform in-place updates and use lazy deletion to discard stale entries in the heap. To relate to our key metric, which is the number of chunks shared by all active requests, we also maintain a shared prefix tip, storing the depth and hash of the deepest node that is shared by all the active requests. Figure~\ref{fig:chunked_hash_tree_activation} shows the Chunked Hash Tree after the request $R_1$ is added to the batch. The nodes marked in green denote the working set. Since there is just a single request, the shared prefix tip is at Level-3 of $R_1$. The heap contains the missing counts of the waiting requests along with stale entries.

\subsubsection*{Finding the Best Request}
The goal of \textsc{FindBest} is to select the waiting request that best complements the active batch, i.e., the one with the minimum missing count, so that we maximize the depth of the shared prefix tip\footnote{An astute reader might ask why we are choosing a request based on its difference from the working set rather than its impact on the length of the shared prefix tip. A brief explanation is that we want to pick a request that helps us revert back to a longer shared prefix tip sooner, and a request that differs from the working set by the least number of chunks helps us achieve this goal best. We refer the reader to Appendix~\ref{subsec:find_best_reques_alternative_heuristic} for more details.}. Since the min-heap is keyed by this count, the best candidate appears at the top after skipping stale entries. As shown in Figure~\ref{fig:chunked_hash_tree_find_best_request}, $R_2$ appears at the top of the min-heap and is therefore selected as the best request. Note that adding $R_2$ to the batch reduces the shared prefix tip by one level. \textsc{FindBest} runs in $\mathcal{O}(\log W)$, where $W$ is the number of waiting requests. For the selected request, we compute two metrics that are inputs to the RL policy (\S\ref{subsec:batching_rl}), which needs to decide whether adding the request to the current batch is beneficial for the overall throughput. One of the metrics is the change in the depth of the shared prefix tip, and the other is the number of additional waiting requests that are present at this new depth, which can all be added to the batch without any further impact on the shared prefix tip in the future.

\subsubsection*{Adding a Request to the Active Batch} 
When a request is promoted to the active batch, we update the working set to include its prefixes and recompute the shared prefix tip. We iterate over the hash vector backwards; for each level, if the corresponding prefix hash isn't present in the working set, it is inserted, and all waiting requests sharing it have their missing count decremented. For the shared prefix tip, admitting a new request can only \textit{shorten} the common prefix, so we seek the deepest level where it agrees with the existing tip hash. This corresponds to the least common ancestor of all active request leaves in the hash trie. \textsc{AddToBatch} runs in $\mathcal{O}(C \cdot W_h \cdot \log W)$ time, where $C = \lceil T/K \rceil$ is the number of chunks for a request with $T$ tokens, and $W_h$ is the number of waiting requests that share the same hash ($W_h \ll W$).

\subsubsection*{Removing a Request from the Active Batch}
\textsc{Finish} is called when a request completes, and we remove it from the active batch. It is the symmetric counterpart of \textsc{AddToBatch}. We iterate over the hash vector from the start. A hash unused by any active request is removed from the working set, and all waiting requests containing it have their missing count incremented. A request completion can only \textit{lengthen} the shared prefix tip since the remaining active requests may agree on a longer common prefix. Starting from the tip's previous position, we scan forward, checking whether all remaining active requests share the same hash at each level. (We do this efficiently by maintaining reference counts of each hash.) The extension stops at the first level of disagreement or at the shortest active request. This scan is inexpensive in practice, as the tip rarely advances more than a few levels. \textsc{Finish} also runs in $\mathcal{O}(C \cdot W_h \cdot \log W)$ time. Figure~\ref{fig:chunked_hash_tree_finish} shows the state after $R_1$ completes: the tip extends to $R_2$'s final level, and waiting heap costs are updated accordingly.

\subsubsection*{Optimizations}
We employ several optimizations to improve efficiency. First, \textsc{FindBest} may be invoked multiple times within a scheduling round without state changes, so we cache its last result and reuse it until it is invalidated by any mutation (e.g., \textsc{AddToBatch} and \textsc{Finish}). Second, we use lazy heap updates: when a request's missing count changes, we push a new entry instead of updating it in place and skip stale entries during \textsc{FindBest}. Third, we maintain the shared prefix tip incrementally rather than recomputing it. This drops the cost from $\mathcal{O}(A \cdot C)$, where $A$ is the number of active requests, and $C$ is the number of chunks, to just $\mathcal{O}(C)$ in the worst case and typically terminates earlier. Finally, all major data structures are pre-allocated at initialization to eliminate rehashing and reallocation overhead.

\subsection{Batching and Reinforcement Learning}
\label{subsec:batching_rl}
The CHT efficiently identifies the next best request to add, maximizing shared prefix length, but it does not determine \textit{when to stop}. While adding requests improves GPU utilization, it can also shrink the shared prefix and degrade locality. We frame this stop/continue decision as a learning problem and present three policies: a heuristic, a contextual bandit, and a Q-learning agent. The scheduler incrementally builds a batch by selecting candidates via the CHT and deciding whether to admit them or dispatch the batch.

\subsubsection*{State, Actions, and Reward}
At each step of batch formation, the scheduler observes the state $s = (\mathit{b},\; \Delta,\; \mathit{w})$ and takes an action $a \in \{\textsc{Add}, \textsc{Stop}\}$. Here, $b$ is the current batch size (a proxy for GPU utilization), $\Delta$ is the loss in the shared prefix, and $w$ is the number of waiting requests sharing the prefix up to the new depth, indicating whether the locality cost of this addition will be compensated by future activations. Both RL policies optimize the reward $\mathcal{R} = \text{throughput}_{\mathrm{decode}}$. Maximizing this encourages large batches and short decode times (better prefix locality). The state is discretized into $(\hat{b}, \hat{\Delta}, \hat{w})$, where $b$ and $w$ use exponential bins, and $\Delta$ uses coarse thresholds. The resulting table is small, hashable, and converges quickly under moderate load.

We choose decode throughput as the reward because it is effectively the inverse of time between tokens (TBT), while remaining a safer objective than minimizing TBT directly --- a batch of size one trivially minimizes per-token latency without regard for GPU utilization, so a raw TBT-minimizing policy would degenerate toward starving the batch. Time to first token (TTFT) is also a weaker signal for this decision, as it is dominated by prefill compute and queuing delay rather than by the decode batch composition choice the policy makes at each step. Finally, decode throughput implicitly captures hardware and model-specific effects --- such as attention cost scaling and memory bandwidth limits --- that are not exposed as explicit state features, allowing the same reward formulation to generalize across the configurations in \S\ref{sec:eval} without redesigning the state space.

\subsubsection*{Heuristic Policy}
As a baseline, we encode domain intuition as simple rules balancing two goals: maximizing GPU utilization (large batches) and preserving prefix locality (avoiding large drops in the shared chain). The heuristic always accepts candidates with zero chunk loss (free additions). When the batch is small, it tolerates higher loss to avoid underutilization. Otherwise, it admits a candidate only if the chunk loss is within a fixed threshold, with slightly relaxed tolerance when many waiting requests share the resulting prefix (anticipating amortization). If none of these conditions hold, it stops and dispatches the batch. This policy requires no training and converges immediately, but its fixed thresholds cannot adapt to workload or hardware changes.

\subsubsection*{Contextual Bandit}
To avoid hand-tuned thresholds, we use a contextual bandit with Upper Confidence Bound (UCB) selection~\cite{auer2002finite}. Each batch-construction step is treated as an independent decision: given the discretized state $\hat{s} = (\hat{b}, \hat{\Delta}, \hat{w})$, the policy selects the action that maximizes

\[
  \mathrm{UCB}(\hat{s},a) = \frac{\mu(\hat{s},a)}{n(\hat{s},a)} + c\sqrt{\frac{\ln S}{n(\hat{s},a)}}.
\]

Here, $\mu(\hat{s}, a)$ is the cumulative reward, $n(\hat{s}, a)$ is the number of times action $a$ was taken in $\hat{s}$, $S$ is the total number of decisions, and $c$ is the exploration coefficient. The first term favors high-reward actions, while the second promotes exploration of under-sampled ones.

\subsubsection*{Q-Learning Policy}
As an alternative to the previous policy, we can also model batch construction as an episodic reinforcement learning problem, where each episode begins when forming a new batch and ends upon \textsc{Stop} or when the waiting queue is exhausted. The Q-table~\cite{watkins1992} stores values $Q(s, a)$ for each (discretized state, action) pair and is updated after each episode using:

\[
Q(s, a) \; \leftarrow\; Q(s, a) + \alpha\bigl[r + \gamma \max_{a'} Q(s', a') - Q(s, a)\bigr].
\]

Here, $r$ is the observed throughput, $s'$ is the next state, $\alpha$ the learning rate, and $\gamma$ the discount factor, which biases the agent toward near-term throughput, consistent with real-time inference. We follow an $\varepsilon$-greedy exploration strategy: at each step, a random action is selected with probability $\varepsilon$, and the greedy action $\arg\max_a Q(s,a)$ otherwise. During training, $\varepsilon$ decays from $1.0$ to $\varepsilon_{\min}$, transitioning from full exploration to near-deterministic behavior. Qualitatively, we expect this policy to perform better in the long term when actions influence future batch composition.

At each step, \textsc{FindBest} returns the lowest-cost candidate along with $(\Delta, w)$, forming the state $s$. The policy (heuristic, bandit, or Q-learning) then selects \textsc{Add} or \textsc{Stop}. If \textsc{Add}, the request is activated and appended to the batch; otherwise, the batch is dispatched. After execution, the observed decode throughput is used as the reward: the bandit updates its statistics and the Q-learning agent applies the Bellman update. This has also been shown in Figure~\ref{fig:pipeline}.

\subsection{Integration with vLLM and SGLang}
We integrate our scheduler as a lightweight layer on top of both vLLM~\cite{kwon2023efficient} and SGLang~\cite{zheng2024sglangefficientexecutionstructured}. The Chunked Hash Tree (implemented in C++) along with shared prefix chain tracking and reinforcement learning policies operates entirely at the scheduler level. Our code is available at \textcolor{blue}{\url{https://github.com/networkedsystemsIITB/Feather}}.

\begin{figure*}[t]
\captionsetup[subfigure]{labelformat=simple}
    \renewcommand\thesubfigure{(\alph{subfigure})}
\centering
\begin{subfigure}{0.32\textwidth}
    \centering
\includegraphics[width=\linewidth]{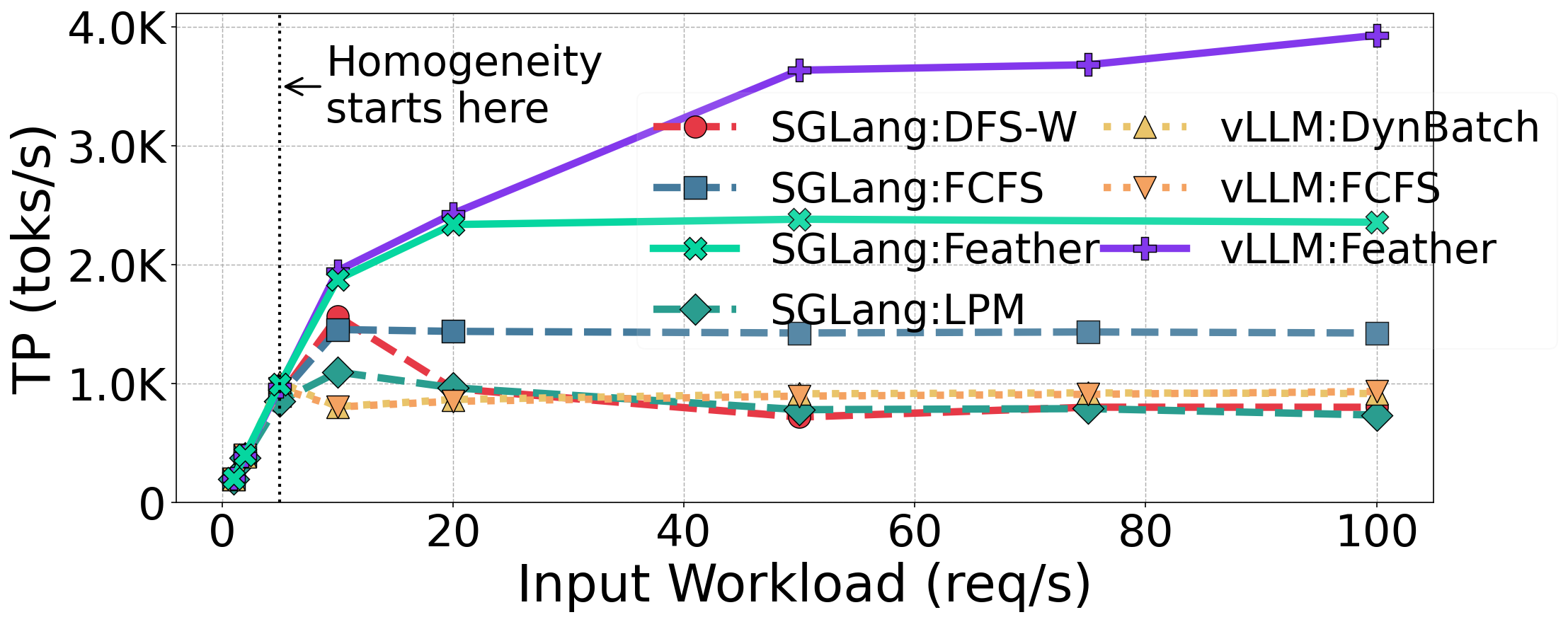}
\caption{Throughput}
\label{fig:tp_vs_rps_l_5000_fam_5}
\end{subfigure}
\hfill
\begin{subfigure}{0.32\textwidth}
    \centering
\includegraphics[width=\linewidth]{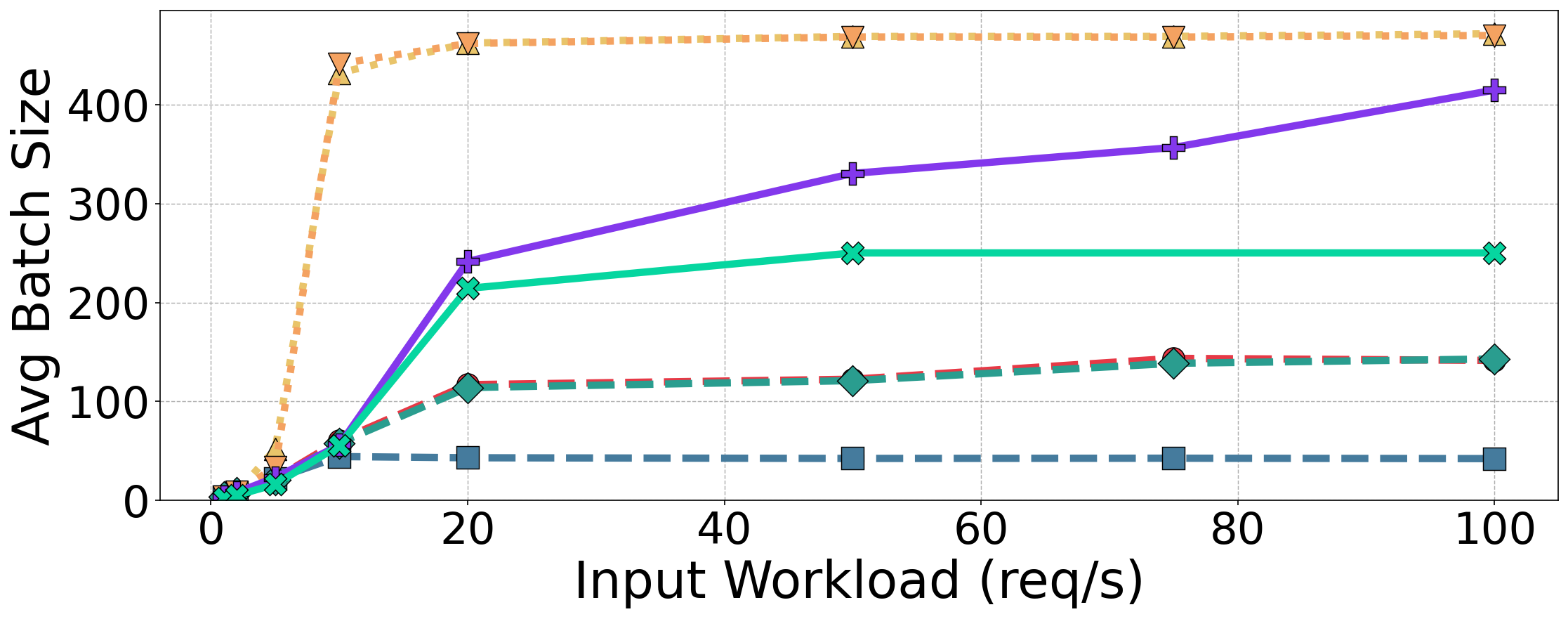}
\caption{Average Batch Size}
\label{fig:abs_vs_rps_l_5000_fam_5}
\end{subfigure}
\hfill
\begin{subfigure}{0.32\textwidth}
    \centering
\includegraphics[width=\linewidth]{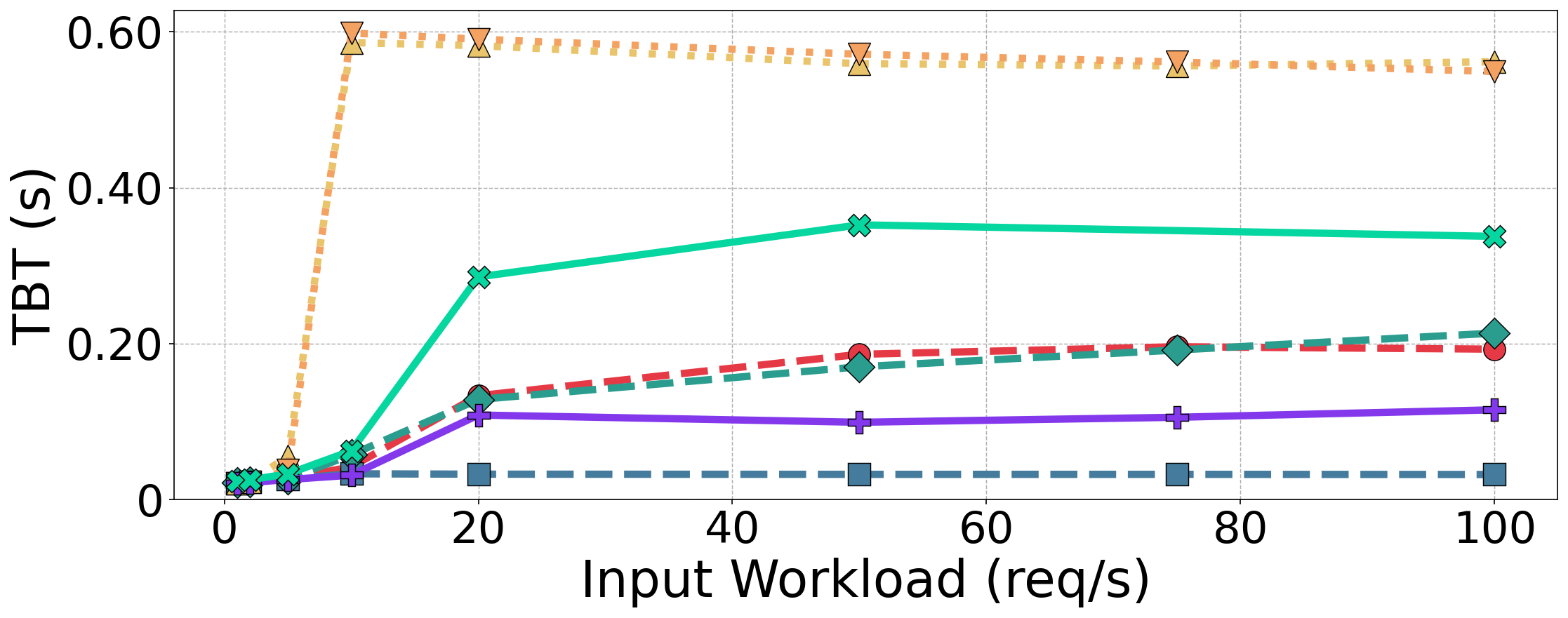}
\caption{Time between tokens (TBT)}
\label{fig:tbt_vs_rps_l_5000_fam_5}
\end{subfigure}
\caption{Poisson Workload}
\label{fig:input_workload_poisson_rates}

\end{figure*}

\section{Evaluation}
\label{sec:eval}

We evaluate \Feather across diverse models, GPU configurations, and real-world datasets and scenarios. Our evaluation answers the following questions: (i) Under what workloads do smaller homogeneous batches outperform larger heterogeneous ones? (ii) How effective is \Feather's Chunked Hash Tree compared to traditional radix-tree-based approaches? (iii) How do different design components of \Feather (RL, CHT) interact to produce end-to-end performance gains?

\subsection{Experimental Setup}
\subsubsection*{Models and Hardware}
We evaluate three model families across multiple sizes: Qwen 0.5B, 1.5B, 8B~\cite{qwen}, LLaMA 3 8B~\cite{llama3modelcard}, and LongChat 13B~\cite{longchat2023}. Experiments are run on an NVIDIA RTX 6000 Ada GPU~\cite{nvidia_rtx6000_datasheet} with 48 GB GDDR6, 96 MB L2, and the maximum batch size is fixed at 500, which fits within the GPU memory for most settings. We additionally evaluate on an A100-80GB GPU for comparisons involving PAT~\cite{yi2026pat}, as it is specifically optimized for this hardware.

\begin{figure}[t]
\captionsetup[subfigure]{labelformat=simple}
    \renewcommand\thesubfigure{(\alph{subfigure})}
\centering
\begin{subfigure}{0.49\columnwidth}
    \centering
    \includegraphics[width=\linewidth]{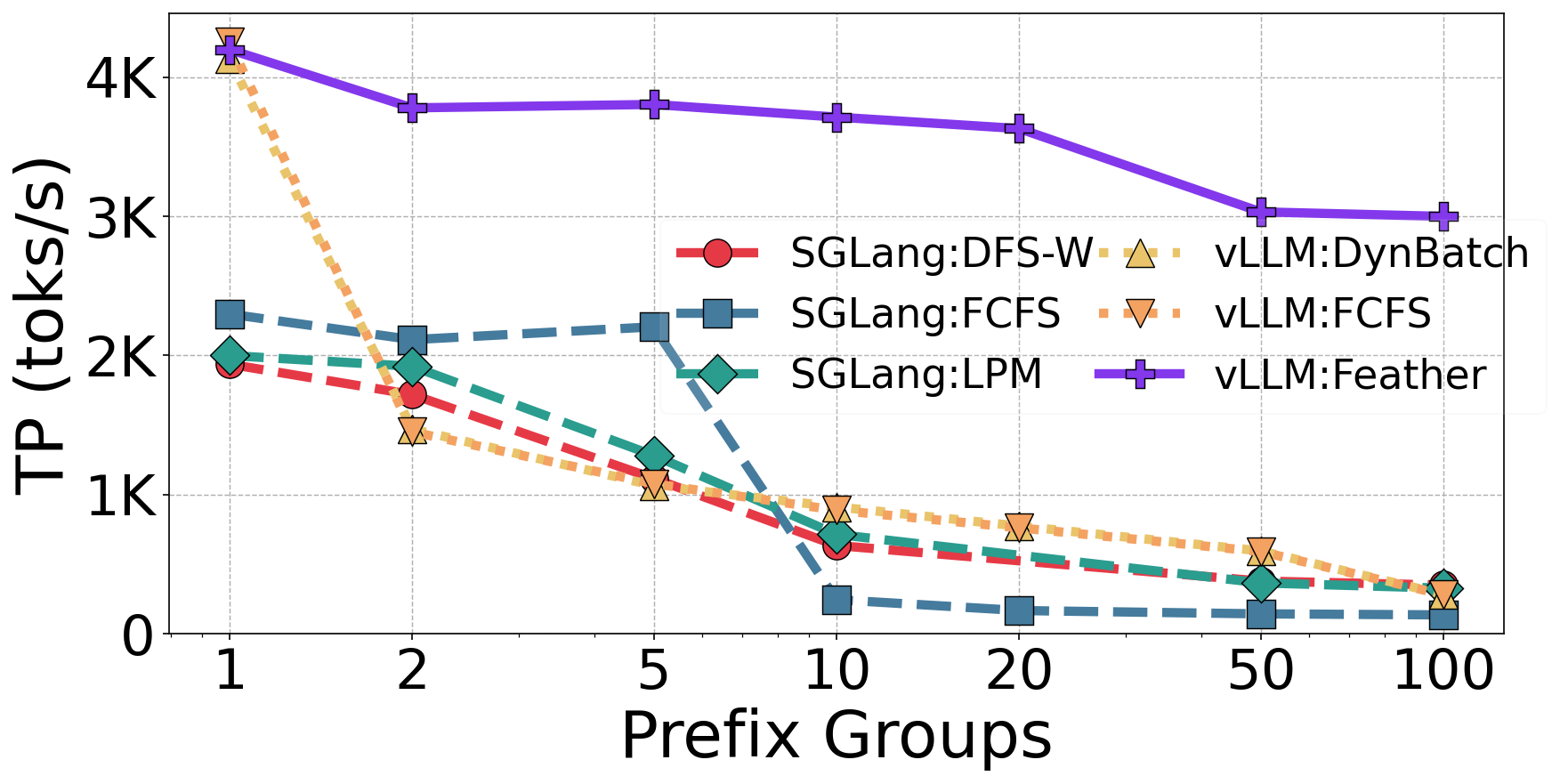}
    \caption{Throughput}
    \label{fig:tp_vs_fam_l_5000}
\end{subfigure}
\hfill
\begin{subfigure}{0.49\columnwidth}
    \centering
    \includegraphics[width=\linewidth]{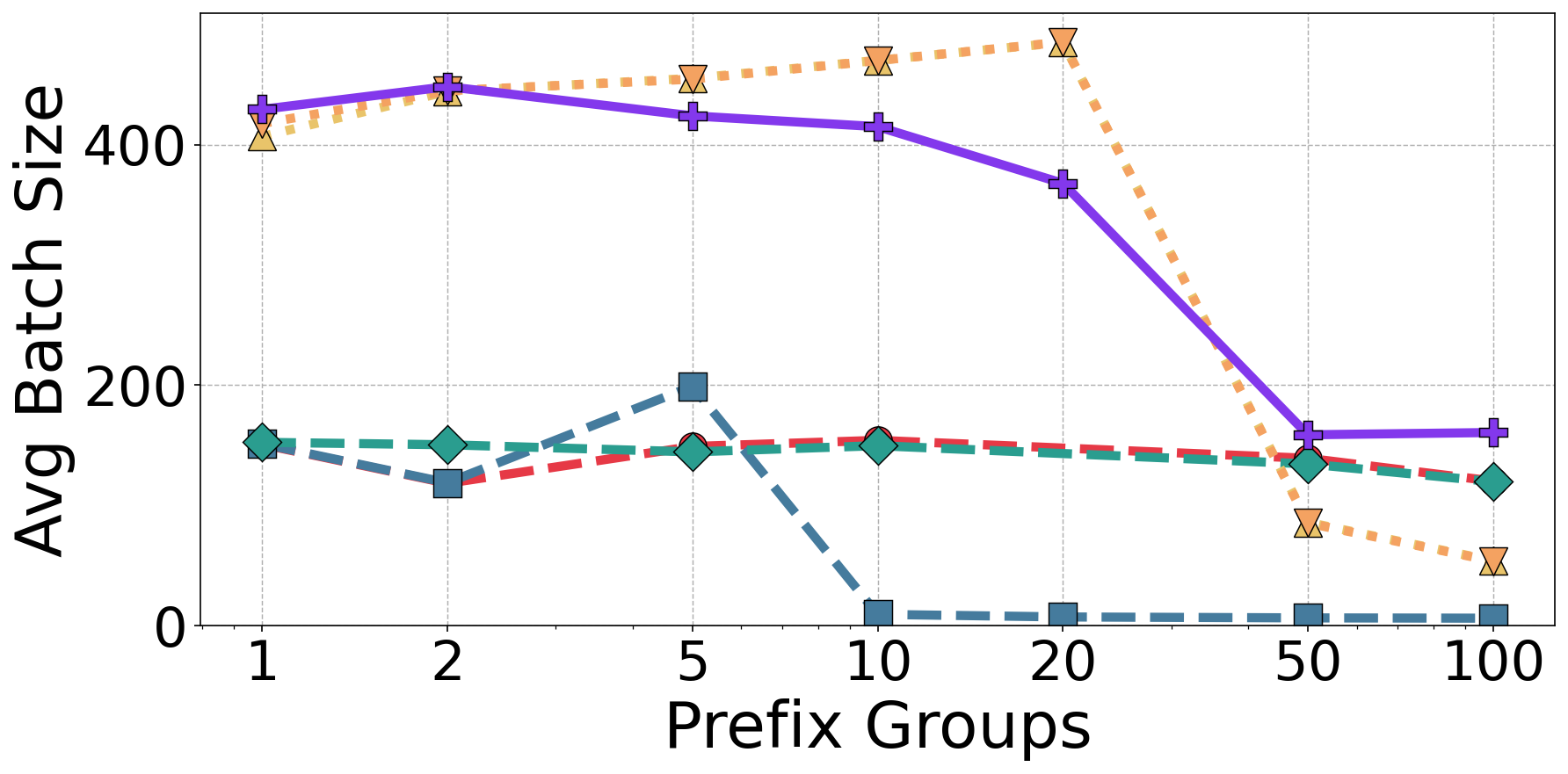}
    \caption{Average Batch Size}
    \label{fig:abs_vs_fam_l_5000}
\end{subfigure}
\caption{Varying Number of Prefix Groups}
\label{fig:num_prefix_groups}

\end{figure}

\subsubsection*{Workloads}
Following prior work, request arrivals are modeled as a Poisson process, with length pruning applied as required by each experiment. L-Eval~\cite{an2023leval} is a benchmark of human-labeled query–response pairs spanning tasks such as summarization and question answering, with sequence lengths ranging from 2.7K to 210.5K tokens. LongBench~\cite{bai2024longbench2} is a collection of long-context samples across six task categories: multi-document QA, summarization, and code completion, with context lengths between 4K and 10K tokens.

\subsubsection*{Baselines}
We compare \Feather against state-of-the-art LLM serving systems, all with prefix caching enabled. vLLM \cite{kwon2023efficient} uses its default FCFS scheduling policy. SGLang~\cite{zheng2024sglangefficientexecutionstructured} is evaluated under FCFS, LPM, and DFS-W to cover its design space. We also evaluate DynamicBatching~\cite{pang2025optimizing} and PAT~\cite{yi2026pat}, both implemented on top of vLLM: DynamicBatching adjusts batch sizes based on runtime memory utilization and latency constraints, while PAT is a prefix-aware attention kernel that reduces redundant KV cache accesses via query packing. \Feather operates purely at the batch formation layer and is agnostic to how prefill tokens are scheduled, so it composes naturally with chunked-prefill and prefill-decode multiplexing techniques such as Sarathi-Serve~\cite{agrawal2024tamingthroughputlatencytradeoffllm}. Chunked prefill is enabled by default in vLLM, the backend used throughout our experiments unless otherwise noted.

\subsubsection*{Metrics}
We report three key metrics. Throughput (toks/s) measures the output tokens generated per second, reflecting serving capacity and GPU utilization. Time between tokens (TBT) measures the average interval between decoded tokens, reflecting responsiveness, especially in interactive settings. We also track the average batch size per forward pass to relate scheduling decisions to these metrics.

\subsection{End-to-End Comparison: \Feather vs Baselines}
We evaluate a workload where the default configuration has 5 prefix groups, each 5K tokens long, with roughly equal request distribution, a request rate of 100 req/s, and 200 output tokens per request. Such workload configurations are common in prior work~\cite{yi2026pat,MLSYS2025_96894468} evaluating LLM serving systems. We vary these parameters to examine how \Feather compares to the baselines. We also evaluate workloads without prefix sharing and show \Feather performs on par with FCFS. The token budget is 32K tokens. For \Feather's RL policy, we find that bandit and Q-learning approaches perform similarly across most settings; unless stated otherwise, we report results using the bandit policy.

\subsubsection*{Varying Input Request Rate}
Figure~\ref{fig:tp_vs_rps_l_5000_fam_5} shows throughput (toks/s) as a function of the request rate. Across all policies, throughput increases with the request rate before saturating. \Feather (on vLLM) consistently outperforms all baselines at every request rate. At 100 req/s, \Feather achieves roughly $4\times$ the throughput of vLLM (FCFS). Notably, this advantage is not driven by larger batch sizes. As shown in Figure~\ref{fig:abs_vs_rps_l_5000_fam_5}, \Feather in general maintains a \textit{smaller} average batch size than vLLM (FCFS), and beyond the vertical line shown in the plot, \Feather's RL policy decides to have completely homogeneous batches. This reinforces a key insight: \textit{smaller, homogeneous batches outperform larger, heterogeneous ones}. The gradual growth in \Feather's batch size with increasing request rate reflects its ability to efficiently aggregate more requests from the same prefix group into the active batch without sacrificing locality. SGLang policies consistently operate at smaller batch sizes than vLLM, as vLLM avoids double-counting shared prefix tokens toward the token budget, allowing more requests per batch. Dynamic Batching performs similarly to vLLM (FCFS), as all KV caches fit within GPU memory. Figure~\ref{fig:tbt_vs_rps_l_5000_fam_5} shows time between tokens (TBT). \Feather (on vLLM) achieves substantially lower TBT than all baselines except SGLang (FCFS), whose smaller batch sizes yield faster per-token generation. While SGLang's LPM and DFS-W policies also use small batch sizes, their TBT remains significantly higher than \Feather due to non-trivial scheduling overhead during batch construction.  \Feather implemented on SGLang also shows similar qualitative trends and outperforms all other SGLang policies. Unless otherwise stated, all subsequent results refer to \Feather implemented on vLLM.

\begin{figure}[t]
\captionsetup[subfigure]{labelformat=simple}
    \renewcommand\thesubfigure{(\alph{subfigure})}
\centering
\begin{subfigure}{0.49\columnwidth}
    \centering
    \includegraphics[width=\linewidth]{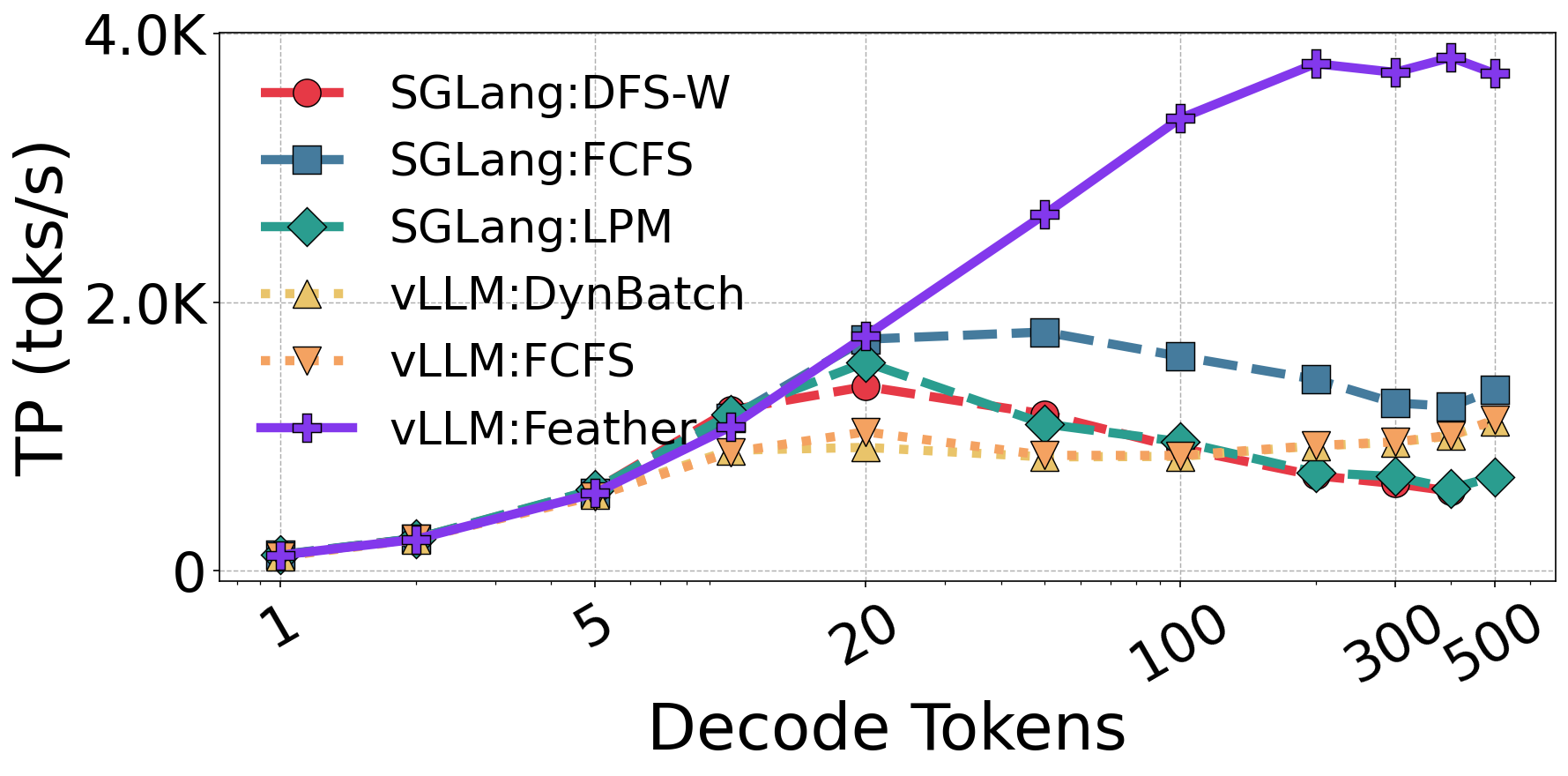}
    \caption{Throughput}
    \label{fig:tp_vs_tok_l_5000}
\end{subfigure}
\hfill
\begin{subfigure}{0.49\columnwidth}
    \centering
    \includegraphics[width=\linewidth]{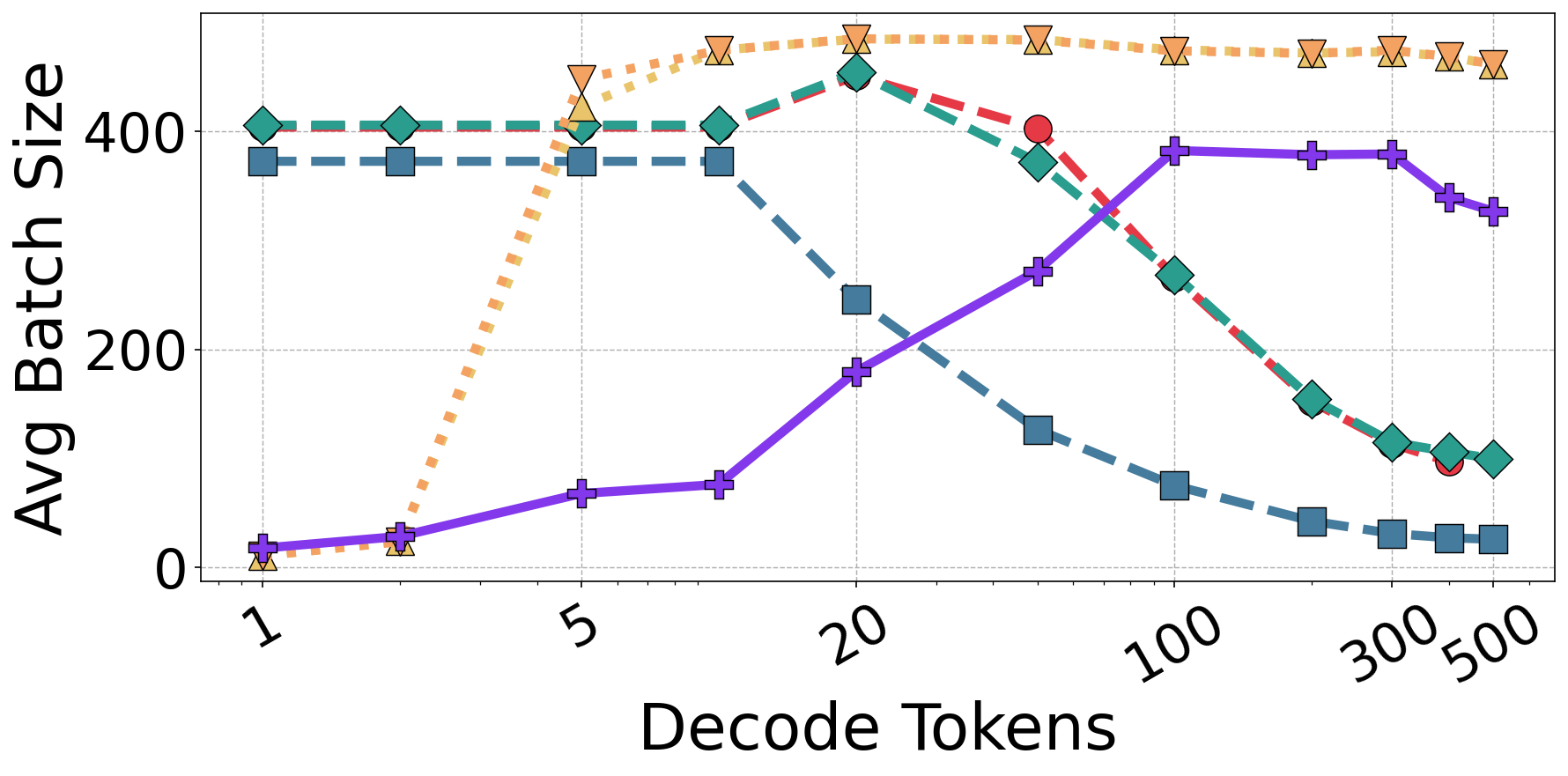}
    \caption{Average Batch Size}
    \label{fig:abs_vs_tok_l_5000}
\end{subfigure}
\caption{Varying Number of Decode Tokens}
\label{fig:num_decode_tokens}

\end{figure}

\begin{figure*}[t]
\centering
\begin{minipage}{0.33\textwidth}
    \centering
    \includegraphics[width=\linewidth]{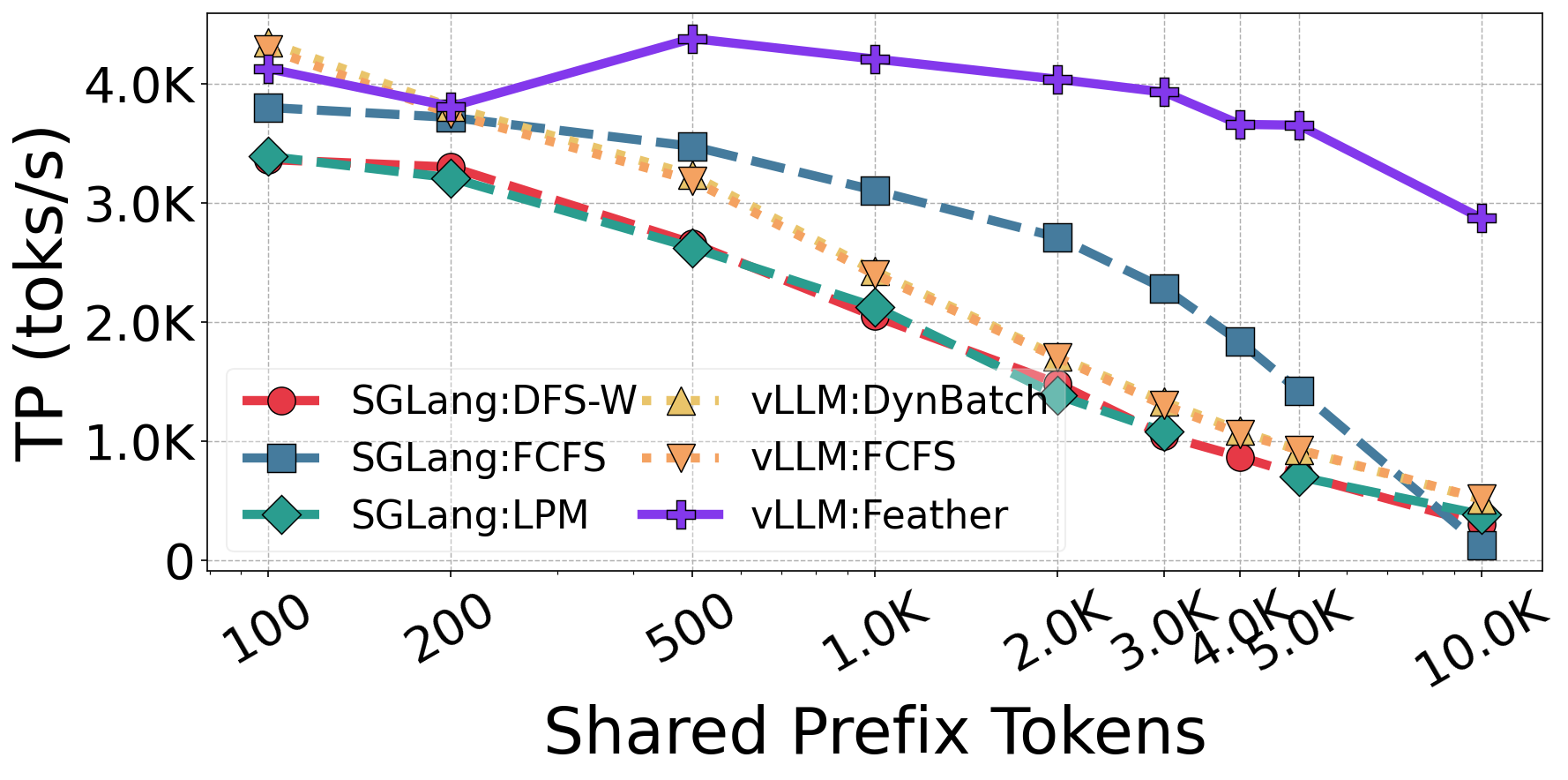}
    \caption{Varying Shared Prefix Tokens}
    \label{fig:tp_vs_sys_eval}
\end{minipage}
\hfill
\begin{minipage}{0.33\textwidth}
    \centering
\includegraphics[width=\linewidth]{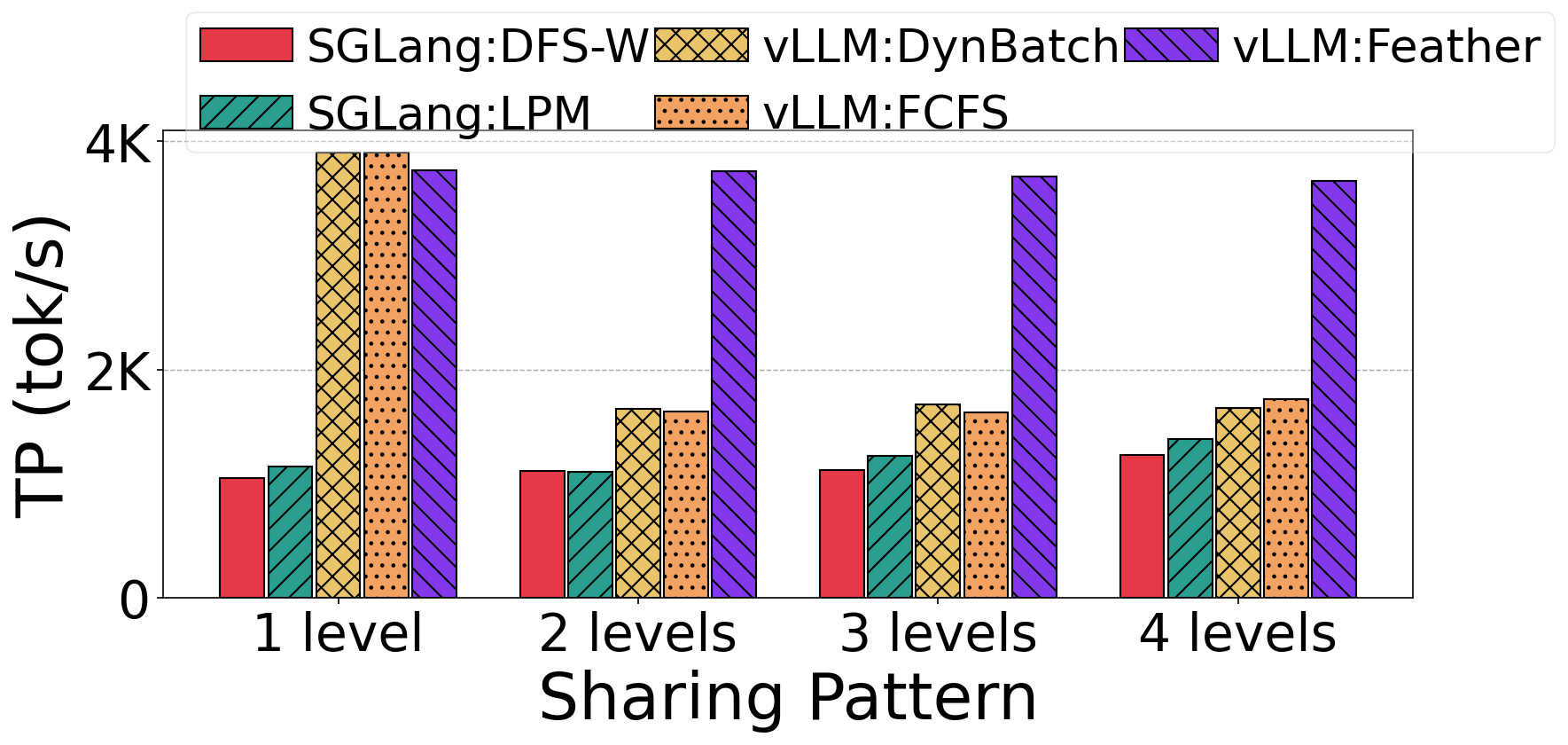}
\caption{Varying Radix Tree Sharing Levels}
\label{fig:tp_vs_rad_eval}
\end{minipage}
\hfill
\begin{minipage}{0.33\textwidth}
   \centering

\includegraphics[width=\linewidth]{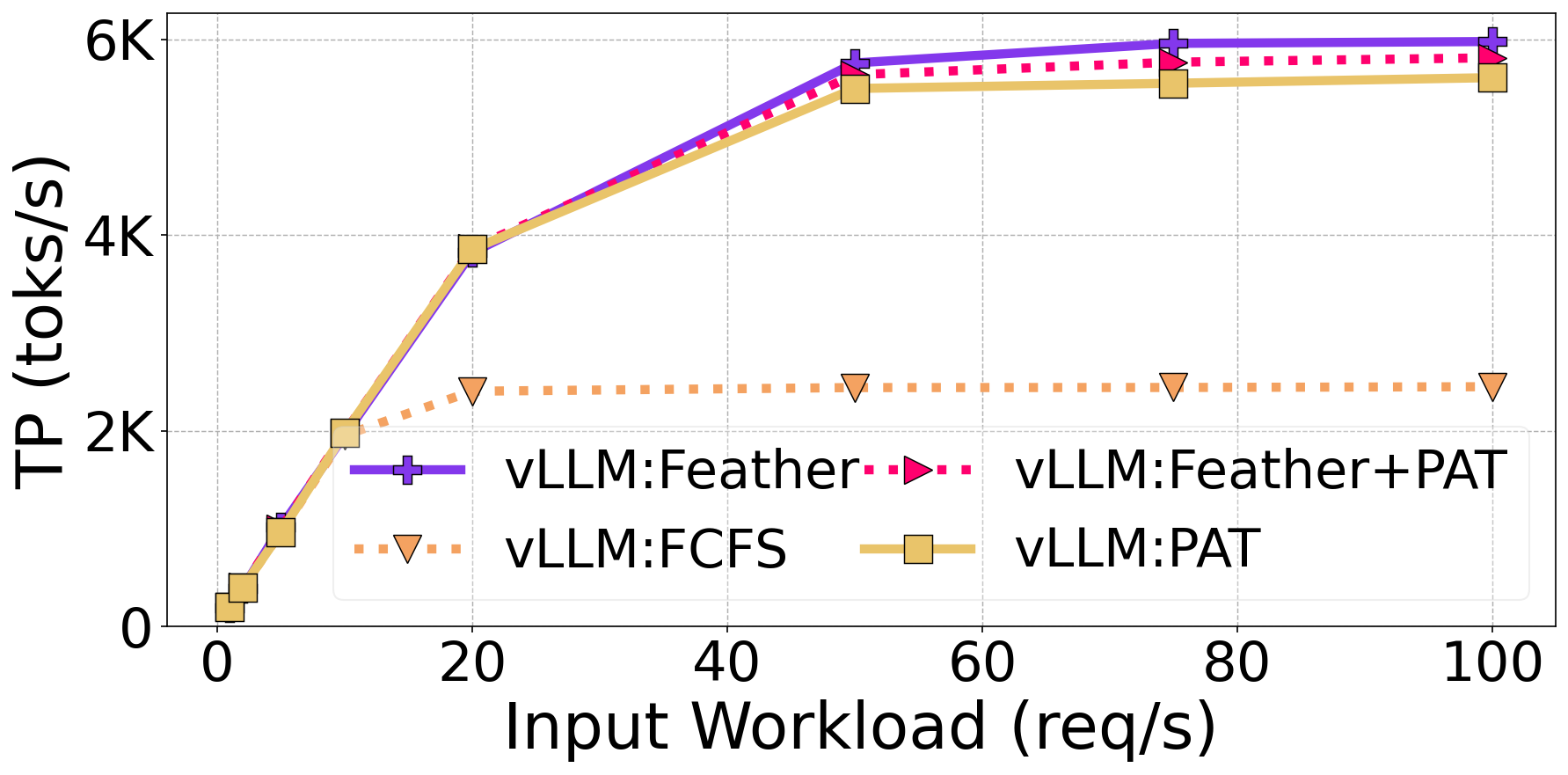}
\caption{\Feather vs PAT}
\label{fig:throughput_vs_rps_pat_fcfs_feather_eval}
\end{minipage}

\end{figure*}

\begin{figure*}[t]
\centering
\begin{minipage}{0.33\textwidth}
   \centering
\includegraphics[width=\linewidth]{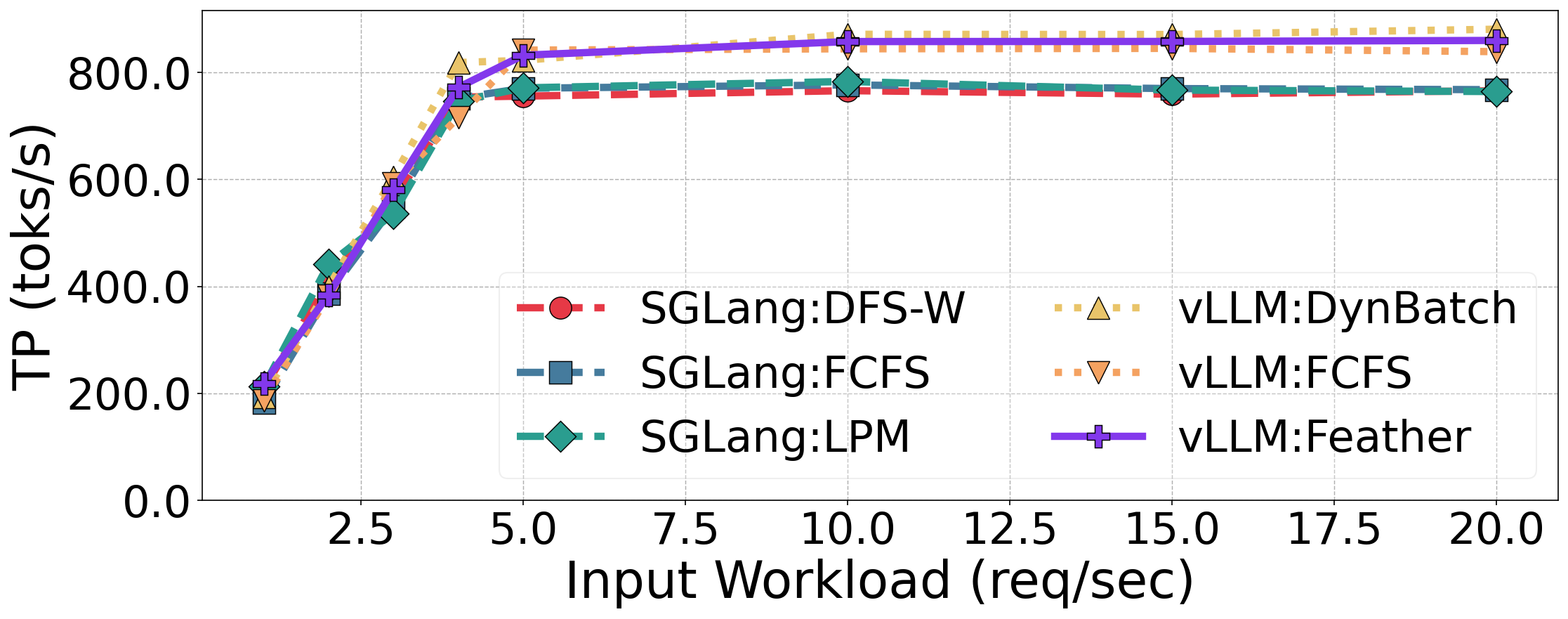}
\caption{Throughput with no prefix sharing}
\label{fig:tp_no_pref}
\end{minipage}
\hfill
\begin{minipage}{0.33\textwidth}
   \centering
\includegraphics[width=\linewidth]{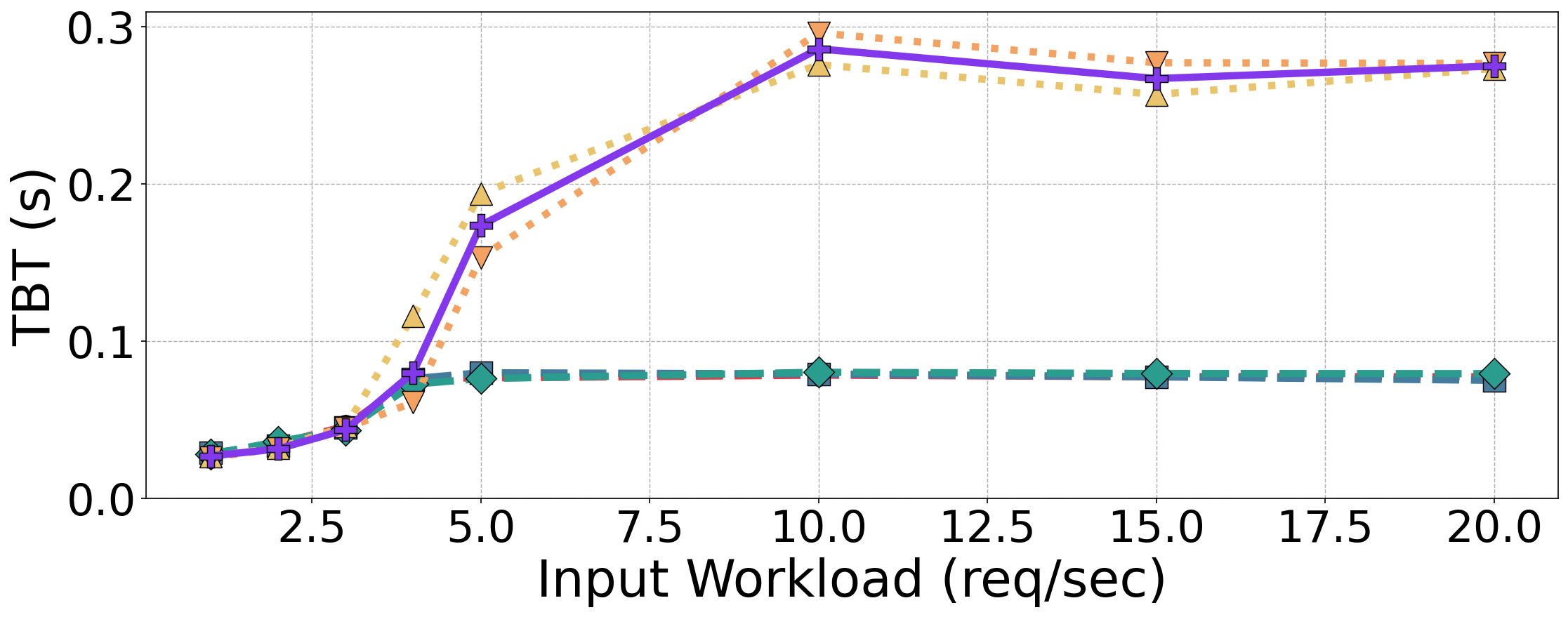}
\caption{Time between tokens (TBT) with no prefix sharing}
\label{fig:tbt_no_pref}
\end{minipage}
\hfill
\begin{minipage}{0.33\textwidth}
   \centering
\includegraphics[width=\linewidth]{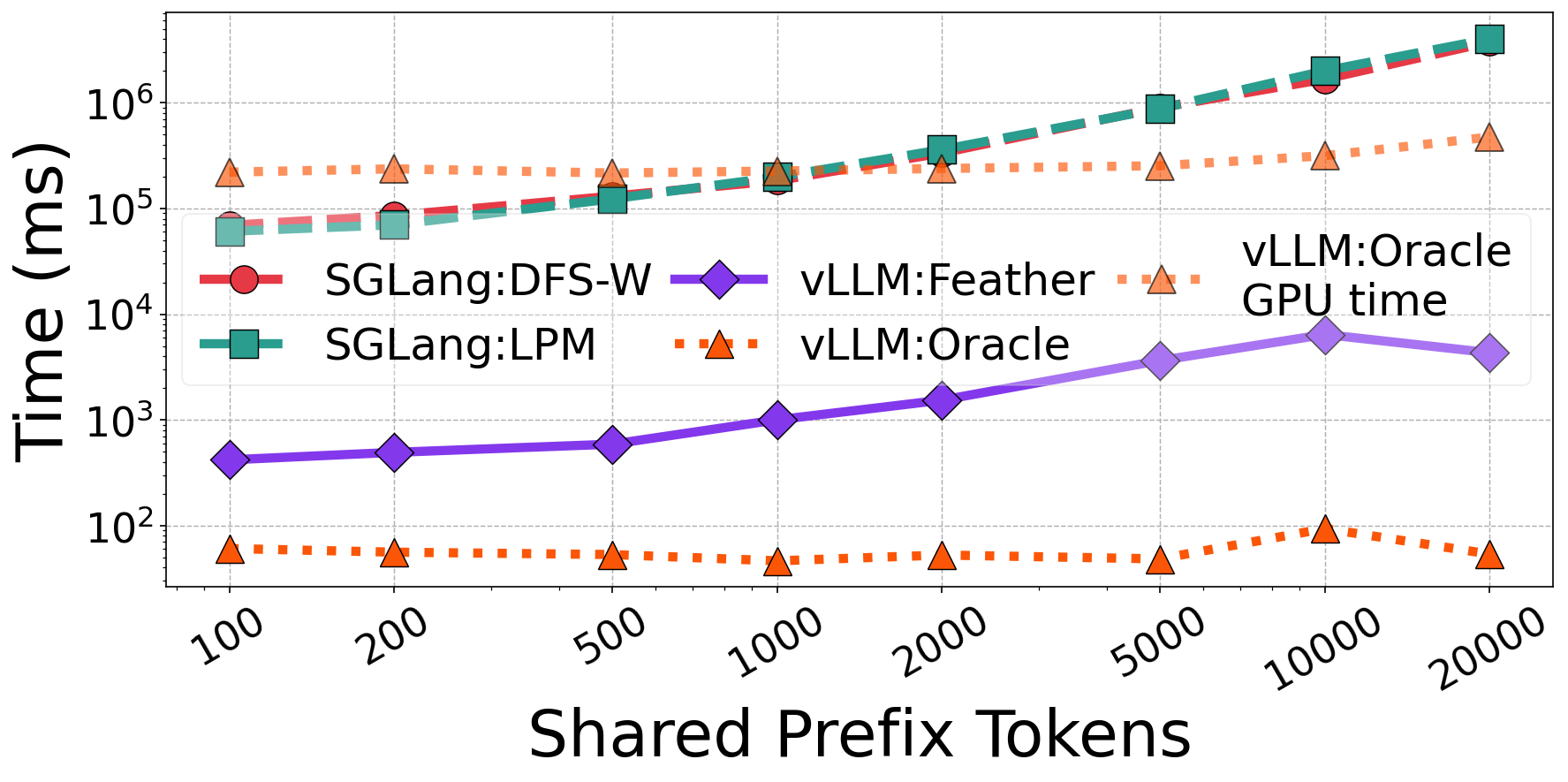}
\caption{CPU overhead across scheduling policies}
\label{fig:profile_sum_vs_sys_prompt_tokens_eval}
\end{minipage}

\end{figure*}

\subsubsection*{Varying Number of Prefix Groups}
Figures~\ref{fig:tp_vs_fam_l_5000} and \ref{fig:abs_vs_fam_l_5000} show throughput and average batch size as functions of the number of prefix groups. Increasing the number of prefix groups introduces greater request heterogeneity, reducing KV cache locality. This leads to a sharp throughput drop for vLLM (FCFS), even when increasing from 1 to 2 prefix groups. As the number of groups increases further, FCFS degrades rapidly: beyond 20 groups, the KV cache footprint exceeds HBM capacity, causing a sharp reduction in batch size and increased recomputation. \Feather mitigates this by grouping requests with shared prefixes, increasing KV cache reuse and avoiding premature evictions. As a result, \Feather maintains a more stable batch size and throughput across all prefix group configurations, significantly outperforming all baselines. At 100 prefix groups, \Feather achieves $10\times$ higher throughput than vLLM (FCFS). SGLang's LPM and DFS-W policies partially mitigate evictions, achieving higher throughput than FCFS (SGLang) at larger group counts.

\subsubsection*{Varying Number of Decode Tokens}
Figures~\ref{fig:tp_vs_tok_l_5000} and \ref{fig:abs_vs_tok_l_5000} show throughput and average batch size as functions of decode tokens per request. Each decode requires a full sweep over the KV cache. The policies exhibit two regimes separated by a crossover around 20 decode tokens. In the prefill-dominated regime (1--20 tokens), all policies behave similarly: throughput increases modestly, and batch sizes converge as prefills limit the impact of locality-aware scheduling. Beyond 20 tokens, the policies diverge sharply. \Feather scales strongly with decode length, reaching $\sim$3800 toks/s at 200 tokens—over $3\times$ higher than SGLang policies. This gain arises from two factors. First, longer decode phases amplify locality across repeated KV cache sweeps. Second, a greater number of decode iterations allows the scheduler to accumulate more same-prefix requests, forming larger, more homogeneous batches. Together, these effects drive the steady increase in \Feather's batch size and throughput. SGLang policies follow the opposite trajectory in batch size. Because shared prefix tokens count toward the token budget, each prefill consumes $\sim$5K tokens, while each decode step contributes only one token. As the decode length increases, fewer prefills occur per step, reducing the number of new requests for batching.

\subsubsection*{Varying Number of Shared Prefix Tokens}
Figure~\ref{fig:tp_vs_sys_eval} shows throughput across the length of the shared prefix. As expected, longer prefixes increase prefill compute and KV cache pressure, reducing throughput across policies. \Feather's advantage grows with prefix length, driven by higher locality gains as sequences lengthen.

\subsubsection*{Varying Models}
\Feather consistently achieves the highest throughput across models and outperforms baselines in nearly all cases. Its relative throughput advantage increases with model size, as larger models spend more time in attention, amplifying the benefits of KV-cache locality exploited by \Feather. All graphs are shown in \S\ref{subsec:across_models}. We also compare \Feather with other policies (\S\ref{subsec:results_across_longchat_model}) for the LongChat 13B model, where we achieve up to 22$\times$ higher throughput compared to vLLM's FCFS for larger number of prefix groups.

\subsubsection*{Varying Radix Tree Levels}
Figure~\ref{fig:tp_vs_rad_eval} shows throughput under different radix-tree sharing levels (seen in \S\ref{subsec:significance_of_prefix_homogeneity}). All experiments use a fixed total sequence length of 4000 tokens per request. For baseline policies like vLLM's FCFS, going from 1 to 2 or more levels of the radix tree reduces the locality in KV cache traversal and, hence, degrades throughput. In contrast, \Feather actively preserves prefix homogeneity within batches, leading to stable throughput at all levels.

\subsubsection*{Comparison with PAT}
Figure~\ref{fig:throughput_vs_rps_pat_fcfs_feather_eval} compares throughput across varying request rates for PAT, vLLM (FCFS), and \Feather on an A100-80GB GPU. PAT is specifically optimized for this hardware through custom prefix-aware kernels. Despite this, \Feather consistently matches or outperforms PAT across all request rates, demonstrating strong performance even on different hardware. This is particularly notable because \Feather operates purely at the scheduling layer, without any kernel-level modifications. A naive integration of \Feather and PAT performs quite similarly to \Feather over most regimes. As future work, we aim to more tightly integrate \Feather with prefix-aware attention kernels, enabling deeper co-design between the scheduler and kernel-level optimizations to further improve performance.

\subsubsection*{Workloads with No Prefix Sharing}
\label{subsec:workloads_with_no_prefix_sharing}
We next evaluate \Feather and compare it against other scheduling policies on a workload without prefix sharing; that is, no two requests share a common prefix. Figures~\ref{fig:tp_no_pref} and \ref{fig:tbt_no_pref} report throughput and time between tokens (TBT) across a range of input rates. At higher input rates, vLLM-based policies begin to outperform SGLang policies, primarily due to their ability to form larger batches. Notably, even in the absence of prefix sharing, \Feather, powered by its adaptive RL-based policy—matches the performance of vLLM's FCFS scheduler. In contrast, SGLang policies achieve lower TBT, which can be attributed to their smaller batch sizes.

\subsubsection*{Latency and Starvation}
\begin{figure}[t]
    \centering
    \includegraphics[width=0.8\linewidth]{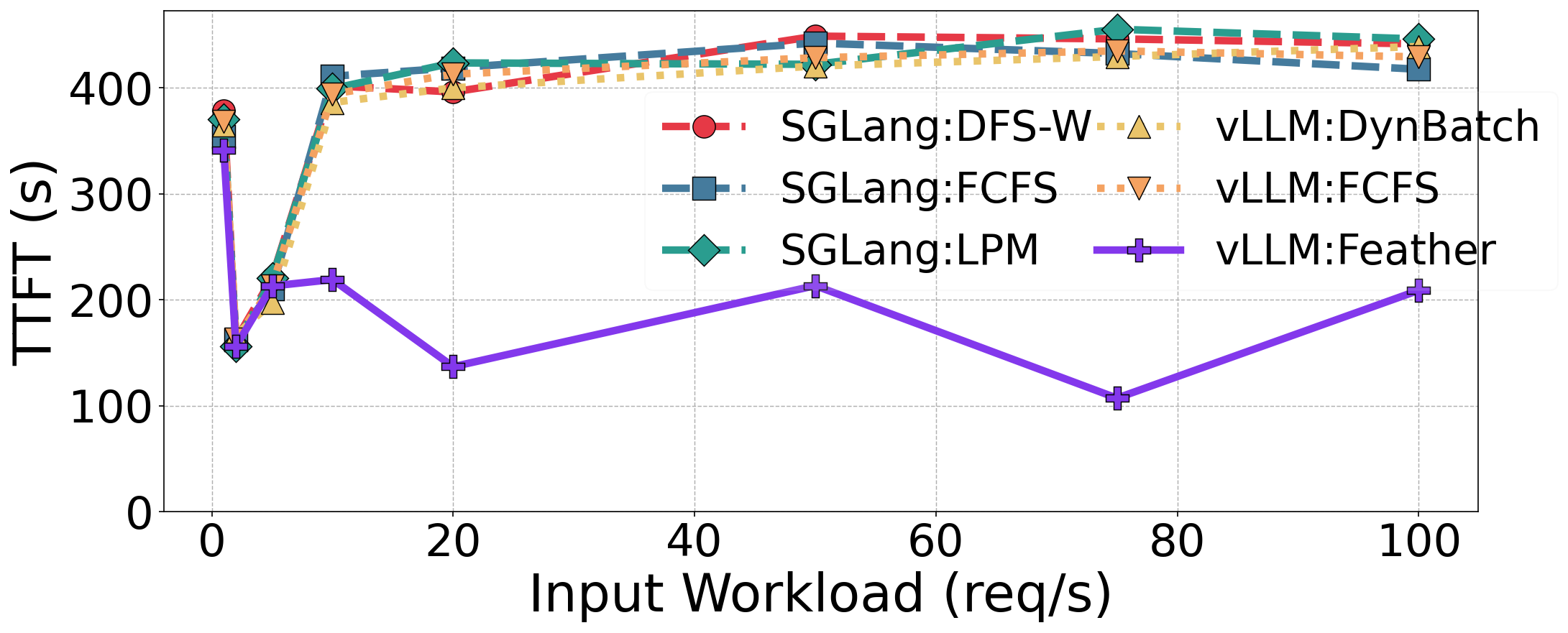}
    \caption{Time to First Token (TTFT) vs. req/sec}
    \label{fig:ttft_vs_rps}
    
\end{figure}

While our main evaluation focuses on aggregate throughput and TBT, a natural concern is whether \Feather's prefix-homogeneous batching starves less popular prefix groups in favor of larger ones. Two design choices mitigate this risk. First, within \textsc{FindBest}'s min-heap, entries with equal missing count are ordered by request ID, which corresponds to arrival order. Among requests that are otherwise equally compatible with the working set, earlier-arriving requests are therefore preferred, preventing indefinite deferral of any single request. Second, the exploration phase of the RL policy (\S\ref{subsec:batching_rl}) periodically constructs heterogeneous batches that include less popular prefixes, which ensures prefixes are not excluded from scheduling over the long term. As shown in Figure~\ref{fig:ttft_vs_rps}, \Feather consistently achieves substantially lower TTFT than all SGLang and vLLM baselines across the tested workload range. This gap stems from \Feather's prefix-homogeneous batching completing requests faster on average, which frees scheduling slots sooner and lets newly arriving requests enter execution with shorter queueing delay, rather than waiting behind a backlog of slow, heterogeneous batches. The RL policy's periodic exploration further contributes to this outcome. By regularly admitting less popular prefixes into the batch rather than deferring them indefinitely, \Feather reduces starvation risk in the tested workloads.

\subsection{Micro-Benchmarks}
\subsubsection*{Comparing Scheduler Compute Overhead across Different Policies}
Figure~\ref{fig:profile_sum_vs_sys_prompt_tokens_eval} shows the CPU scheduling overhead for each policy across different shared prefix lengths. We include an Oracle policy where the caller supplies prefix group identifiers with each request. This allows the scheduler to batch same-prefix requests without any prefix matching overhead, resulting in effectively zero scheduling cost while preserving locality. We also show the GPU execution time for this Oracle scheduling policy. All three realistic policies—\Feather, LPM, and DFS-W— incur higher overhead as the prefix length increases due to deeper radix-tree traversals and larger key comparisons. However, their growth rates diverge significantly. At 20K tokens, LPM and DFS-W incur roughly 1,000$\times$ higher overhead than \Feather. At larger sequence lengths, their scheduling overhead exceeds the Oracle GPU execution time by up to 10$\times$. This indicates that CPU overhead alone can negate the benefits of cache-aware batching. In contrast, \Feather remains below 1\% of GPU execution time across the entire range. We do a detailed analysis of the overheads of LPM and DFS-W policies in \S\ref{subsec:scheduling-complexity}.

\begin{figure}[t]
\captionsetup[subfigure]{labelformat=simple}
    \renewcommand\thesubfigure{(\alph{subfigure})}
\centering
\begin{subfigure}{0.49\columnwidth}
    \centering
\includegraphics[width=\linewidth]{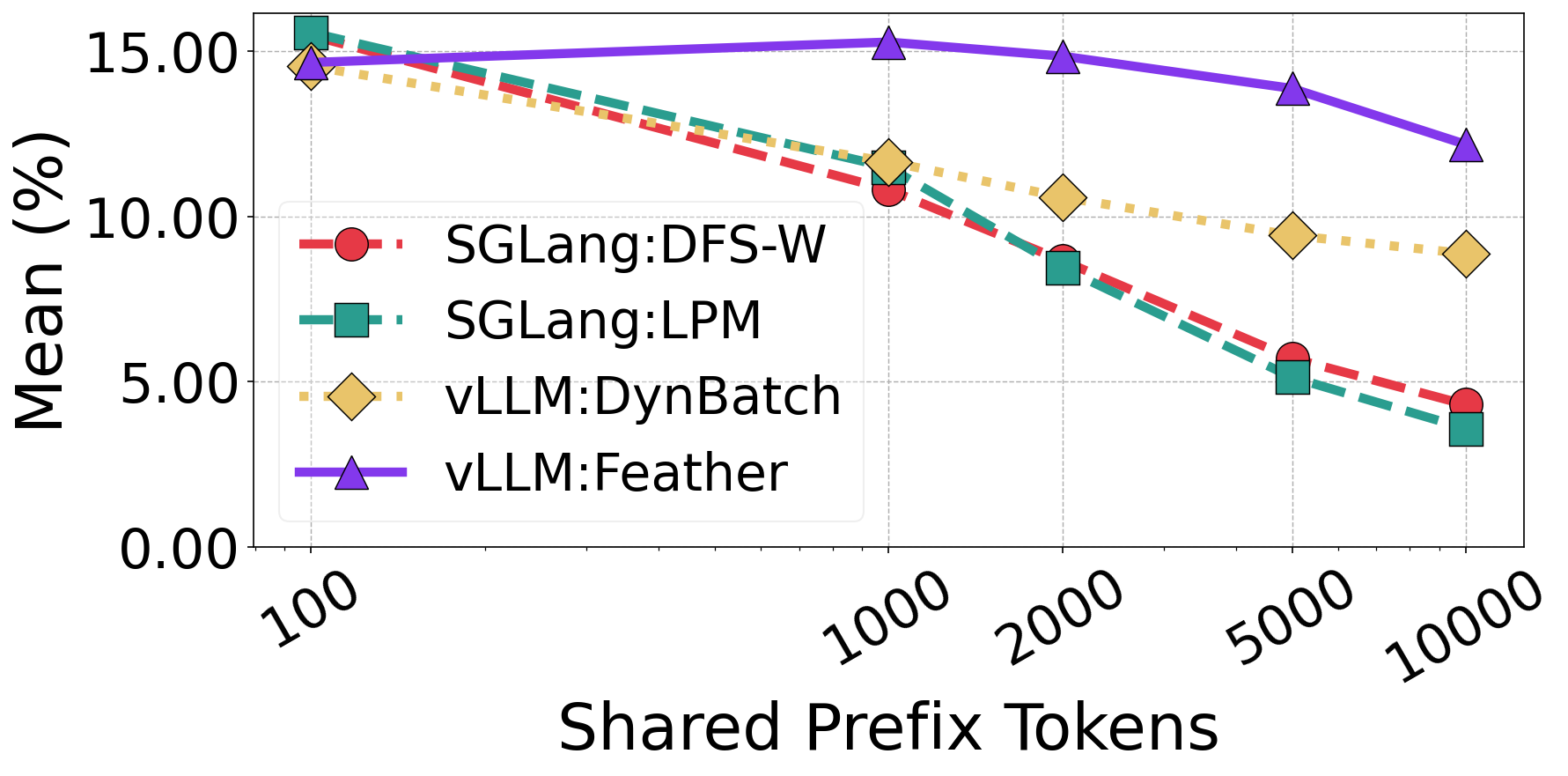}
\caption{Mean}
\label{fig:mean_drama_vs_sys_prompt_tokens}
\end{subfigure}
\hfill
\begin{subfigure}{0.49\columnwidth}
   \centering
\includegraphics[width=\linewidth]{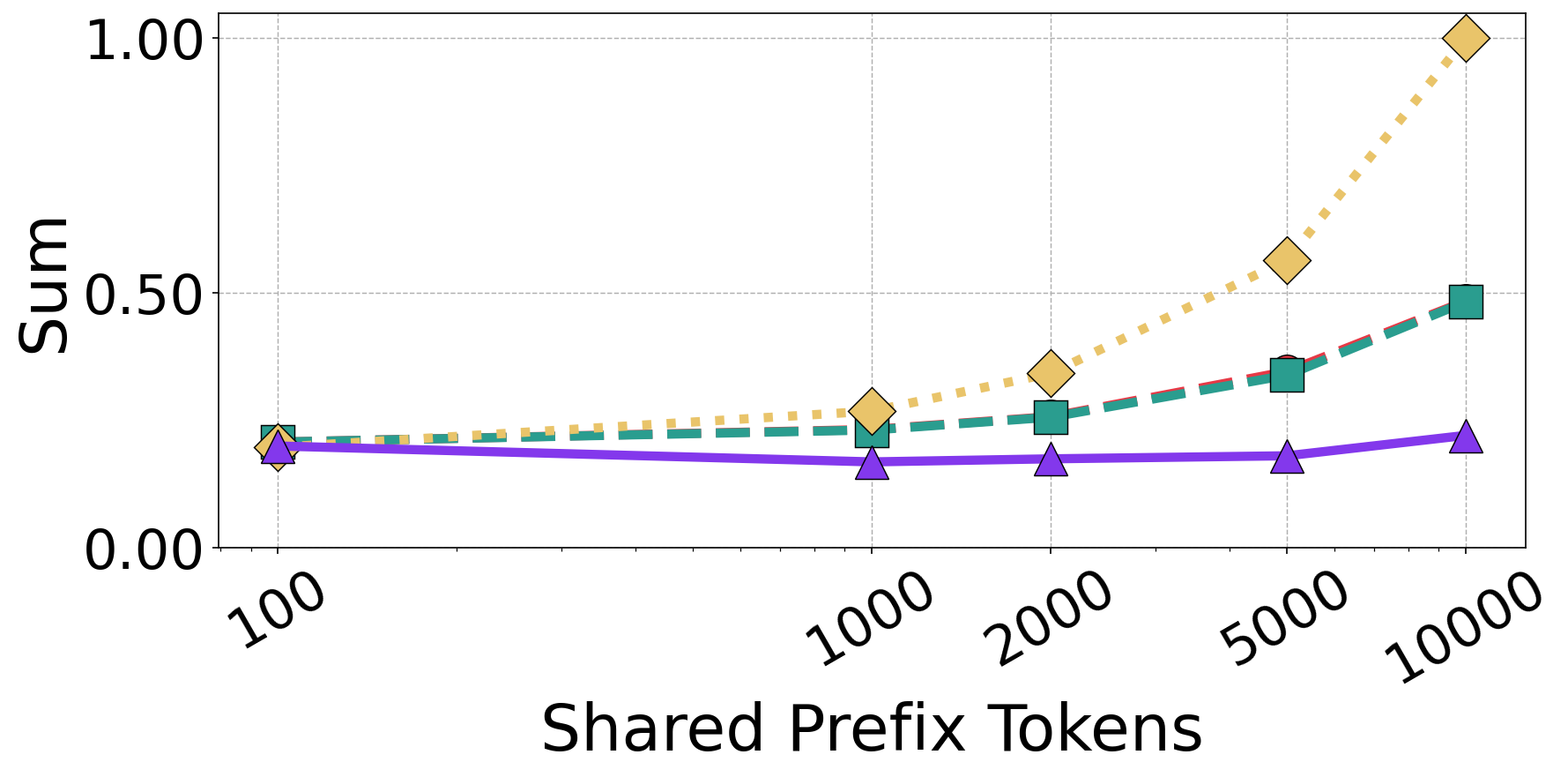}
\caption{Sum}
\label{fig:sum_dram_vs_sys_prompt_tokens}
\end{subfigure}
\caption{DRAM Bandwidth Utilization}
\label{fig:microbenchmark}

\end{figure}

\begin{figure}[t]
\captionsetup[subfigure]{labelformat=simple}
    \renewcommand\thesubfigure{(\alph{subfigure})}
\centering
\begin{subfigure}{0.49\columnwidth}
    \centering
\includegraphics[width=\linewidth]{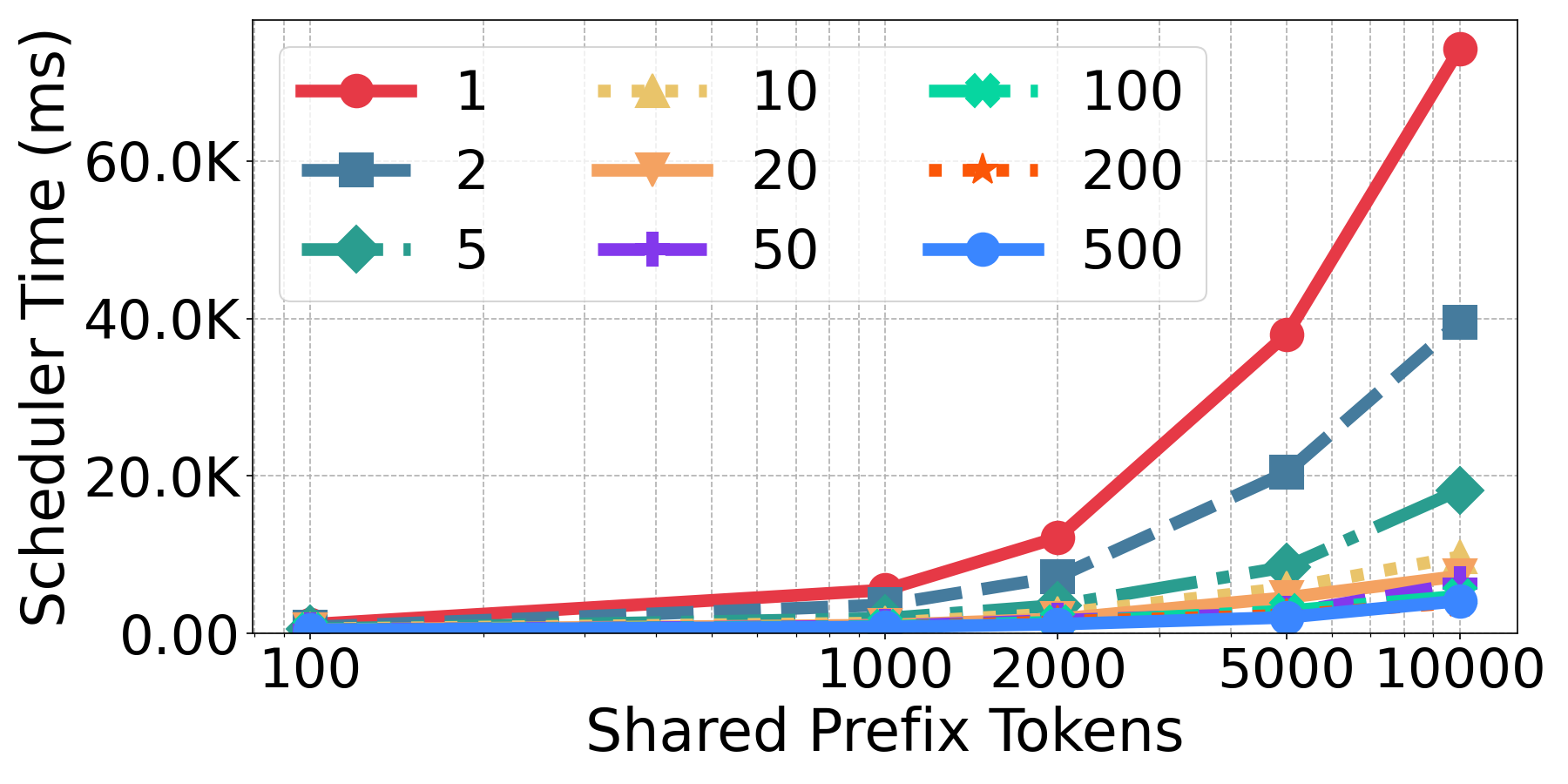}
\caption{CPU Scheduler Overhead}
\label{fig:functotals_vs_systok_chunks}
\end{subfigure}
\hfill
\begin{subfigure}{0.49\columnwidth}
    \centering
\includegraphics[width=\linewidth]{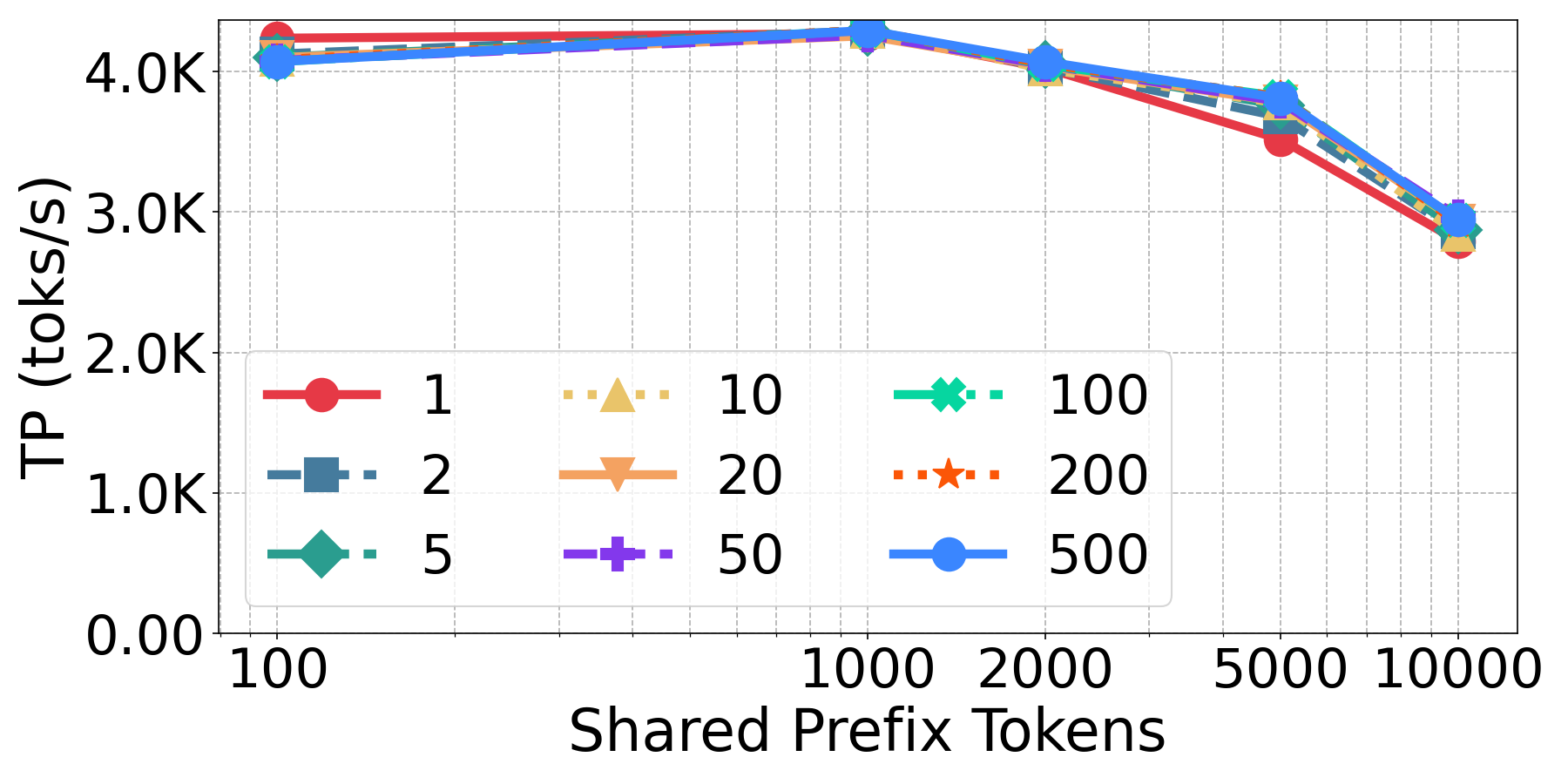}
\caption{Throughput}
\label{fig:throughput_vs_systok_chunks}
\end{subfigure}
\caption{Sensitivity to Chunk Size}
\label{fig:chunk_size_sensitivity}

\end{figure}

\subsubsection*{Mean DRAM Activity across Various Sequence Lengths}
Figure~\ref{fig:mean_drama_vs_sys_prompt_tokens} shows mean DRAM bandwidth utilization, while Figure~\ref{fig:sum_dram_vs_sys_prompt_tokens} shows the total DRAM usage over the full inference duration (a proxy for total memory accessed) across policies and shared prefix lengths. \Feather consistently achieves the highest bandwidth utilization due to improved spatial locality, leading to more effective prefetching. Furthermore, more homogeneous batches also increase temporal locality, which is reflected in lower total DRAM usage. SGLang's DFS-W and LPM exhibit lower mean DRAM utilization because of CPU scheduler stalls. However, they also achieve lower normalized total DRAM usage, as their prefix-aware scheduling occasionally forms homogeneous batches, allowing them to partially benefit from memory locality.

\subsubsection*{Effect of Chunk Size}
Figures~\ref{fig:functotals_vs_systok_chunks} and~\ref{fig:throughput_vs_systok_chunks} show CPU overhead and throughput across shared prefix lengths for different chunk sizes in \Feather's CHT. Increasing the chunk size reduces CPU overhead by enabling coarser-grained matching; a chunk size of 1 is equivalent to an exact radix tree. Overhead rises with prefix length due to more hash comparisons, while throughput declines due to higher KV cache memory pressure. At larger prefix lengths, the overhead gap between chunk size 1 and larger sizes widens substantially, yet throughput remains similar across all chunk sizes, suggesting that the CPU scheduling cost, even at the finest granularity, is not the primary bottleneck in our CHT. Overall, CHT is robust against chunk size selection, and even its radix tree-equivalent configuration (chunk size 1) achieves significantly better efficiency than a conventional radix tree.

\subsection{Ablation Study}
\subsubsection*{What if we Replace the Chunked Hash Tree with a Radix Tree?}
\begin{figure}[t]
\centering
\begin{minipage}{0.49\columnwidth}
   \centering
\includegraphics[width=\linewidth]{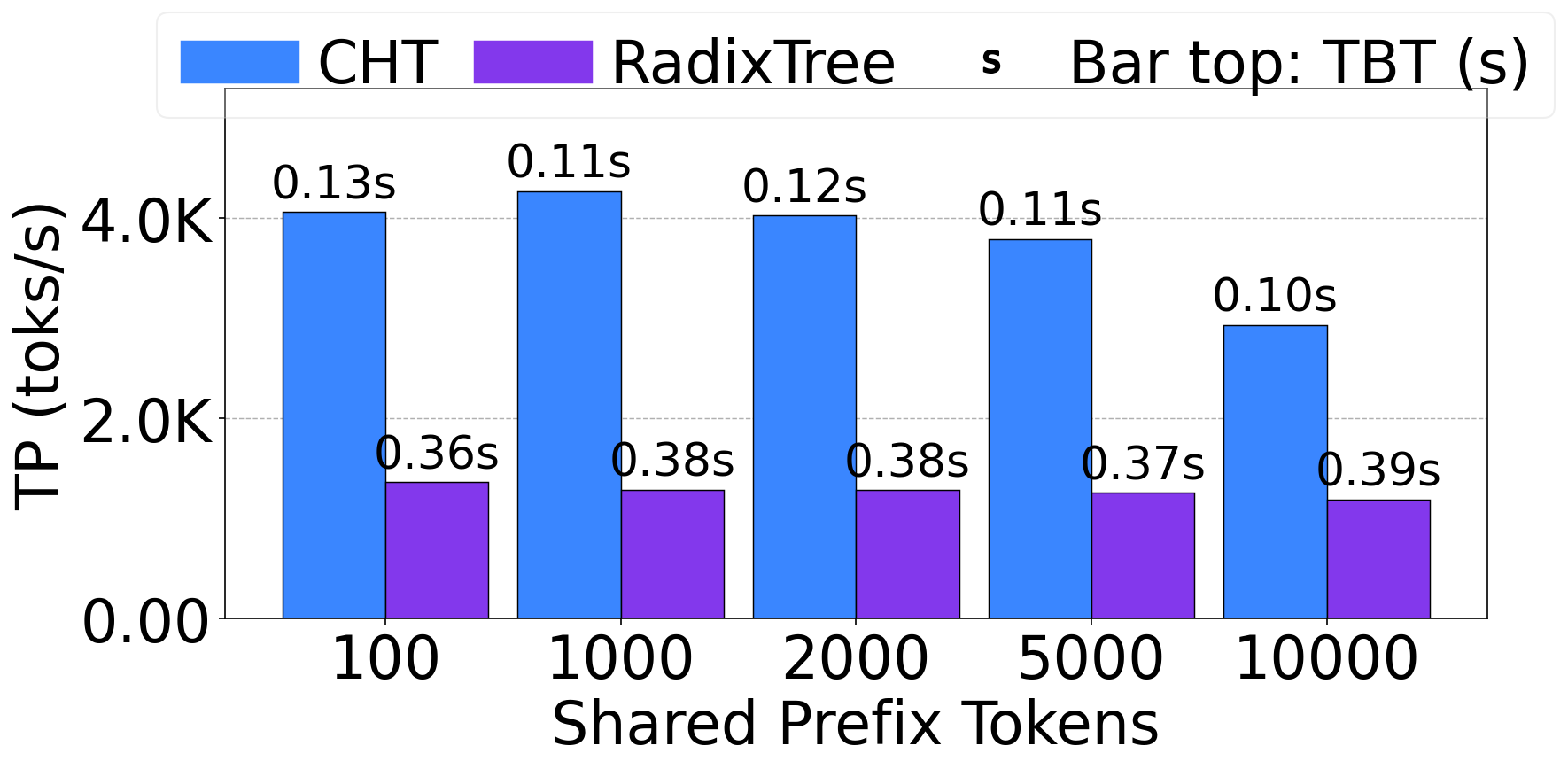}
\caption{CHT vs. Radix Tree}
\label{fig:throughput_vs_sys_prompt_tokens_bar}
\end{minipage}
\hfill
\begin{minipage}{0.49\columnwidth}
   \centering
\includegraphics[width=\linewidth]{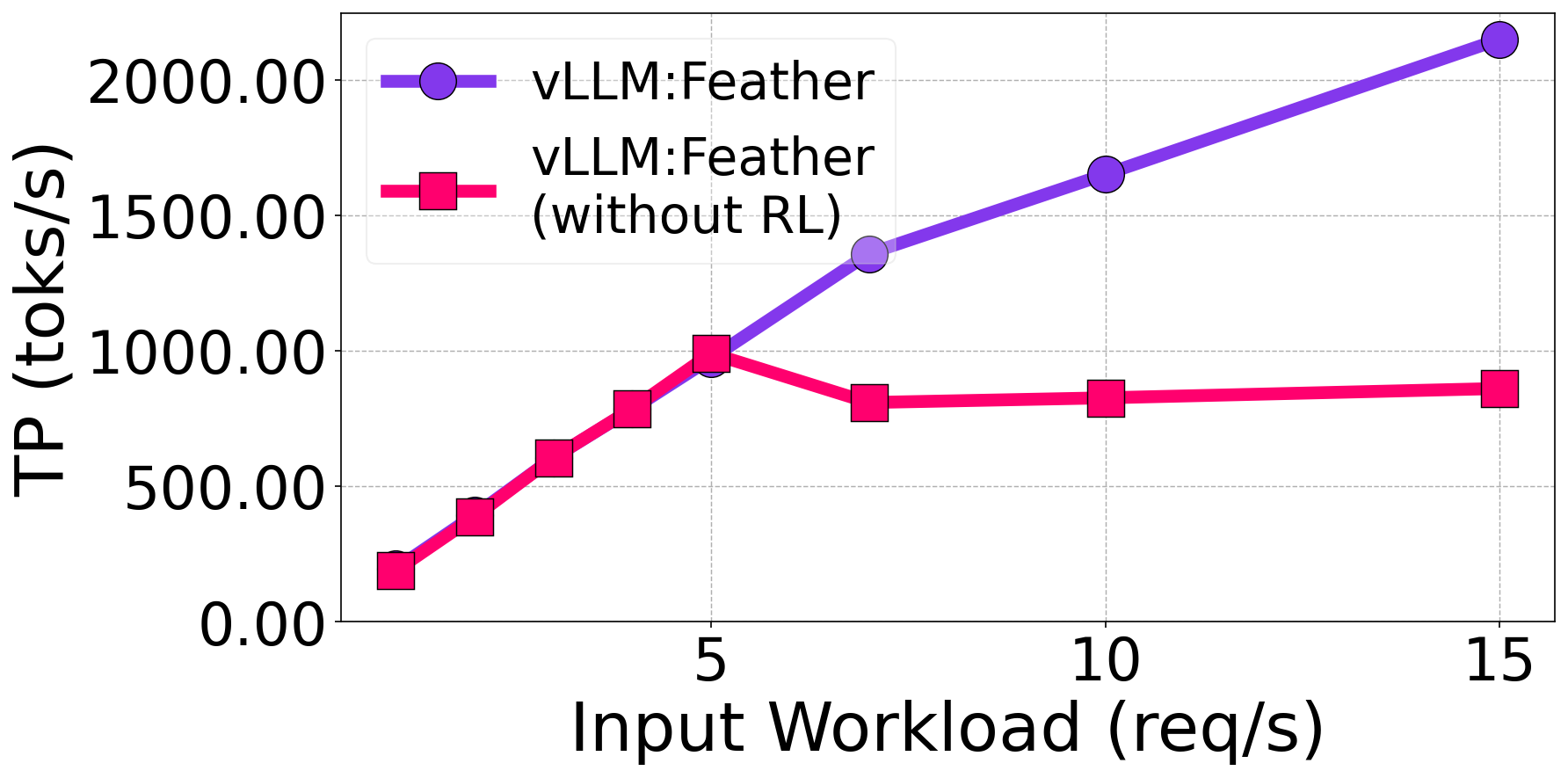}
\caption{\Feather: with and without RL}
\label{fig:tp_vs_req_per_sec_ablation}
\end{minipage}

\end{figure}

Figure~\ref{fig:throughput_vs_sys_prompt_tokens_bar} compares throughput and TBT when replacing CHT with a radix tree under the same RL policy. Across all prefix lengths, the radix tree yields much lower throughput. Despite homogeneous RL batching, higher CPU overhead causes frequent GPU stalls. This appears as higher TBT, indicating GPU idle time due to scheduling delays. CHT avoids this by reducing scheduling overhead, enabling higher GPU utilization and throughput.

\subsubsection*{What if We Remove RL?}
Figure~\ref{fig:tp_vs_req_per_sec_ablation} compares throughput with and without RL at low input rates. Without RL, CHT still selects the best request, but batching never halts and always maximizes batch size. At <5 req/s, removing RL has little effect. Both fall back to FCFS, adding requests to keep GPUs utilized. Beyond this, the behaviors diverge. With RL, \Feather detects homogeneous requests and stops batching at the right time, achieving higher throughput despite smaller average batch sizes. At higher rates than shown in the figure, large pools enable even non-RL to form ${\sim}500$-request homogeneous batches, leading the two curves to converge. These results show that \Feather's RL policy makes load-aware decisions: it falls back to FCFS under low utilization to avoid GPU starvation, and closes batch formation when homogeneity is achievable, maximizing overall throughput.

\subsubsection*{Convergence of Bandit Policy}
\label{subsec:convergence_of_bandit_policy}
\begin{figure}[t]
    \centering
    \includegraphics[width=0.8\linewidth]{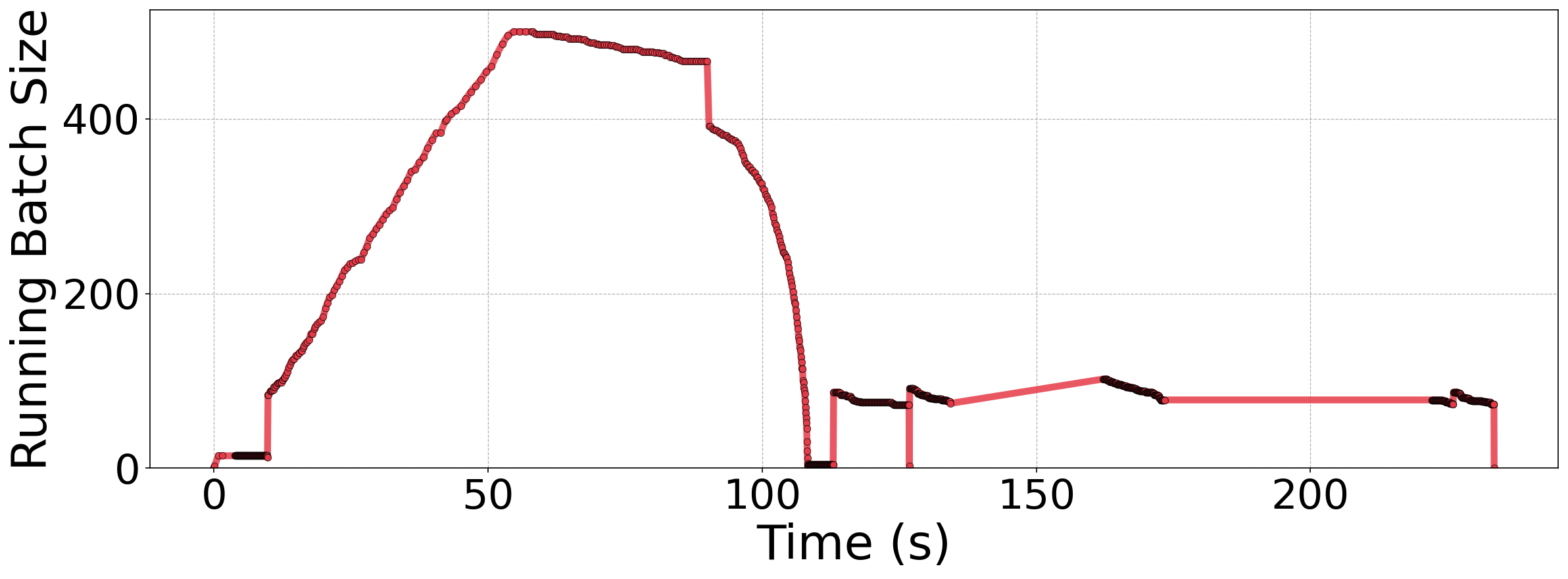}
    \caption{Convergence of Bandit Policy}
    \label{fig:convergence_of_bandit_policy}
    
\end{figure}

Having shown that RL improves throughput over a non-RL baseline, we also evaluate how quickly \Feather's bandit-based policy converges to a stable and efficient batching configuration. We construct a workload with five prefix groups, each containing 5K tokens, and issue requests at a rate of 5 per second. Figure~\ref{fig:convergence_of_bandit_policy} illustrates how the running batch size evolves over time for \Feather under its bandit-based policy. At the beginning, the system enters an exploration phase, during which \Feather experiments with different configurations and reaches relatively large batch sizes (around 500). However, these batches are heterogeneous. After this phase concludes (around 100 seconds), the system realizes that homogeneous batching yields higher throughput, and converges to a significantly smaller batch size for the remainder of the experiment.

\subsubsection*{RL Policy across Varying Workload}
\label{subsec:rl_policy_across_varying_worklaod}
\begin{figure}[t]
    \centering
    \includegraphics[width=0.8\linewidth]{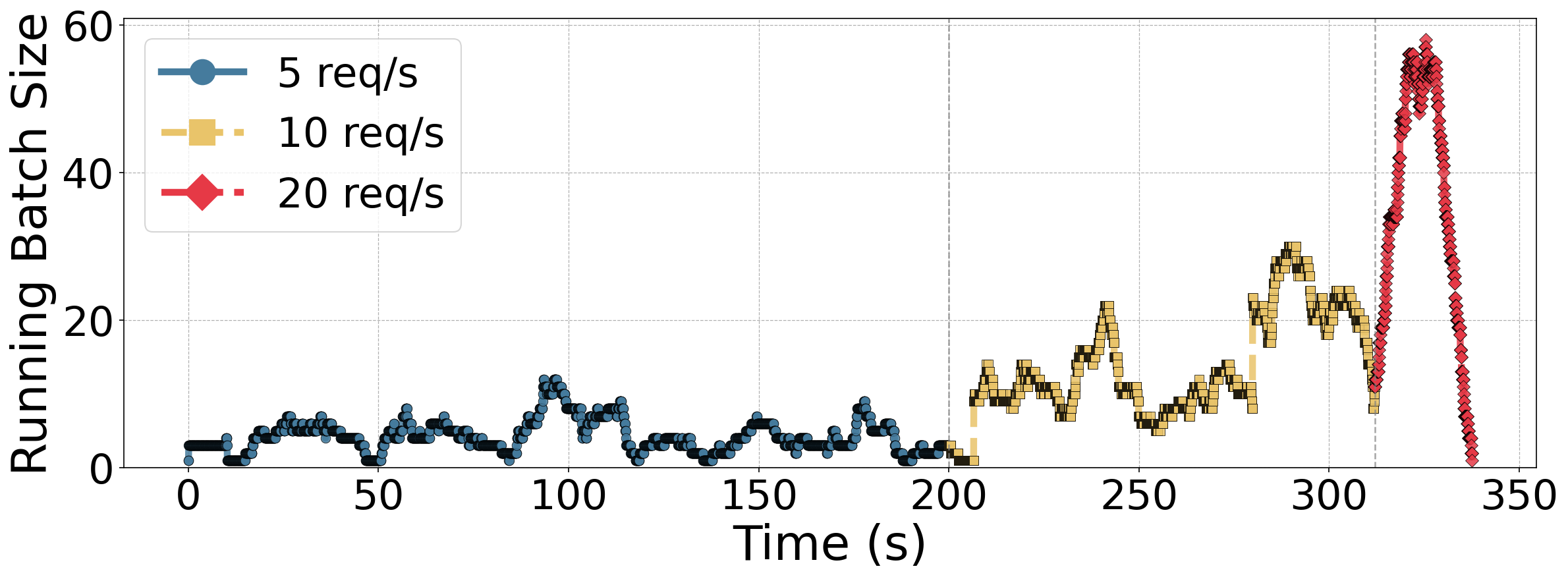}
    \caption{Performance of Bandit across Varying Workload}
    \label{fig:rl_policy_across_varying_worklaod}
    
\end{figure}

In all experiments so far, including those across models (\S\ref{subsec:across_models}) and hardware (Figure~\ref{fig:throughput_vs_rps_pat_fcfs_feather_eval}), the RL policy was reset and learned online from scratch, with no state carried over from other models or GPUs.
Next, we generate a workload consisting of five prefix groups, each containing 5K tokens. The input request rate varies over time---from 5 to 10 to 20 requests per second---and we measure the average batch size as a function of time for \Feather using its Bandit policy. We see in Figure~\ref{fig:rl_policy_across_varying_worklaod} that the average batch size increases in response to increasing incoming load that makes larger homogeneous batches possible. Note that fluctuations in the plot are due both to RL exploration and the workload being Poisson. Because Feather's RL state and reward are defined relative to the current workload and hardware rather than in absolute terms, the learned policy generalizes across configurations without retraining.

Interested readers can see \S\ref{sec:additional_results_of_feather} for additional results.

\section{Conclusion}
\label{sec:conclusion}
In this paper, we showed that maximizing batch size alone is not sufficient for efficient LLM serving during the decode phase. Instead, we identified \textit{prefix homogeneity}, the extent of the prefix shared across \textit{all} requests in a batch, as a more important factor for improving memory locality and reducing redundant KV cache accesses. Higher batch sizes lead to better GPU utilization but also reduce the length of the prefix shared across all requests. Existing approaches ignore this aspect and also rely on expensive prefix detection mechanisms that introduce significant CPU overhead. To address these gaps, we proposed \Feather, a lightweight prefix-aware scheduler that balances this trade-off between batch size and prefix homogeneity. It uses a Chunked Hash Tree (CHT) for fast prefix detection and a reinforcement learning-based batching policy that decides when to stop batch formation based on the loss in prefix homogeneity due to the addition of a new request. Our results show that \Feather consistently improves end-to-end throughput across diverse workloads while keeping scheduling overhead low. As future work, we plan to extend this approach to distributed and multi-GPU settings. We are also looking into more adaptive and lightweight learning policies that can improve robustness across different workloads and deployment scenarios.

\bibliographystyle{ACM-Reference-Format}
\bibliography{references}

\appendix

\section{Additional Results from \S\ref{subsec:significance_of_prefix_homogeneity}}
\label{sec:additional_results_section_3}
Below, we present a set of experiments to strengthen our observations from \S\ref{subsec:significance_of_prefix_homogeneity} regarding prefix homogeneity.

\subsection{Effect of Total KV Cache Size}
\label{subsec:cache_working_set_theory}
\begin{figure}[t]
\captionsetup[subfigure]{labelformat=simple}
    \renewcommand\thesubfigure{(\alph{subfigure})}
\centering
\begin{subfigure}{0.49\columnwidth}
    \centering
\includegraphics[width=\linewidth]{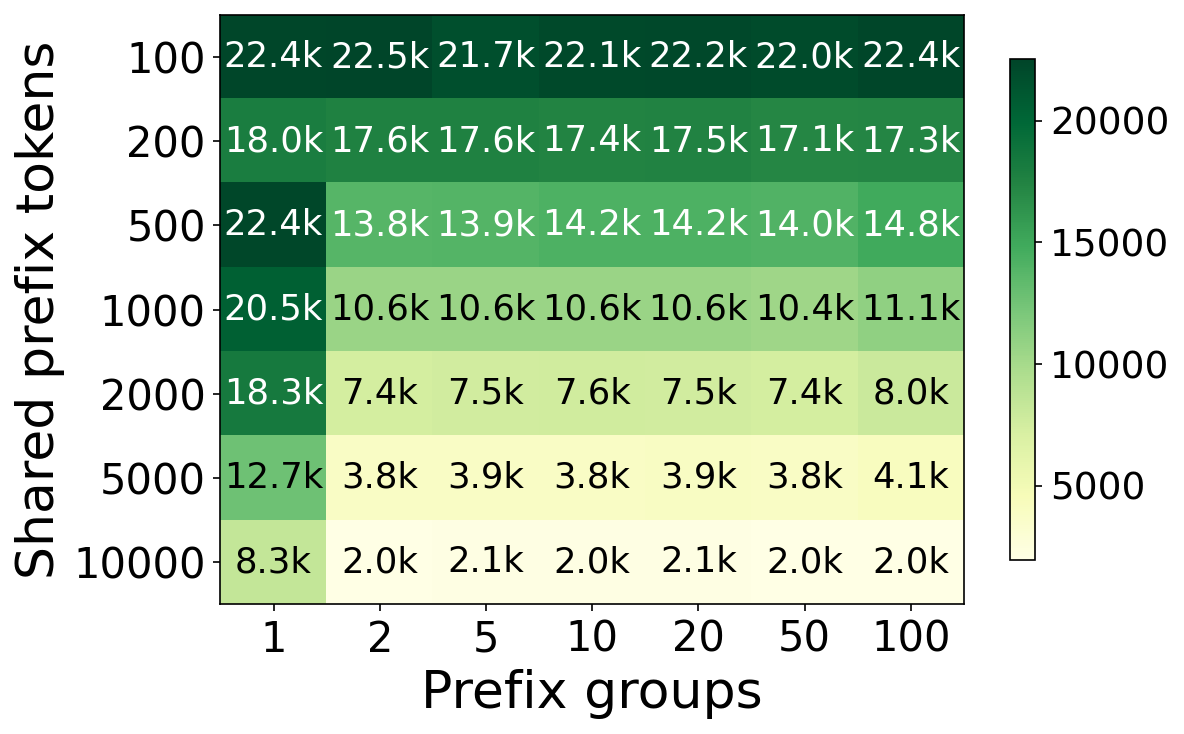}
\caption{Throughput}
\label{fig:tp_qwen}
\end{subfigure}
\hfill
\begin{subfigure}{0.49\columnwidth}
    \centering
\includegraphics[width=\linewidth]{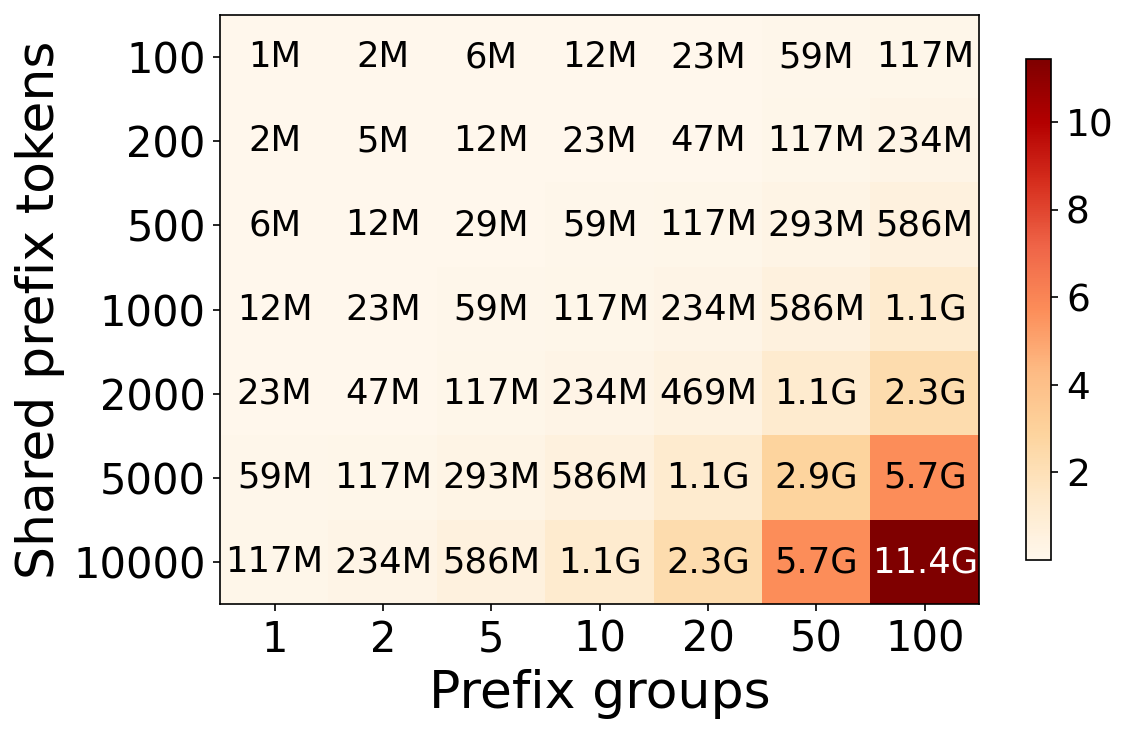}
\caption{KV Cache Memory}
\label{fig:kvcache}
\end{subfigure}
\caption{Throughput and KV Cache Size}
\label{fig:qwen_tp_kvcache}
\end{figure}

Figure~\ref{fig:qwen_tp_kvcache} illustrates vLLM's FCFS throughput alongside the total KV cache memory footprint for prefix tokens in the Qwen 0.5B model, evaluated across varying prefix lengths and numbers of prefix groups. As expected, throughput decreases as the prefix sequence length increases. A more pronounced drop is observed when the number of prefix groups increases from 1 to 2, which can be attributed to a loss of locality. Beyond this point, however, throughput remains relatively stable. This suggests that increasing the number of prefix groups further does not significantly degrade temporal or spatial locality, indicating that most of the locality loss occurs in the transition from a single group to multiple groups. Moreover, this drop from 1 to 2 prefix groups is observable only for larger shared prefix lengths where temporal and spatial locality benefits kick in.

Another question to ask if the high throughput values are correlated with the entire KV cache working set fitting into one of the cache levels. L1 cache size is typically in KBs, which is too small even for tens of tokens. Therefore, we focus on the L2 cache, which is 96 MB in size for the GPU on which these experiments were run. In the top-left region of the heatmap, the entire KV cache fits within this size. However, no clear throughput degradation is observed once this threshold is exceeded. This indicates that effective spatial and temporal locality depend more on whether the subset of KV blocks actively accessed by batched requests at a given time (i.e. the dynamic working set) fits in the L2 cache. It is intuitive that this dynamic working set is smaller in size for prefix-homogeneous batches which explains the higher throughput gains achieved for such batches. Moreover, even for such a small model, the model weights are already large enough to prevent the entire KV cache to reside in L2. Therefore, performance gains due to prefix homogeneity primarily stem from a smaller dynamic working set of the KV cache.

\begin{figure}[t]
\captionsetup[subfigure]{labelformat=simple}
    \renewcommand\thesubfigure{(\alph{subfigure})}
\centering
\begin{subfigure}{0.49\columnwidth}
    \centering
\includegraphics[width=\columnwidth]{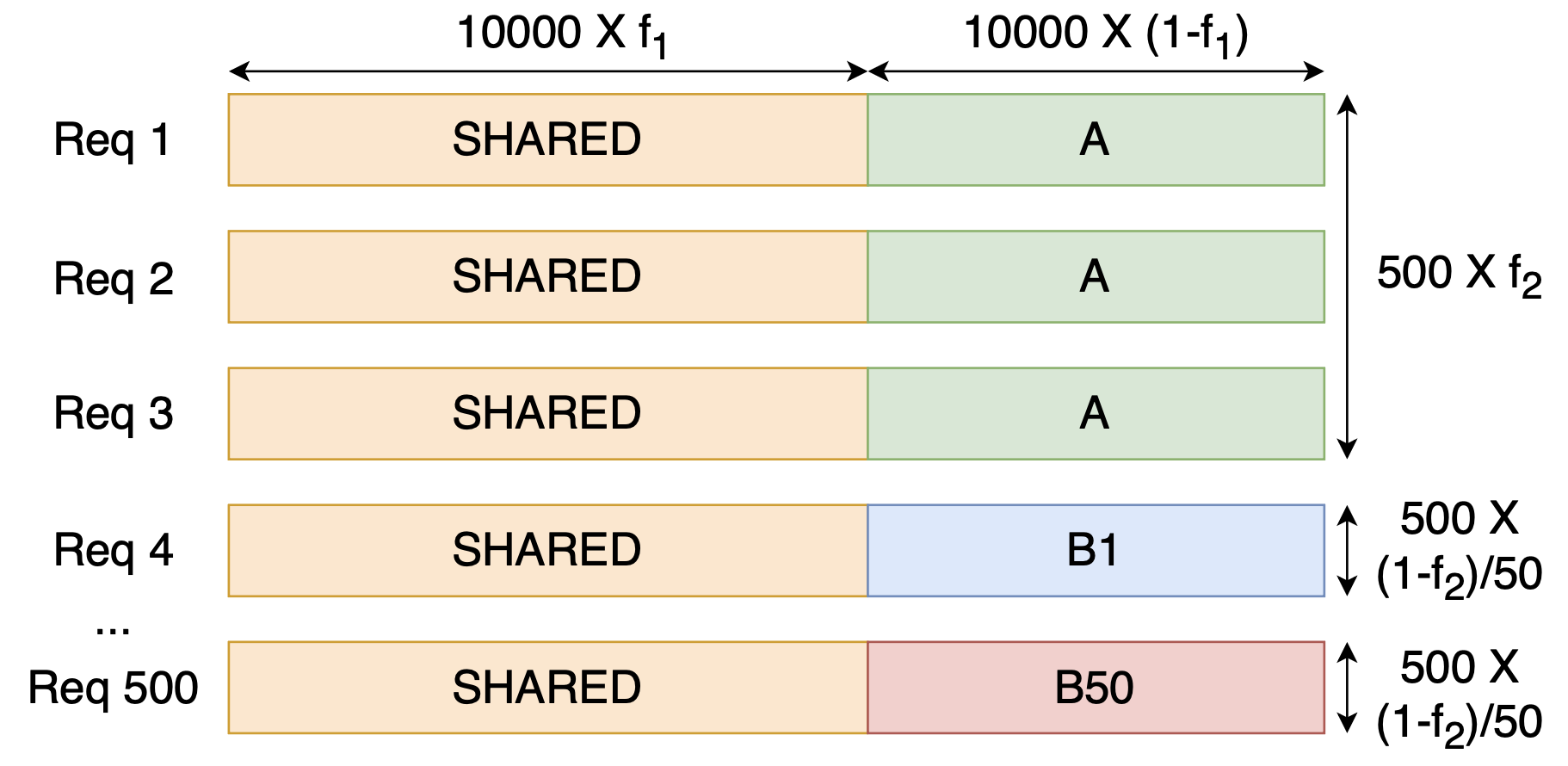}
\caption{Visualization}
\label{fig:vis_tiered_prefix_sharing}
\end{subfigure}
\hfill
\begin{subfigure}{0.49\columnwidth}
   \centering
\includegraphics[width=\columnwidth]{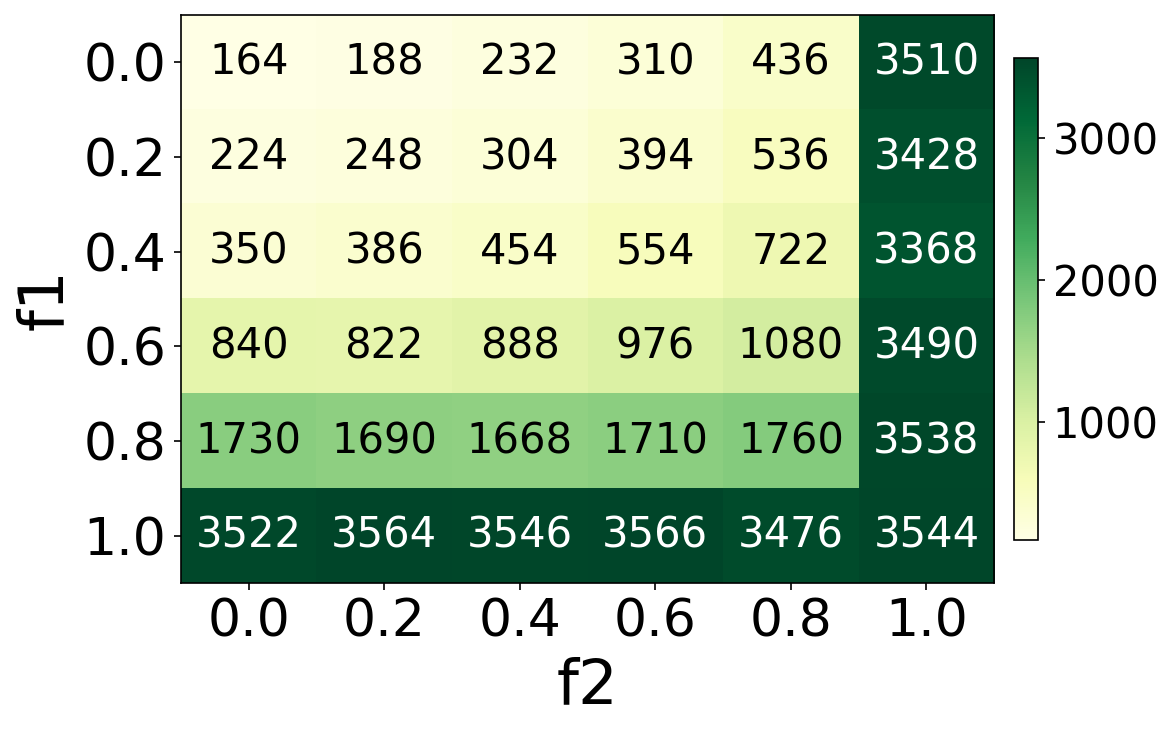}
\caption{Throughput}
\label{fig:tp_tiered_prefix_sharing}
\end{subfigure}
\caption{Key Observations on Prefix Homogeneity}
\label{fig:tiered_prefix_sharing_fig}
\end{figure}

\subsection{Key Observations on Prefix Homogeneity}
\label{subsec:tiered_prefix_sharing}
We consider an additional experiment where each request has a sequence length of $10K$ tokens. As illustrated in Figure~\ref{fig:vis_tiered_prefix_sharing}, a fraction $f_1 \in [0, 1]$ of this sequence is shared as a common prefix across all $500$ requests in the batch. The remaining $10K(1 - f_1)$ tokens constitute the suffix of each request. Within the batch, a fraction $f_2 \in [0, 1]$ of requests share an identical suffix, denoted as $A$. The remaining fraction $(1 - f_2)$ has suffixes drawn from one of $50$ distinct groups, $B_1, \dots, B_{50}$. Figure~\ref{fig:tp_tiered_prefix_sharing} presents the throughput as we vary $f_1$ and $f_2$. Consistent with the observations in \S\ref{subsec:significance_of_prefix_homogeneity}, increasing $f_1$ improves throughput due to enhanced temporal and spatial locality in KV cache accesses. Additionally, we observe a sharp increase in throughput when $f_2$ changes from $0.8$ to $1$. This occurs because at $f_2 = 1$, the entire batch becomes homogeneous, resulting in fully aligned KV cache traversal. Throughput is significantly lower at smaller values of $f_1$, primarily because limited GPU memory leads to frequent cache evictions. This is further supported by the increase in throughput with rising $f_2$ at low $f_1$, as more requests share the suffix $A$, reducing eviction frequency. However, once the working set fits within GPU memory (around $f_1 = 0.8$), throughput remains nearly constant for all $f_2 < 1$, indicating that residual heterogeneity has minimal impact. Overall, this experiment consolidates the key insights discussed in \S\ref{subsec:significance_of_prefix_homogeneity}.

\section{Operations of Chunked Hash Tree}
\label{sec:algorithms}
This section provides pseudo-codes and details of various algorithms used in the Chunked Hash Tree (\S\ref{subsec:chunked_hash_tree}).

\subsection{Variables and Notations}
Table~\ref{tab:cht_variables_and_notations} provides a list of all the variables and notations used in the following pseudo-codes.

\begin{table}[t]
\centering
\small
\resizebox{\columnwidth}{!}{\begin{tabular}{l p{0.8\columnwidth}}
\toprule
\textbf{Variable} & \textbf{Explanation} \\
\midrule
$T$      & Sequence length in tokens  \\
$C$      & Number of chunks in a sequence  \\
$K$      & Chunk size  \\
$H$      & Hash vector  \\
$r$      & Request  \\
$W$      & Number of waiting requests \\
$W_h$    & Average number of waiting requests sharing a hash \\
$(l^*, h^*)$ & Shared prefix tip \\
$h_l^r$ & Hash at level $l$ of request $r$ \\
$\texttt{working\_set}$ & Working set \\
$\texttt{waiting}$ & Reverse index from hashes to waiting requests \\
$\texttt{waiting\_heap}$ & Min-heap \\
$\texttt{request\_hashes}$ & Map from request to its hash vector \\
$\texttt{miss}$ & Map from request to its missing count \\
$\texttt{active}$ & Set of requests present in the current batch\\
$\texttt{waiting\_requests}$ & Set of waiting requests \\
$\texttt{ref}$ & Map from hash to number of active requests sharing it \\
\bottomrule
\end{tabular}}
\caption{Variables and Notations Used}
\label{tab:cht_variables_and_notations}
\end{table}

\subsection{Prefix Hash Computation}
\label{subsec:chunked_hash_tree_prefix_representation}
\begin{figure}[t]
\centering
\includegraphics[width=0.7\columnwidth]{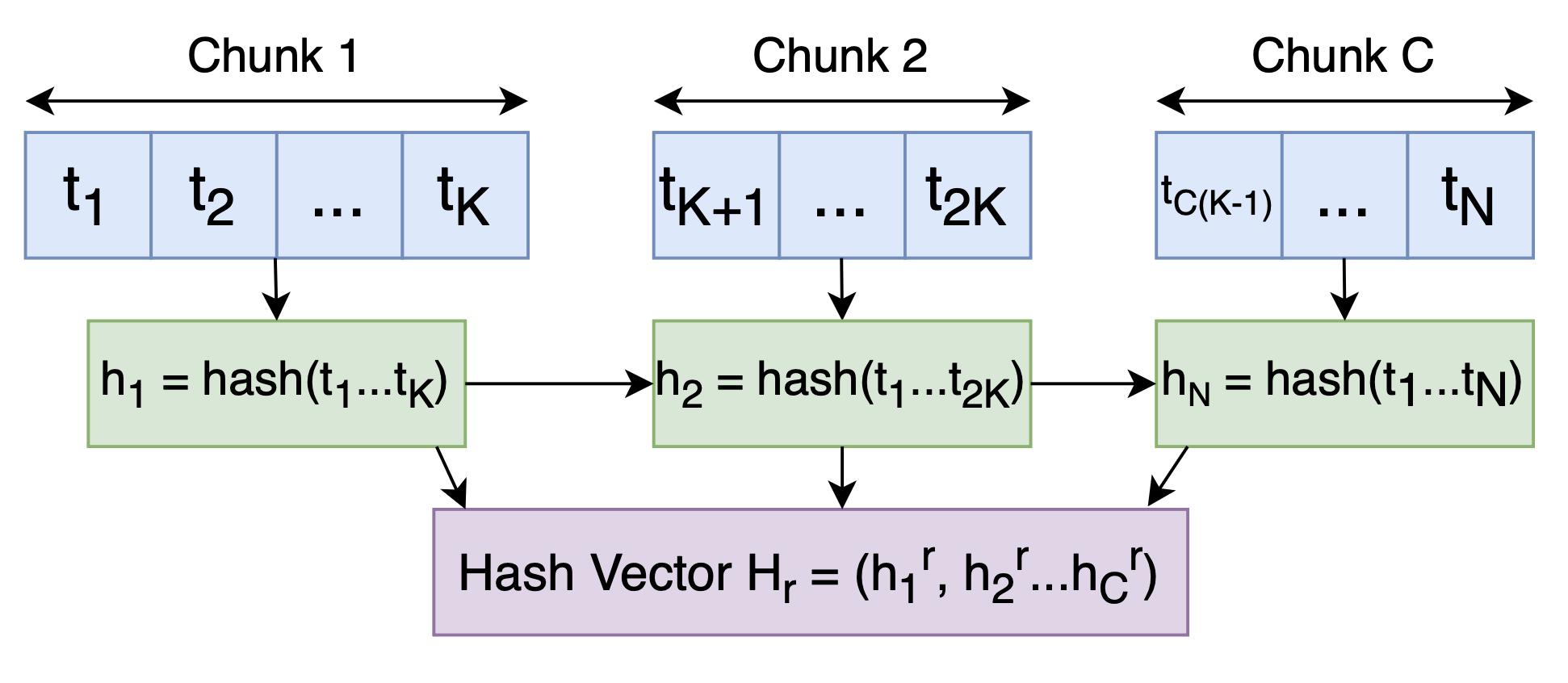}
\caption{Hash Computation in Chunked Hash Tree}
\label{fig:chunked_hash_tree_prefix_representation}
\end{figure}

The goal is to efficiently compute prefix-aligned hashes at fixed chunk boundaries. Algorithm~\ref{alg:compute_hashes} computes the hash vector for a token sequence using an incremental hashing scheme (Figure~\ref{fig:chunked_hash_tree_prefix_representation}). Instead of recomputing hashes for each prefix independently, the algorithm maintains a streaming hash state that is updated once per token. The algorithm processes each token exactly once and performs constant-time updates to the hash state. As a result, it runs in $\mathcal{O}(T)$ time and produces $C = \lceil T / K \rceil$ hash values, yielding $\mathcal{O}(C)$ output space. While Algorithm~\ref{alg:compute_hashes} processes tokens sequentially, the prefix-hash vector can equivalently be computed in parallel across chunks. The key property we want is that all requests sharing the first $c \cdot K$ tokens produce an identical $h_c$. This is achieved by partitioning the token sequence into chunks $S_c = (t_{(c-1)K+1}, \dots, t_{cK})$, hashing each chunk independently in parallel, and then chaining the results via $h_c = \textsc{Hash}(h_{c-1} \,\|\, \textsc{Hash}(S_c))$, with $h_0$ set to a fixed initialization constant (Algorithm~\ref{alg:compute_hashes_parallel}). With parallel workers, the dominant hashing phase runs in $\mathcal{O}(K)$ per worker rather than $\mathcal{O}(T)$ sequentially, reducing overall wall-clock time to $\mathcal{O}(K + C)$; the subsequent chaining pass costs only $\mathcal{O}(C)$ and is negligible in practice. This parallelization is most beneficial when $K$ is large, as each chunk contains more tokens and therefore represents a heavier independent unit of work, allowing parallel workers to amortize scheduling overhead across a larger payload and yielding greater absolute speedup over the sequential baseline. Both algorithms produce a hash vector satisfying the prefix-consistency property---all requests sharing the first $c \cdot K$ tokens yield the same $h_c$, yet the two vectors differ in their actual hash values, as Algorithm~\ref{alg:compute_hashes_parallel} is a distinct formulation rather than a parallelized execution of Algorithm~\ref{alg:compute_hashes}.

\begin{algorithm}[b]
\caption{\textsc{ComputeHashes}(S, K)}
\label{alg:compute_hashes}
\begin{algorithmic}[1]
\Require Token sequence $(t_1, \dots, t_T)$; chunk size $K$
\Ensure Cumulative prefix-hash vector $H = (h_1, \dots, h_C)$
\State $H \gets []$;\quad $C \gets \lceil T / K \rceil$
\State Initialise incremental \textsc{xxHash-64} state $s$
\For{$i \gets 1$ \textbf{to} $T$}
  \State Feed $t_i$ into $s$
  \If{$i \bmod K = 0$ \textbf{or} $i = T$}
    \State $H.\textsc{Append}(\textsc{Digest}(s))$
  \EndIf
\EndFor
\State \Return $H$
\end{algorithmic}
\end{algorithm}

\begin{algorithm}[t]
\caption{\textsc{ComputeHashesParallel}(S, K)}
\label{alg:compute_hashes_parallel}
\begin{algorithmic}[1]
\Require Token sequence $(t_1, \dots, t_T)$; chunk size $K$
\Ensure Cumulative prefix-hash vector $H = (h_1, \dots, h_C)$
\State $C \gets \lceil T / K \rceil$
\State $h_0 \gets \textsc{InitConst}$
\State \textbf{parallel for} $c \gets 1$ \textbf{to} $C$ \textbf{do} \Comment{Independent per-chunk hashes}
  \State \hspace{\algorithmicindent} $S_c \gets (t_{(c-1)K+1}, \dots, t_{\min(cK,\,T)})$
  \State \hspace{\algorithmicindent} $\ell_c \gets \textsc{Hash}(S_c)$ \Comment{Each chunk hashed independently}
\State \textbf{end parallel for}
\For{$c \gets 1$ \textbf{to} $C$} \Comment{Sequential prefix chaining}
  \State $h_c \gets \textsc{Hash}(h_{c-1} \,\|\, \ell_c)$
\EndFor
\State \Return $H = (h_1, \dots, h_C)$
\end{algorithmic}
\end{algorithm}

\subsection{Insertion}
Algorithm~\ref{alg:insert} provides the pseudo-code for the \textsc{Insert} procedure. Given a request with a token sequence, we compute its cumulative prefix-hash vector using \textsc{ComputeHashes} and store it in a dictionary that maps each request to its hashes for easier reuse later (lines 1-2). For each level, we retrieve the corresponding hash and check whether the tuple is present in the working set; if not, the missing count is incremented (lines 6-7). Simultaneously, the request is inserted into the reverse index (line 9), which maps each hash to the set of waiting requests containing it and supports efficient lookup of prefix-sharing candidates. After processing all levels, we store the miss count in a map and insert the tuple of miss count and the request into the min-heap, which is keyed by missing count, enabling efficient retrieval of the most compatible request (lines 11-12). The design avoids any updates to the working set or reference counts during insertion, ensuring low overhead. 

\begin{algorithm}[t]
\caption{\textsc{Insert}(r, S)}
\label{alg:insert}
\begin{algorithmic}[1]
\Require $r$; token sequence $S$; $\mathit{working\_set}$; $\mathit{waiting}$; $\mathit{waiting\_heap}$
\Ensure $r$ registered and ready for scheduling
\State $H_r \gets \textsc{ComputeHashes}(S, K)$
\State $\mathit{request\_hashes}[r] \gets H_r$
\State $m_r \gets 0$
\For{$\ell \gets 1$ \textbf{to} $|H_r|$}
  \State $h \gets h_\ell^r$
  \If{$(\ell, h) \notin \mathit{working\_set}$}
    \State $m_r \gets m_r + 1$
  \EndIf
  \State $\mathit{waiting}[h] \gets \mathit{waiting}[h] \cup \{r\}$
\EndFor
\State $\mathit{miss}[r] \gets m_r$
\State $\mathit{waiting\_heap}.\textsc{Push}(m_r,\; r)$
\end{algorithmic}
\end{algorithm}

\subsection{Finding the Best Request}
The \textsc{FindBest} procedure selects the most compatible waiting request by querying the min-heap while skipping stale entries arising from lazy updates until a valid minimum is found (lines 6-8). Once a candidate is identified, it is reinserted into the heap for future activations in case the RL policy decides to add this request to the batch (line 9), and three quantities are computed without modifying the system state: the current tip depth, the prospective depth, and the number of peers sharing the resulting prefix. The prospective depth is obtained via a backward scan over the cumulative hash vector, selecting the deepest level such that the corresponding level-hash tuple exists in the working set, which corresponds to the least common ancestor with the active set under the current state. The peer count is computed by counting waiting requests whose hash at that level matches that of the candidate request. The result is cached and reused until either the working set or the waiting queue changes, avoiding redundant recomputation across repeated calls. If no valid request is found, the procedure returns $\bot$.

\begin{algorithm}[t] 
\caption{\textsc{FindBest}()} 
\label{alg:find_best} 
\begin{algorithmic}[1] 
\Require $\mathit{waiting\_heap}$; $\mathit{miss}$; $\mathit{active}$; tip $(\ell^*, h^*)$; $\mathit{working\_set}$; $\mathit{request\_hashes}$ 
\Ensure $(r^*,\; \ell^*_{\mathit{before}},\; \ell^*_{\mathit{after}},\; \mathit{peers})$ or $\bot$ 
\If{$\mathit{waiting\_requests} = \emptyset$}\ 
\Return $\bot$ 
\EndIf 
\If{$\mathit{cache\_valid}$}\ 
\Return $\mathit{cached\_result}$ 
\EndIf 
\While{$\mathit{waiting\_heap} \neq \emptyset$} 
\State $(m, r) \gets \mathit{waiting\_heap}.\textsc{Pop}()$ 
\If{$r \in \mathit{active}$ \textbf{or} $r \notin \mathit{miss}$ \textbf{or} $\mathit{miss}[r] \neq m$}\ 
\textbf{continue} 
\EndIf 
\State $r^* \gets r$;\quad $\mathit{waiting\_heap}.\textsc{Push}(m, r^*)$ \Comment{restore for future calls} 
\State $\ell^*_{\mathit{before}} \gets \ell^*$ 
\State $\ell^*_{\mathit{after}} \gets \ell$ where $\ell$ is the largest $\ell \leq \min(\ell^*, |H_{r^*}|)$ s.t. $(\ell, h_\ell^{r^*}) \in \mathit{working\_set}$ \State $\mathit{peers} \gets |\{w \in \mathit{waiting\_requests} : h_{\ell^*_{\mathit{after}}}^w = h_{\ell^*_{\mathit{after}}}^{r^*}\}|$ 
\State Cache and \Return $(r^*,\; \ell^*_{\mathit{before}},\; \ell^*_{\mathit{after}},\; \mathit{peers})$ 
\EndWhile 
\State \Return $\bot$ 
\end{algorithmic} 
\end{algorithm}

\subsection{Adding a Request to the Active Batch}
Given a request with its prefix-hash vector, the \textsc{AddToBatch} procedure promotes it to the active set (line 1) and updates the working set, reference counts, and other metadata in a single backward pass over the prefix-hash vector. For each level, if the reference count of the corresponding hash is zero (no other active request shares it), the level-hash tuple is inserted into the working set, and all corresponding waiting requests have their miss counts decremented and are reinserted into the waiting heap (lines 7-12). The reference counts are incremented at all levels (line 14). In parallel, the shared prefix tip is recomputed by tracking the deepest level at which the level-hash tuple remains in the working set; the first such match during the backward traversal determines the new tip (lines 15-17), while the loop continues to ensure that all state updates are applied. If this is the first active request, the tip is initialized directly to the end of its prefix-hash vector.

\begin{algorithm}[t]
\caption{\textsc{AddToBatch}(r)}
\label{alg:add_to_batch}
\begin{algorithmic}[1]
\Require $H_r$; $\mathrm{ref}$; $\mathit{working\_set}$; $\mathit{waiting}$; $\mathit{miss}$; $\mathit{waiting\_heap}$; tip $(\ell^*, h^*)$; $\mathit{active}$
\Ensure Updated $\mathrm{ref}$, $\mathit{working\_set}$, $\mathit{miss}$, $(\ell^*, h^*)$

\State $\mathit{active} \gets \mathit{active} \cup \{r\}$;\quad $\mathit{waiting\_requests} \gets \mathit{waiting\_requests} \setminus \{r\}$

\If{$|\mathit{active}| = 1$}
  \State $(\ell^*, h^*) \gets (|H_r|,\; h_{|H_r|}^r)$;\quad \Return
\EndIf

\State $\ell_{\mathrm{tip}} \gets \min(\ell^*,\; |H_r|)$;\quad $(\ell^*, h^*) \gets (0, \bot)$

\For{$\ell \gets |H_r|$ \textbf{downto} $1$}
  \If{$\mathrm{ref}[h_\ell^r] = 0$}
    \State $\mathit{working\_set} \gets \mathit{working\_set} \cup \{(\ell, h_\ell^r)\}$
    \For{$w \in \mathit{waiting}[h_\ell^r],\; w \notin \mathit{active}$}
      \State $\mathit{miss}[w] \mathrel{-}= 1$; 
      \State $\mathit{waiting\_heap}.\textsc{Push}(\mathit{miss}[w], w)$
    \EndFor
  \EndIf
  \State $\mathrm{ref}[h_\ell^r] \mathrel{+}= 1$
  \If{$\ell^* = 0$ \textbf{and} $\ell \leq \ell_{\mathrm{tip}}$ \textbf{and} $(\ell, h_\ell^r) \in \mathit{working\_set}$}
    \State $(\ell^*, h^*) \gets (\ell,\; h_\ell^r)$
    \Comment{Least common ancestor found; continue loop for remaining working set updates}
  \EndIf
\EndFor
\end{algorithmic}
\end{algorithm}

\subsection{Removing a Request from the Active Batch}
Given a completed request, the \textsc{Finish} procedure removes it from the active set (line 1) and updates the working set, reference counts, and other metadata. A forward pass over the prefix-hash vector decrements the reference count at each level (line 3); if it drops to zero, the level-hash tuple is removed from the working set, and all the corresponding waiting requests have their miss counts incremented and are reinserted into the min-heap (lines 4-10). After updating the working set, the shared prefix tip is recomputed by attempting to extend it forward: starting from the previous tip level, we check successive levels up to the minimum length among active requests and extend the tip whenever the reference count at that level equals the number of active requests, indicating agreement across all of them; the scan terminates at the first disagreement (lines 19-26). If the active set becomes empty, the tip is reset to $(0, \bot)$ (lines 12-14), and if only one request remains, the tip is set to its full length (lines 15-18). Finally, \textsc{Remove} is invoked to clean auxiliary structures such as the reverse index and the stored hash vector (line 27).

\begin{algorithm}[t]
\caption{\textsc{Finish}(r)}
\label{alg:finish}
\begin{algorithmic}[1]
\Require $H_r$; $\mathrm{ref}$; $\mathit{working\_set}$; $\mathit{waiting}$; $\mathit{miss}$; $\mathit{waiting\_heap}$; tip $(\ell^*, h^*)$; $\mathit{active}$
\Ensure Updated $\mathrm{ref}$, $\mathit{working\_set}$, $\mathit{miss}$, $(\ell^*, h^*)$

\State $\mathit{active} \gets \mathit{active} \setminus \{r\}$

\For{$\ell \gets 1$ \textbf{to} $|H_r|$}
  \State $\mathrm{ref}[h_\ell^r] \mathrel{-}= 1$
  \If{$\mathrm{ref}[h_\ell^r] = 0$}
    \State $\mathit{working\_set} \gets \mathit{working\_set} \setminus \{(\ell, h_\ell^r)\}$
    \For{$w \in \mathit{waiting}[h_\ell^r],\; w \notin \mathit{active}$}
      \State $\mathit{miss}[w] \mathrel{+}= 1$;
      \State $\mathit{waiting\_heap}.\textsc{Push}(\mathit{miss}[w], w)$
    \EndFor
  \EndIf
\EndFor

\If{$\mathit{active} = \emptyset$}
  \State $(\ell^*, h^*) \gets (0, \bot)$;\quad \Return
\EndIf
\If{$|\mathit{active}| = 1$}
  \State let $q$ be the sole element of $\mathit{active}$
  \State $(\ell^*, h^*) \gets (|H_q|,\; h_{|H_q|}^q)$;\quad \Return
\EndIf

\While{$\ell^* + 1 \leq \min_{q \in \mathit{active}} |H_q|$}
  \Comment{try to extend tip forward}
  \State $\ell \gets \ell^* + 1$
  \If{$\mathrm{ref}[h_\ell^{q_0}] = |\mathit{active}|$ for arbitrary $q_0 \in \mathit{active}$}
    \Comment{all active requests agree at level $\ell$}
    \State $(\ell^*, h^*) \gets (\ell,\; h_\ell^{q_0})$
  \Else
    \State \textbf{break}
  \EndIf
\EndWhile
\State \textsc{Remove}(r)
\Comment{cleans reverse index and request hashes}
\end{algorithmic}
\end{algorithm}

\subsection{Batch Formation}
\textsc{BuildBatch} describes the entire pipeline, where we incrementally construct a batch by repeatedly querying \textsc{FindBest} to obtain the most compatible candidate, along with summary statistics, which are combined with the current batch size to form the RL state (lines 3-6). A policy (heuristic, bandit, or Q-learning) maps this state to an action, either \textsc{Add} or \textsc{Stop} (line 7); if \textsc{Add} is selected, the request is activated via \textsc{AddToBatch} and appended to the batch; otherwise, the loop terminates (lines 8-10). The procedure continues until the waiting queue is exhausted or the policy decides to stop, after which the constructed batch is executed for one decode iteration to obtain the observed throughput (line 12). This throughput serves as the reward signal used to update the policy (line 13), with bandit methods updating their statistics and Q-learning applying a Bellman update with parameters $\alpha$ and $\gamma$, while heuristic policies incur no update.

\begin{algorithm}[t]
\caption{\textsc{BuildBatch}()}
\label{alg:rl_batch}
\begin{algorithmic}[1]
\Require Waiting queue $Q$; policy $\pi$ (heuristic, bandit, or Q-learning); discount $\gamma$; learning rate $\alpha$
\Ensure Batch $B$ ready for one decode iteration
\State $B \gets \emptyset$
\While{$Q \neq \emptyset$}
    \State $(r^*, c_{\mathrm{before}}, c_{\mathrm{after}}, w) \gets \textsc{FindBest}()$
    \If{$r^* = \bot$} \textbf{break} \EndIf
    \State $s \gets (|B|,\; \Delta = c_{\mathrm{before}} - c_{\mathrm{after}},\; w)$
    \State $a \gets \pi(s)$ \Comment{heuristic rule / UCB look-up / $\varepsilon$-greedy Q-table}
    \If{$a = \textsc{Stop}$} \textbf{break} \EndIf
    \State \textsc{AddToBatch}($r^*$); \quad $B \gets B \cup \{r^*\}$
\EndWhile
\State $\text{throughput}_{\mathrm{decode}} \gets \textsc{ExecuteBatch}(B)$
\State \textbf{Update} $\pi$ with observed reward $\text{throughput}_{\mathrm{decode}}$ \Comment{no-op for heuristic}
\State \Return $B$
\end{algorithmic}
\end{algorithm}

\subsection{Complexity Analysis}
Table~\ref{tab:complexity} summarizes the time complexity of each operation. Tip maintenance in \textsc{AddToBatch} and \textsc{Finish} is subsumed within their $\mathcal{O}(C \cdot W_h \cdot \log W)$ passes and adds no asymptotic cost. In practice, the dominant cost is miss-count propagation, which updates all affected waiting requests per working-set change. Result caching avoids redundant \textsc{FindBest} traversals between mutations, keeping scheduling overhead low. We present the time taken by individual functions for a particular experimental setup in \S\ref{subsec:time_taken_individual_functions_chunked_hash_tree}. 

\begin{table}[t]
\centering
\begin{tabular}{ll}
\hline
\textbf{Operation} & \textbf{Complexity} \\
\hline
\textsc{Insert}     & $\mathcal{O}(T + C \log W)$ \\
\textsc{AddToBatch}  & $\mathcal{O}(C \cdot W_h \cdot \log W)$ \\
\textsc{Finish}      & $\mathcal{O}(C \cdot W_h \cdot \log W)$ \\
\textsc{FindBest}    & $\mathcal{O}(\log W)$ amortized; $\mathcal{O}(1)$ cached \\
\textsc{Remove}      & $\mathcal{O}(C)$ \\
\hline
\end{tabular}
\caption{Time complexity of CHT scheduler operations}
\label{tab:complexity}
\end{table}

\subsection{Find Best Request - Alternative Heuristic}
\label{subsec:find_best_reques_alternative_heuristic}
\begin{figure}[t]
\centering
    \includegraphics[width=0.25\linewidth]{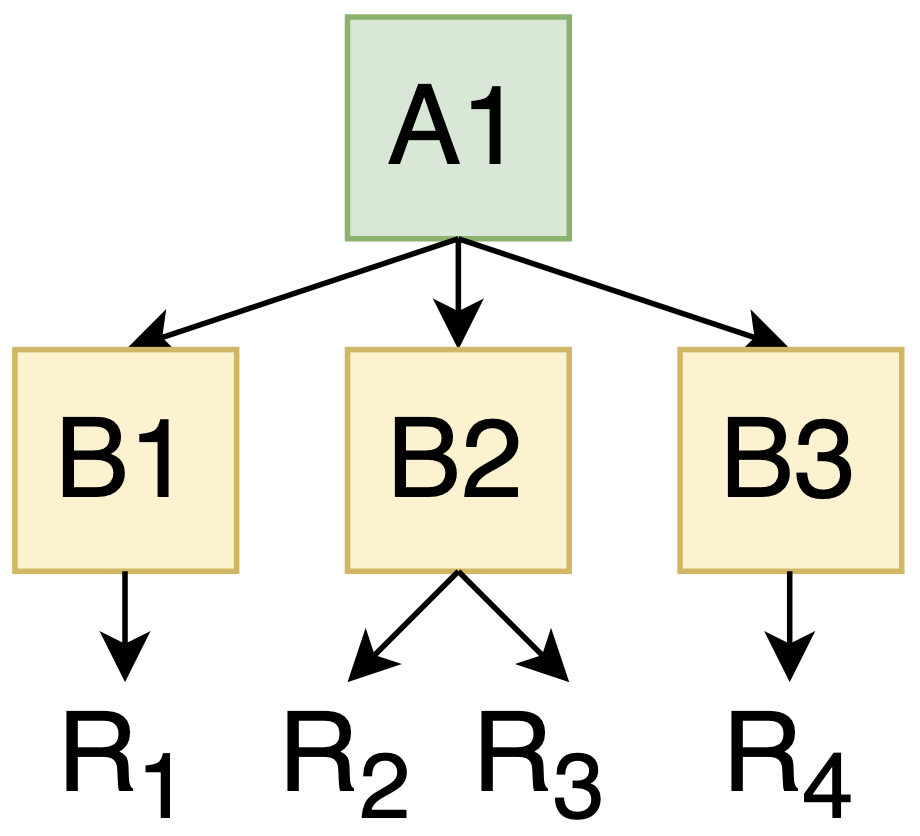}
    \caption{\textsc{FindBest} - Alternative Heuristic Example}
    \label{fig:find_best_request_different_approach_example}
\end{figure}

An intuitive alternative to using the missing count (which is based on the difference with the working set) is to select a waiting request that maximizes the depth of the current shared prefix tip. However, this approach has a subtle flaw: the tip is inherently bottlenecked by the most divergent active request, limiting future extension. Consider a scenario where Level 1 is shared by all requests, but Level 2 splits into three branches: one for $R_1$, one shared by $R_2$ and $R_3$, and one for $R_4$ (Figure~\ref{fig:find_best_request_different_approach_example}). If the active batch contains $R_1$ and $R_2$, the shared prefix tip is restricted to Level 1. Selecting based only on this tip makes $R_3$ and $R_4$ appear equally viable. If the scheduler admits $R_4$, the tip remains at Level 1; when $R_1$ finishes, the remaining requests ($R_2$ and $R_4$) lie on divergent branches, leaving the tip still stuck. Our Chunked Hash Tree avoids this by evaluating candidates against the full working set using the missing count $m_r$, rather than anchoring to the bottlenecked tip. The working set tracks all prefix chunks present in any active request. In the example, since $R_2$ is active, its Level 2 chunk is already in the working set. Thus, $R_3$ requires fewer missing chunks than $R_4$, i.e., $m_{R_3} < m_{R_4}$. Selecting $R_3$ preserves future overlap: although the immediate tip remains bottlenecked, once $R_1$ finishes and is removed via \textsc{Finish}, the remaining batch $\{R_2, R_3\}$ shares a deeper prefix, and the tip extends to Level 2. 

So, we have two ways to find the best waiting request to be added to the batch. One is based on the difference from the working set, and the other is based on the shared prefix tip. Next, we try to formalize the metrics on which both of these ways are based. Let $W$ be the hashes in the current working set, which can be defined as: $\cup_{r \in A}H^r$, where $A$ is the active set of requests and $H^r$ is the set of hashes for request $r$. CHT selects the request that has the minimum miss-count, which can be defined as $m_W(r) = |H^r \setminus W|$. The shared prefix tip is the deepest level shared by all the requests of the active batch, which can be defined as $\max\{\ell : \forall r \in A, (\ell, h_\ell^r) \in W\}$. The alternative heuristic selects the request that leads to the maximum depth of the shared prefix tip, which is equivalent to minimizing the number of hashes missing from the working set at or above the current tip for that request. Let $m_S(r) = |S \setminus H^r|$, where $S$ is the set of all hashes at or above the current tip level in the working set, i.e., $S = \{h : (\ell, h) \in W, \ell \leq \ell^*\}$, where $\ell^*$ is the current tip level. That is, $m_S(r)$ denotes the number of hash chunks present in $S$, but not present in the request $r$. Therefore, we explore the alternative heuristic that minimizes $m_S(r)$ instead of $m_W(r)$.

Let us assume that all the requests have a length of $C$ chunks, and the best requests emitted from the two metrics are $r_W$ and $r_S$, respectively. We adopt this equal-length assumption only to simplify this illustrative proof; it is not required by the Chunked Hash Tree system itself, which handles requests of differing hash-vector lengths directly---for instance, \textsc{FindBest} and \textsc{AddToBatch} (Algorithm~\ref{alg:find_best}, line~11, and Algorithm~\ref{alg:add_to_batch}, line~5) explicitly bound the relevant traversal by $\min(\ell^*, |H_r|)$ to accommodate variable-length prefixes. Let us also assume that the shared prefix tip is at level $\ell^*$. We aim to show that $m_S(r_W) = m_S(r_S)$, that is, the working set metric causes equal loss in the shared prefix tip, implying that $r_W$ does not lead to less depth in the shared prefix tip compared to $r_S$. Let us assume, to the contrary, that $m_S(r_S) < m_S(r_W)$. Now, we compare $m_W(r_W)$ and $m_W(r_S)$ and aim to show that $m_W(r_W) > m_W(r_S)$, which contradicts the fact that $r_w$ minimizes the metric $m_W$.

\begin{figure}[t]
\captionsetup[subfigure]{labelformat=simple}
    \renewcommand\thesubfigure{(\alph{subfigure})}
\centering
\begin{subfigure}{0.49\linewidth}
    \centering
\includegraphics[width=\linewidth]{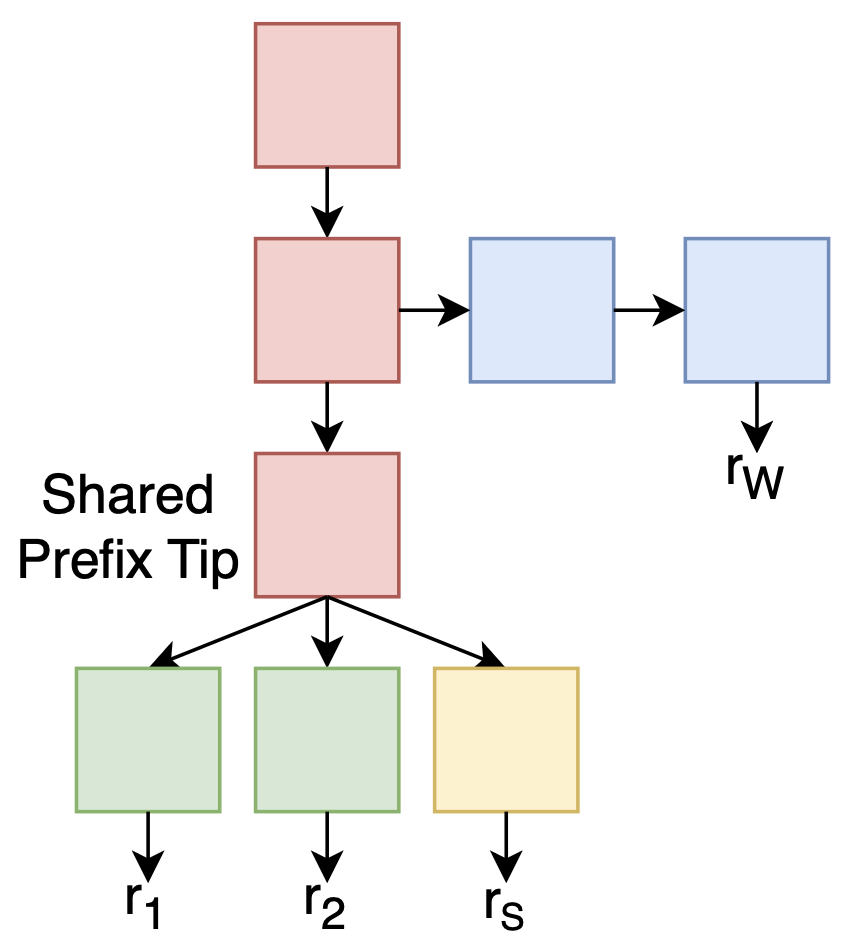}
\caption{Case 1}
\label{fig:alternative_heuristic_case1}
\end{subfigure}
\hfill
\begin{subfigure}{0.49\linewidth}
    \centering
\includegraphics[width=\linewidth]{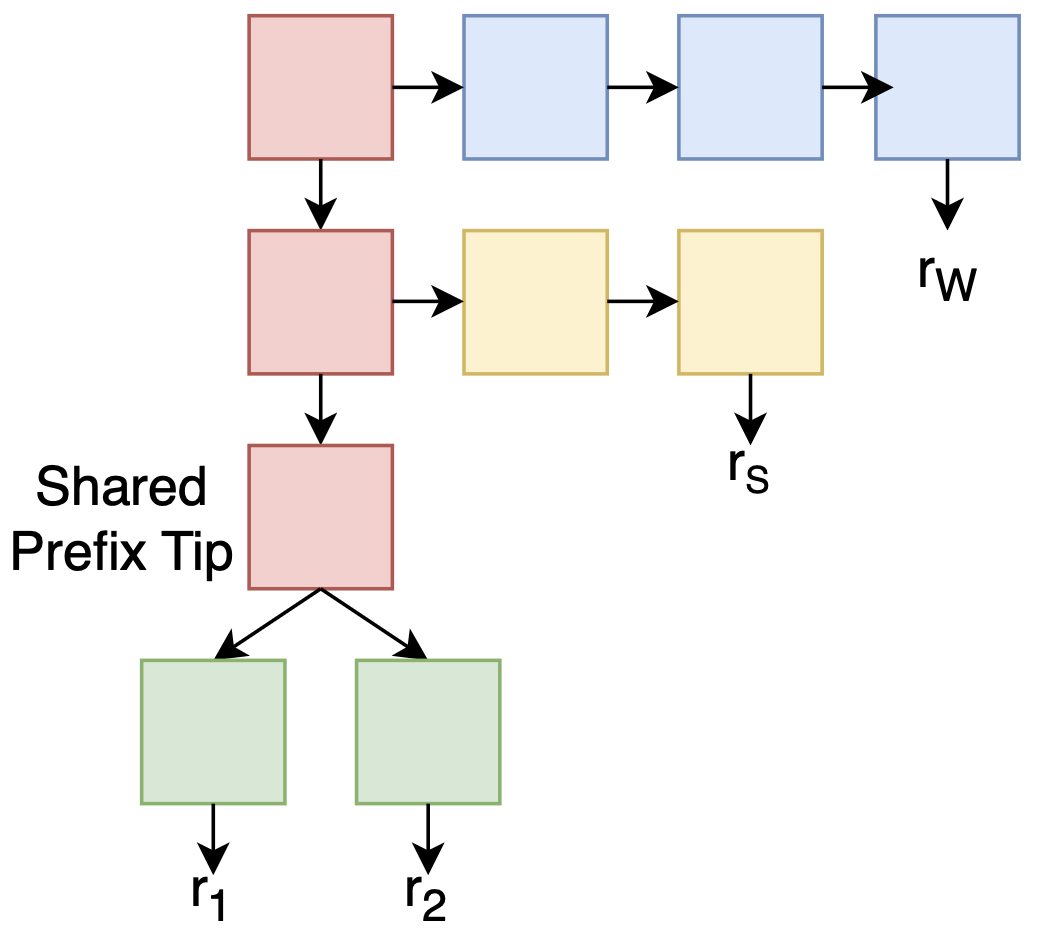}
\caption{Case 2}
\label{fig:alternative_heuristic_case2}
\end{subfigure}
\caption{\textsc{FindBest} cases: $r_1$ and $r_2$ are both present in the active batch, and hence the working set is the union of all of their chunks. The shared prefix tip is at the third level, and the set $S$ of all shared prefix chunks are marked in red. In the first case, $r_S$ fully matches the shared prefix tip, whereas $r_W$ hangs from the middle. For this case, $m_W(r_W) = 2, m_W(r_S) = 1, m_S(r_W) = 1, m_S(r_S) = 0$. In the second case, both the requests $r_S$ and $r_W$ do not fully match the shared prefix tip, however $r_W$ hangs off from an earlier chunk. For this case, $m_W(r_W) = 3, m_W(r_S) = 2, m_S(r_W) = 2, m_S(r_S) = 1$. Therefore, for both the cases, $m_W$ and $m_S$ exhibit similar behaviour.}
\label{fig:alternative_heuristic_cases}
\end{figure}

Since $S$ represents the shared prefix of all active requests up to depth $\ell^*$, the working set $W$ contains exactly one unique hash per level for all levels $\ell \leq \ell^*$, which constitutes the set $S$. This property dictates that if any request $x$ misses a hash in $S$ at some level $k < \ell^*$, its prefix diverges from the \textit{entire} active batch at level $k$. Consequently, due to the cumulative nature of the chunked hashes, $x$ cannot share any hashes with $W$ at any level $\geq k$. Specifically, if a request does not fully match the shared tip (i.e., $m_S(x) > 0$), it cannot overlap with any deeper divergent branches in $W$. For such a request, its overlap with the working set is exactly its overlap with the shared tip: $|H^x \cap W| = |H^x \cap S| = \ell^* - m_S(x)$. Using our contrary assumption $m_S(r_S) < m_S(r_W)$, we evaluate the cost of both requests under the working set metric, $m_W(x) = C - |H^x \cap W|$. We consider two cases for $r_S$: 

\noindent \textbf{Case 1:} $r_S$ fully matches the shared prefix tip ($m_S(r_S) = 0$). Since $r_S$ matches $S$ entirely, $|H^{r_S} \cap S| = \ell^*$. It may also overlap with deeper branches in $W$, meaning $|H^{r_S} \cap W| \geq \ell^*$. Therefore, its cost is $m_W(r_S) \leq C - \ell^*$. Because $m_S(r_W) > m_S(r_S) = 0$, $r_W$ diverges before the tip. As established, this means $|H^r_W \cap W| = \ell^* - m_S(r_W) < \ell^*$. Thus, $m_W(r_W) = C - |H^r_W \cap W| > C - \ell^*$. Combining these inequalities yields: $m_W(r_S) < m_W(r_W)$ (Figure~\ref{fig:alternative_heuristic_case1}).

\noindent \textbf{Case 2:} $r_S$ diverges before the shared prefix tip ($m_S(r_S) > 0$). Since $m_S(r_W) > m_S(r_S) > 0$, both requests diverge before $\ell^*$. Their overlap with the working set is strictly limited to their overlap with $S$. Thus, $|H^{r_S} \cap W| = \ell^* - m_S(r_S)$ and $|H^r_W \cap W| = \ell^* - m_S(r_W)$. Because $m_S(r_S) < m_S(r_W)$, it logically follows that $|H^{r_S} \cap W| > |H^r_W \cap W|$. Consequently: $m_W(r_S) < m_W(r_W)$ (Figure~\ref{fig:alternative_heuristic_case2}).

In both cases, assuming $m_S(r_S) < m_S(r_W)$ unavoidably leads to $m_W(r_S) < m_W(r_W)$. However, this directly contradicts our initial premise that $r_W$ is the optimal request chosen by the Chunked Hash Tree metric (which guaranties $m_W(r_W) \leq m_W(q)$ for all waiting requests $q$).  Therefore, our assumption must be false, proving that $m_S(r_W) \leq m_S(r_S)$. Because $r_S$ is defined as the absolute minimum for $m_S$, it must strictly hold that $m_S(r_W) = m_S(r_S)$. This completes the proof that optimizing for the working set metric is mathematically guaranteed to be at least as optimal as the shared prefix tip metric for preserving the immediate tip depth while simultaneously exposing deeper amortization opportunities in $W \setminus S$.

\begin{figure*}[t]
\captionsetup[subfigure]{labelformat=simple}
    \renewcommand\thesubfigure{(\alph{subfigure})}
\centering
\begin{subfigure}{0.32\textwidth}
    \centering
\includegraphics[width=\linewidth]{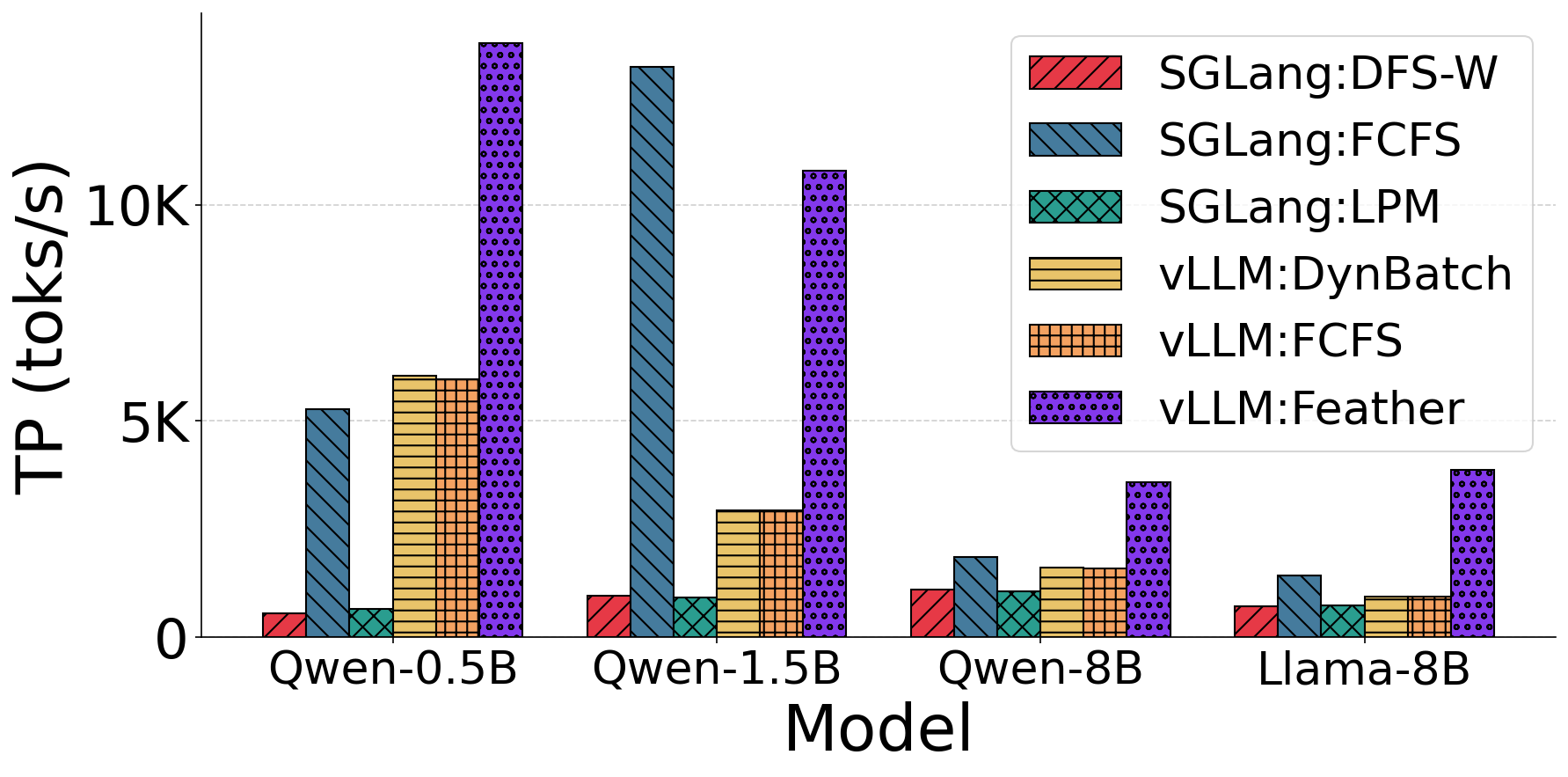}
\caption{Throughput}
\label{fig:tp_vs_mod}
\end{subfigure}
\hfill
\begin{subfigure}{0.32\textwidth}
    \centering
\includegraphics[width=\linewidth]{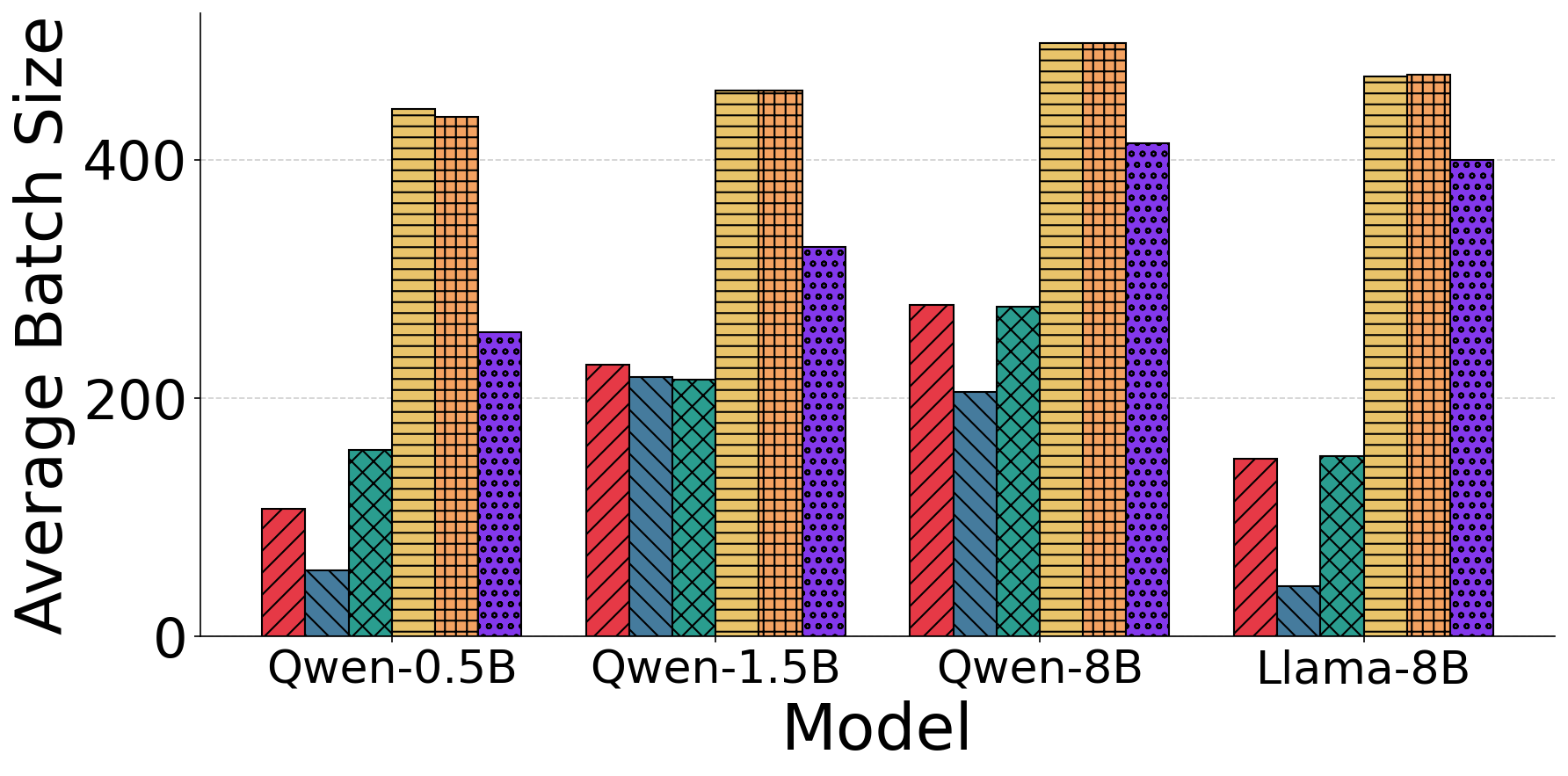}
\caption{Average running batch size}
\label{fig:abs_vs_mod}
\end{subfigure}
\hfill
\begin{subfigure}{0.32\textwidth}
    \centering
\includegraphics[width=\linewidth]{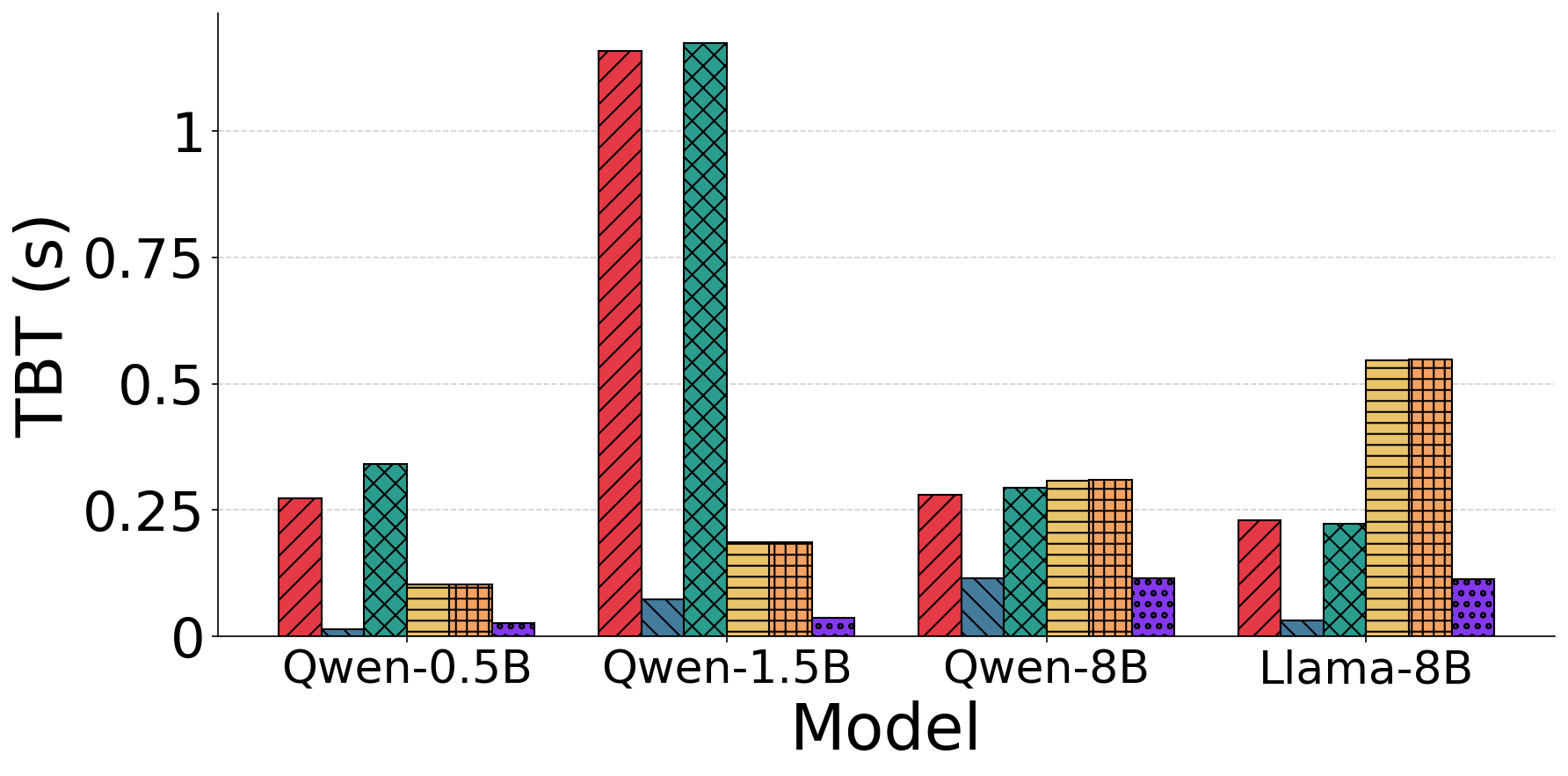}
\caption{Time between tokens (TBT)}
\label{fig:tbt_vs_mod}
\end{subfigure}
\caption{Different Models}
\label{fig:models}
\end{figure*}

\section{Additional Results of \Feather}
\label{sec:additional_results_of_feather}
Our evaluation uses prompts drawn from real datasets (L-Eval and LongBench), but the grouping structure of requests---number of prefix groups, group sizes, and arrival timing---is controlled explicitly to isolate the effect of prefix homogeneity. Production traces would additionally exhibit irregular prefix grouping, variable prefix and decode lengths within a group, and bursty, time-varying arrivals, none of which we evaluate directly due to the unavailability of public production traces exhibiting prefix-sharing structure. We partially stress several of these axes independently: Figure~\ref{fig:tp_vs_rps_l_5000_fam_5} varies the Poisson arrival rate from 1 to 100 req/s, Figure~\ref{fig:tp_vs_fam_l_5000} varies the number of prefix groups from 1 to 100, and Figure~\ref{fig:tp_vs_sys_eval} varies the shared prefix length from 100 to 10,000 tokens. \Feather's gains hold consistently across each of these axes in isolation, which gives us confidence in its robustness to realistic workloads, though we have not validated performance under the joint irregularity of a live production trace. Below, we present additional evaluation results of \Feather\ that could not be included in the main body due to space constraints.

\subsection{Time Taken by Individual Functions of CHT}
\label{subsec:time_taken_individual_functions_chunked_hash_tree}
Table~\ref{tab:chunked_hash_tree_individual_functions} profiles CHT operations over an inference workload of $5K$ requests, each $5K$ tokens long. \textsc{Insert}, \textsc{AddToBatch}, and \textsc{Finish} are each invoked $5K$ times, once per request. \textsc{FindBest}, however, is called $26{,}405$ times because it is queried repeatedly during batch construction. The RL policy may reject candidates and trigger fresh lookups, and the scheduler may inspect the queue multiple times before committing to a \textsc{Stop}. \textsc{Insert} dominates total time because hash computation scans every token once, making it proportional to prompt length. \textsc{Finish} is more expensive than \textsc{AddToBatch}, despite their symmetric roles. This gap arises from \textsc{Finish}'s additional forward tip-extension scan. After updating the working set, it traverses forward, level by level, to check if the shared prefix can be extended. In contrast, \textsc{AddToBatch} performs an early-terminating backward LCA scan, avoiding this overhead. \textsc{FindBest} is the cheapest per-call operation due to result caching and lazy heap traversal.

\begin{table}[t]
\centering
\small
\resizebox{\columnwidth}{!}{\begin{tabular}{lccc}
\toprule
\textbf{Function} & \textbf{Number of Calls} & \textbf{Average Time (ns)} & \textbf{Total Time (ns)} \\
\midrule
\textsc{Insert}      & 5,000  & 153,714 & 768,570,573 \\
\textsc{AddToBatch}  & 5,000  &  11,143 &  55,715,469 \\
\textsc{Finish}      & 5,000  &  38,952 & 194,760,111 \\
\textsc{FindBest}    & 26,405 &   6,641 & 175,372,986 \\
\bottomrule
\end{tabular}}
\caption{Time taken by individual functions of CHT}
\label{tab:chunked_hash_tree_individual_functions}
\end{table}

\subsection{Performance of \Feather across Models}
\label{subsec:across_models}
Figures~\ref{fig:tp_vs_mod}, \ref{fig:abs_vs_mod}, \ref{fig:tbt_vs_mod} present throughput, average batch size, and time between tokens (TBT) across models of varying sizes. The overall trend is consistent with expectations: smaller models achieve higher throughput owing to lower per-token compute costs. \Feather outperforms all baselines across all models. The sole exception is Qwen-1.5B, where SGLang FCFS achieves marginally higher throughput; however, \Feather compensates on the latency axis, exhibiting substantially lower TBT for the same model. A key structural observation is that \Feather's relative throughput advantage over the baselines grows with model size. As model size increases, a larger fraction of the total execution time is spent on attention computation, which amplifies the benefits of temporal and spatial KV-cache locality that \Feather exploits. This effect is visible in Figure~\ref{fig:tp_vs_mod}: the gap between \Feather and the next-best baseline widens from Qwen-0.5B to the 8B-scale models. The performance of SGLang's LPM and DFS-W schedulers degrades noticeably for smaller models. Both policies rely on CPU-side tree traversal, whose overhead scales with sequence length rather than model size; as the model shrinks, GPU execution time decreases while scheduler overhead remains roughly constant, causing the CPU to become an increasing bottleneck. \Feather avoids this through its Chunked Hash Tree, which reduces per-step scheduler complexity and keeps CPU overhead negligible relative to GPU execution time across all model scales. Finally, Figure \ref{fig:tbt_vs_mod} highlights that vLLM's FCFS and DB baselines incur substantially higher TBT at larger model sizes, whereas Feather keeps TBT low and stable. 

\subsection{Results across LongChat Model}
\label{subsec:results_across_longchat_model}
\begin{figure*}[t]
\captionsetup[subfigure]{labelformat=simple}
    \renewcommand\thesubfigure{(\alph{subfigure})}
\centering
\begin{subfigure}{0.30\linewidth}
    \centering
\includegraphics[width=\linewidth]{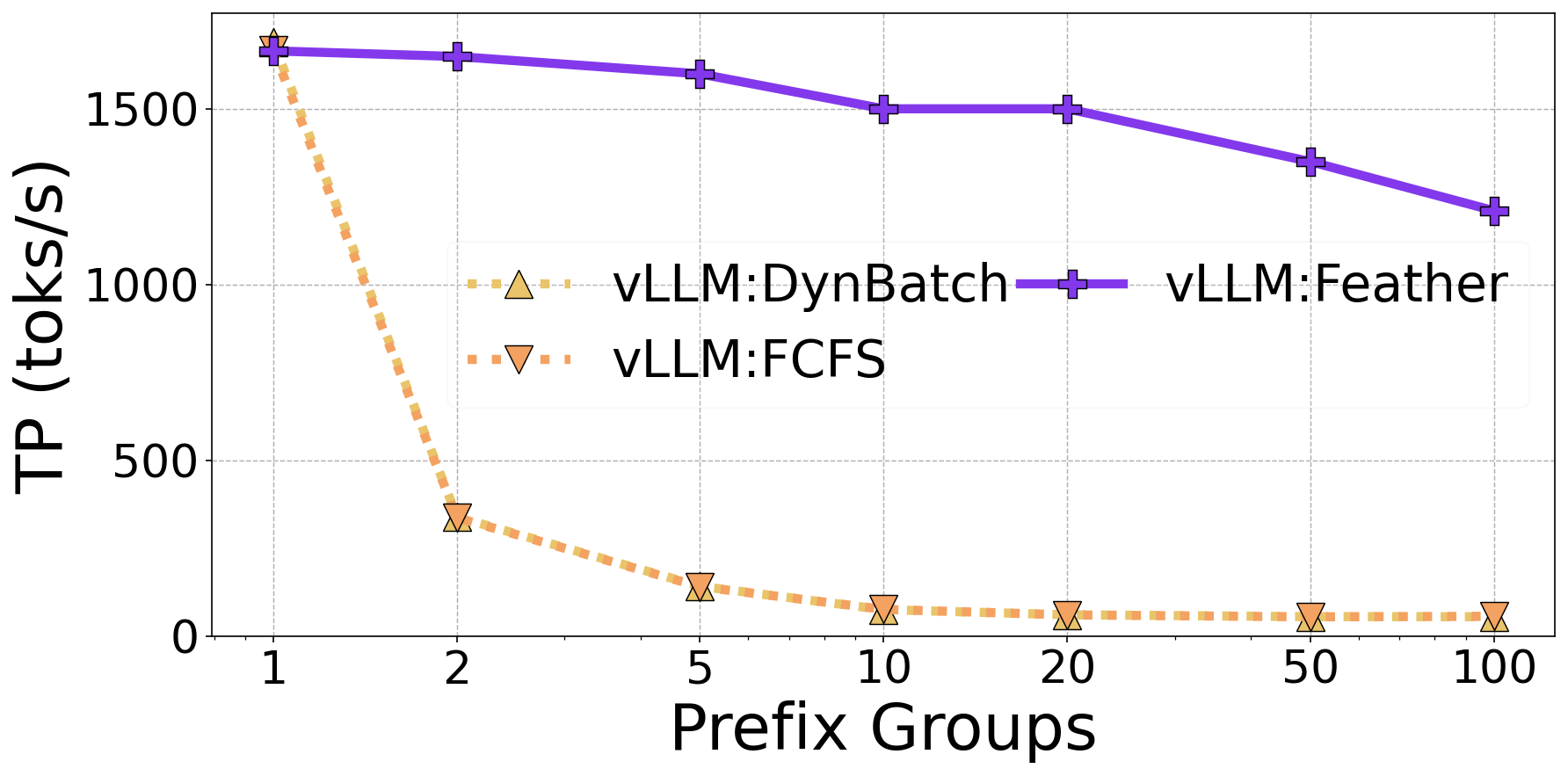}
\caption{Throughput}
\label{fig:tp_vs_pref_long}
\end{subfigure}
\hfill
\begin{subfigure}{0.30\linewidth}
    \centering
\includegraphics[width=\linewidth]{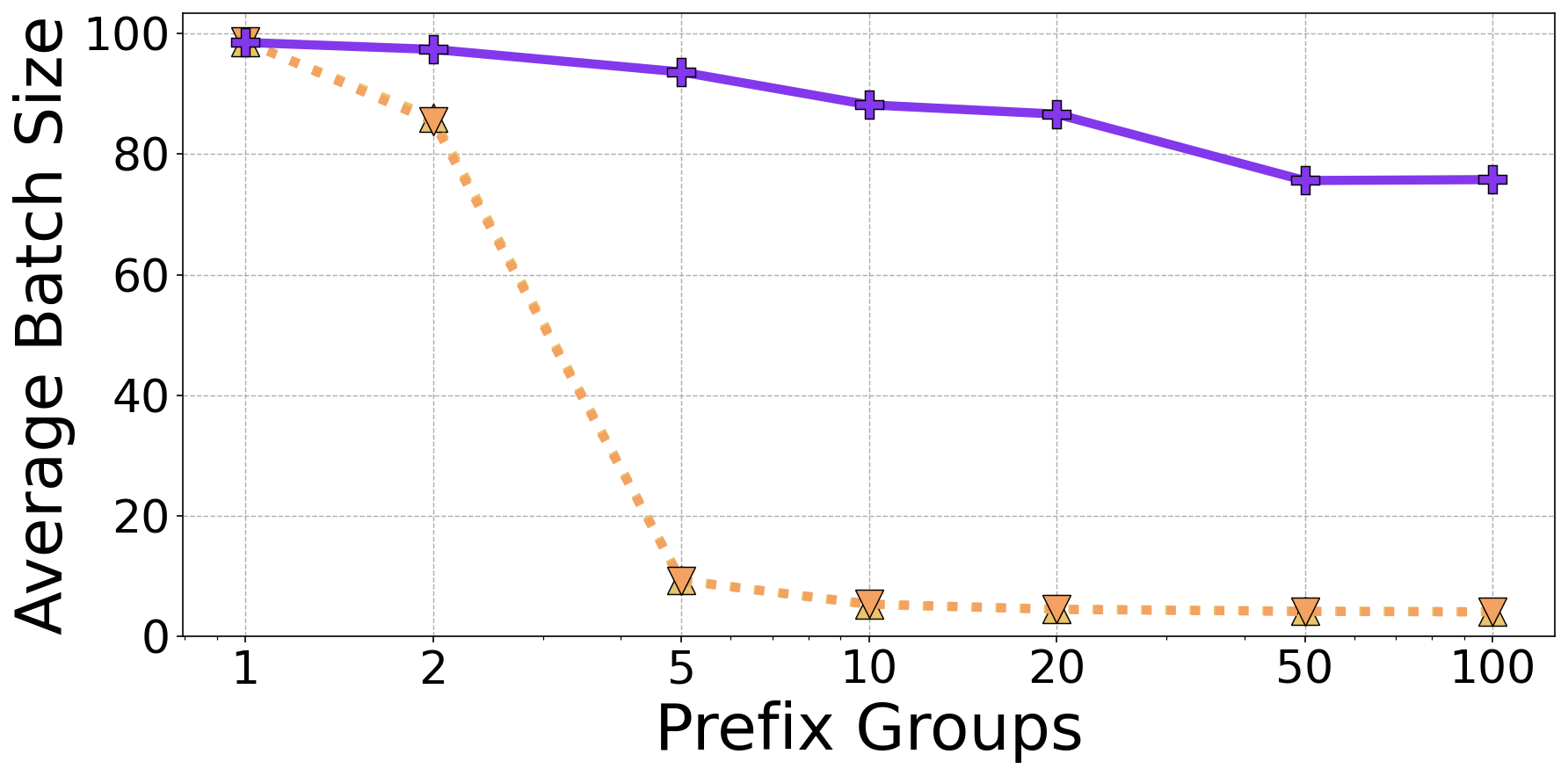}
\caption{Average Batch Size}
\label{fig:abs_vs_pref_long}
\end{subfigure}
\hfill
\begin{subfigure}{0.30\linewidth}
    \centering
\includegraphics[width=\linewidth]{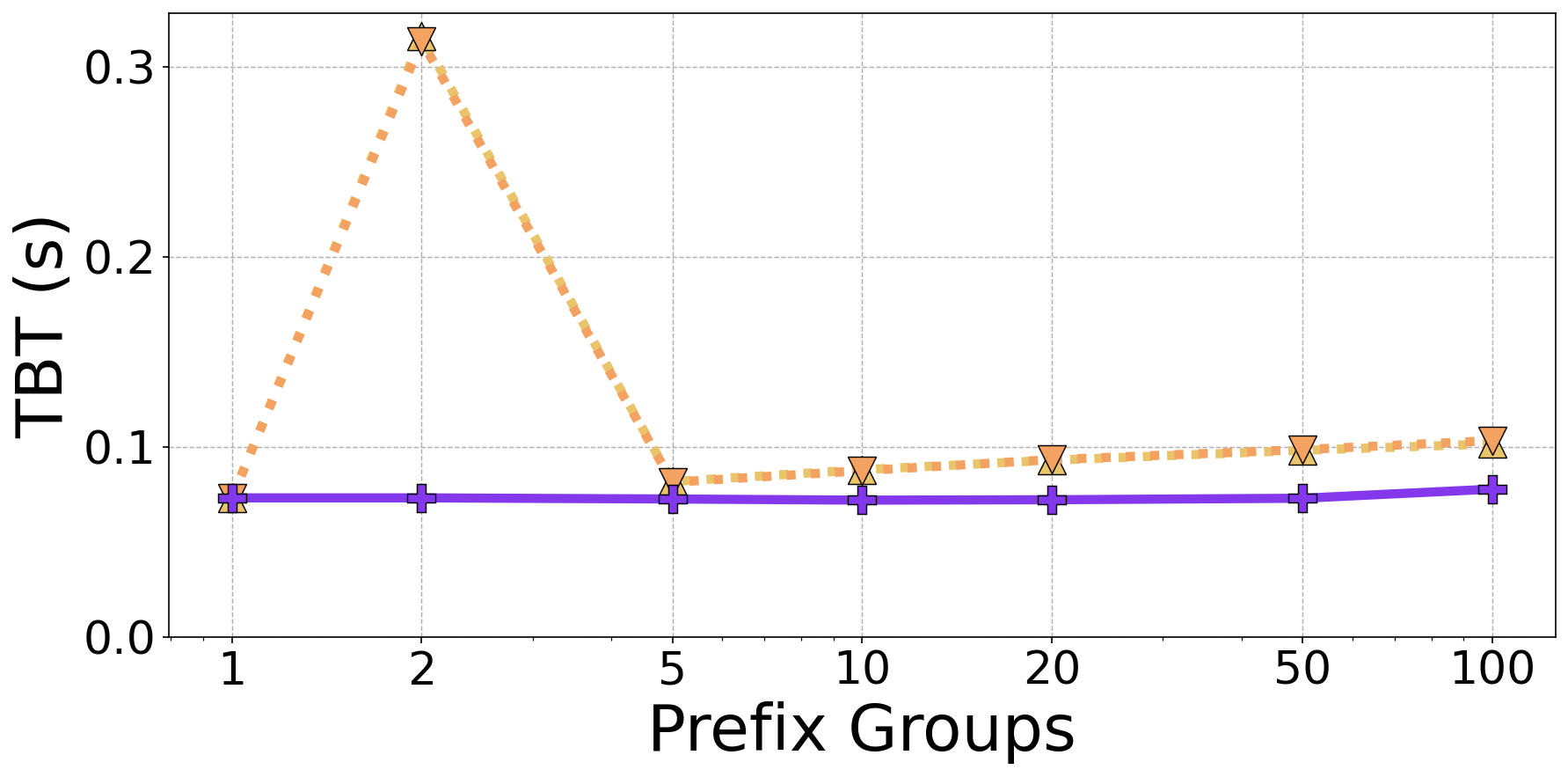}
\caption{Time between tokens (TBT)}
\label{fig:tbt_vs_pref_long}
\end{subfigure}
\caption{Varying number of prefix groups for LongChat 13B}
\label{fig:longhcat_results}
\end{figure*}

Figure~\ref{fig:longhcat_results} presents throughput, average batch size, and time between tokens (TBT) as the number of prefix groups varies for the LongChat 13B model. Overall, vLLM's FCFS and Dynamic Batching behave similarly across most configurations. For FCFS, there is a sharp drop in throughput when moving from 1 to 2 prefix groups, accompanied by an increase in TBT. This is largely due to the loss of locality benefits. The further decline from 2 to 5 prefix groups is driven by KV cache evictions, which is also reflected in the steep reduction in average batch size. Beyond this point, TBT stabilizes at a lower value, primarily because the batches have become much smaller. In contrast, \Feather maintains stable and consistent throughput across all settings. At 100 prefix groups, it achieves up to 22$\times$ higher throughput than FCFS. By forming prefix-homogeneous batches, \Feather not only improves scheduling efficiency but also significantly reduces KV cache evictions through more effective immediate cache reuse.

\subsection{Complexity Analysis of LPM and DFS-W}
\label{subsec:scheduling-complexity}
We provide a detailed complexity analysis of the two SGLang cache-aware scheduling policies discussed in \S\ref{sec:eval}: Longest Prefix Match (LPM) and Depth-First Search with Weighting (DFS-W). Let $W$ denote the number of requests in the waiting queue and $T$ the maximum prefix length. For each request, LPM performs prefix matching on the global radix tree, costing $\mathcal{O}(T)$ per request, and then sorts the queue by matched prefix length, adding $\mathcal{O}(W \log W)$. The total per-round cost is therefore $\mathcal{O}(W \cdot T + W \log W)$. The result is overhead that scales \textit{multiplicatively} with both queue size and context length. LPM implicitly acknowledges this by falling back to FCFS when $W > 128$ (which was disabled in our experiments). DFS-W avoids per-request traversal; instead, it performs a full traversal of the radix tree in each scheduling round. Let $B$ be the total number of radix tree nodes. DFS-W first propagates subtree request counts upward in $\mathcal{O}(B)$, then re-traverses in a depth-first manner, sorting children by weight at every level. The total cost is $\mathcal{O}(W + B \log B)$. The key problem is that $B$ reflects the server's accumulated history, not the current workload. The radix tree grows monotonically and is only pruned under memory pressure, so DFS-W's overhead increases over time, regardless of the current load. A server with a large accumulated cache ($B \gg W$) will incur high scheduling costs even during quiet periods. In summary, a burst of long-context requests penalizes LPM through the $W \cdot T$ term, while a large accumulated cache penalizes DFS-W through the $B \log B$ term. Either way, scheduling overhead can reach the same order of magnitude as GPU execution time. This demonstrates that identifying shared prefixes at runtime is fundamentally expensive and naive cache-aware scheduling can become the dominant bottleneck.

\subsection{Tensor Core Utilization across Policies}
\label{subsec:tensor_core_utilization_policies}
\begin{figure}[t]
\captionsetup[subfigure]{labelformat=simple}
    \renewcommand\thesubfigure{(\alph{subfigure})}
\centering
\begin{subfigure}{0.49\columnwidth}
    \centering
    \includegraphics[width=\linewidth]{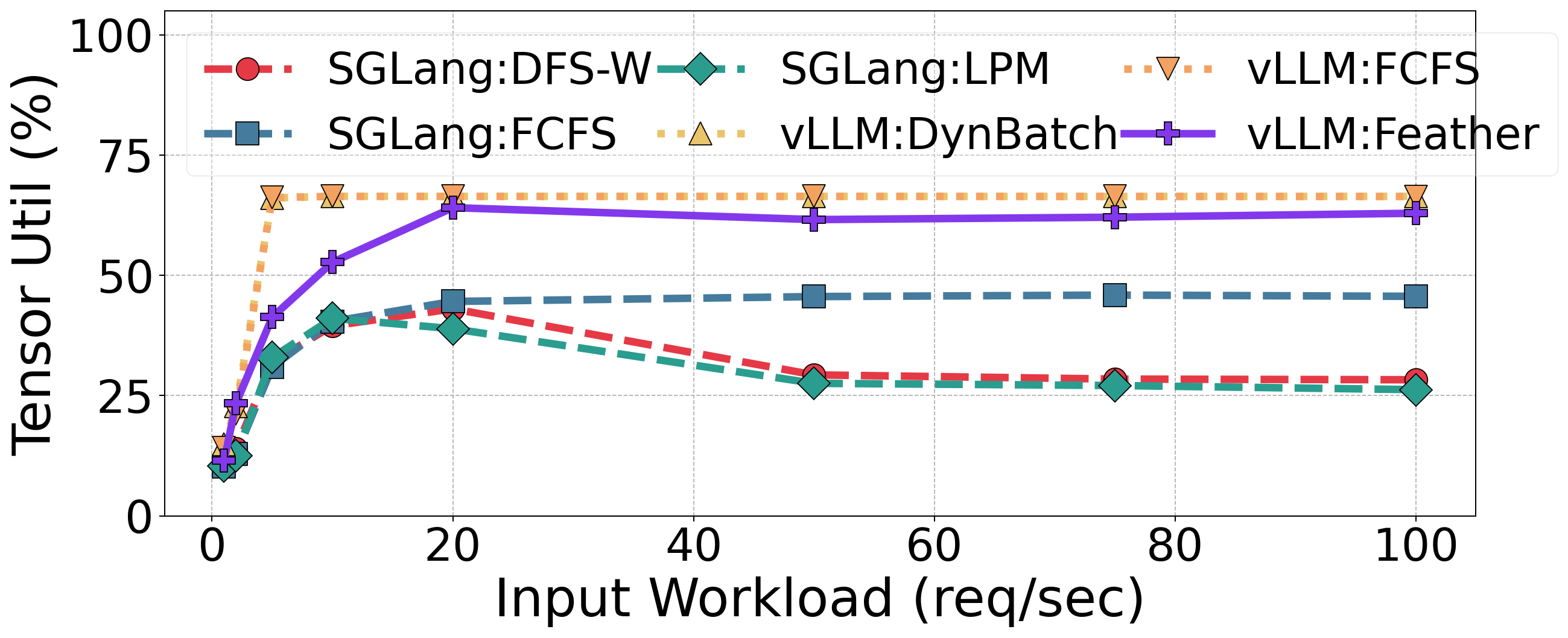}
    \caption{Tensor Core Utilization}
    \label{fig:tensor_core_utilzation_across_policies}
\end{subfigure}
\hfill
\begin{subfigure}{0.49\columnwidth}
    \centering
\includegraphics[width=\linewidth]{results/req_per_sec_plots/avg_batch_vs_rps.png}
\caption{Average Batch Size}
\label{fig:batch_size_across_policies}
\end{subfigure}
\caption{Input Workload}
\label{fig:req_per_sec_tenso_batch}
\end{figure}

Figure~\ref{fig:req_per_sec_tenso_batch} compares tensor core utilization and average batch size across different request rates and scheduling policies. \Feather shows slightly lower utilization at low request rates due to forming smaller batches, but it still delivers higher throughput at these points (as seen earlier in Figure~\ref{fig:tp_vs_rps_l_5000_fam_5}). As the workload increases, \Feather gradually builds larger batches from prefix-homogeneous requests, and its utilization rises to match vLLM's FCFS. In contrast, SGLang's FCFS maintains relatively low utilization because it operates with consistently small batch sizes. Interestingly, even though SGLang's DFS-W and LPM achieve larger batch sizes, their utilization remains low. This is mainly due to CPU-side stalls between GPU executions, during which the GPU stays idle, effectively reducing overall utilization.

\section{Why the name \Feather?}
\label{sec:about_feather}
\Feather is named after the structural pattern of prefix-homogeneous batches in our system: a shared prefix forms a central shaft, from which individual requests branch out as they diverge, much like the barbs of a \Feather. The metaphor captures both the visual structure of our batches and the functional benefit of keeping prefix sharing requests together for better performance.

\end{document}